\RequirePackage{fix-cm}
\documentclass[smallextended]{svjour3}       
\smartqed  
\usepackage{graphicx}
\usepackage{amsmath}

\usepackage{comment}
\usepackage[square,numbers]{natbib}
\usepackage{xcolor}
\usepackage{subfigure}
\usepackage{color}
\usepackage{multirow}
\usepackage{adjustbox}
\usepackage{colortbl}
\usepackage{siunitx}
\usepackage{amssymb}
\usepackage{algorithm}
\usepackage{algorithmic}
\newcolumntype{P}[1]{>{\centering\arraybackslash}p{#1}}
\usepackage[algo2e]{algorithm2e}
\usepackage[title]{appendix}
\usepackage{ragged2e}

\begin{document}

\title{Adaptive Exploitation of Pre-trained Deep Convolutional Neural Networks for Robust Visual Tracking
}


\author{Seyed Mojtaba Marvasti-Zadeh \and
        Hossein Ghanei-Yakhdan  \and
        Shohreh Kasaei 
}

\institute{S. M. Marvasti-Zadeh \at
              \textit{Digital Image and Video Processing Lab} (DIVPL), Department of Electrical Engineering, Yazd University, Yazd, Iran.\\
              \textit{Image Processing Lab} (IPL), Department of Computer Engineering, Sharif University of Technology, Tehran, Iran.\\
              \textit{Vision and Learning Lab}, Department of Electrical and Computer Engineering, University of Alberta, Edmonton, Canada.
              \\
              \email{mojtaba.marvasti@ualberta.ca}           
           \and
              H. Ghanei-Yakhdan (Corresponding Author) \at
              \textit{Digital Image and Video Processing Lab} (DIVPL), Department of Electrical Engineering, Yazd University, Yazd, Iran.
              \\
              \email{hghaneiy@yazd.ac.ir}
           \and
              S. Kasaei \at
              \textit{Image Processing Lab} (IPL), Department of Computer Engineering, Sharif University of Technology, Tehran, Iran.
              \\
              \email{kasaei@sharif.edu}
}

\date{Received: date / Accepted: date}

\maketitle

\begin{abstract}
Due to the automatic feature extraction procedure via multi-layer nonlinear transformations, the deep learning-based visual trackers have recently achieved a great success in challenging scenarios for visual tracking purposes. Although many of those trackers utilize the feature maps from pre-trained \textit{convolutional neural networks} (CNNs), the effects of selecting different models and exploiting various combinations of their feature maps are still not compared completely. To the best of our knowledge, all those methods use a fixed number of convolutional feature maps without considering the scene attributes (e.g., occlusion, deformation, and fast motion) that might occur during tracking. As a pre-requisition, this paper proposes adaptive \textit{discriminative correlation filters} (DCF) based on the methods that can exploit CNN models with different topologies. First, the paper provides a comprehensive analysis of four commonly used CNN models to determine the best feature maps of each model. Second, with the aid of analysis results as attribute dictionaries, an adaptive exploitation of deep features is proposed to improve the accuracy and robustness of visual trackers regarding video characteristics. Third, the generalization of proposed method is validated on various tracking datasets as well as CNN models with similar architectures. Finally, extensive experimental results demonstrate the effectiveness of proposed adaptive method compared with the state-of-the-art visual tracking methods.
\keywords{Discriminative correlation filters \and deep convolutional neural networks \and robust visual tracking}
\end{abstract}
\section{Introduction}
\label{sec:1_Intro}
\begin{figure}
\centering
\includegraphics[width=0.98\linewidth]{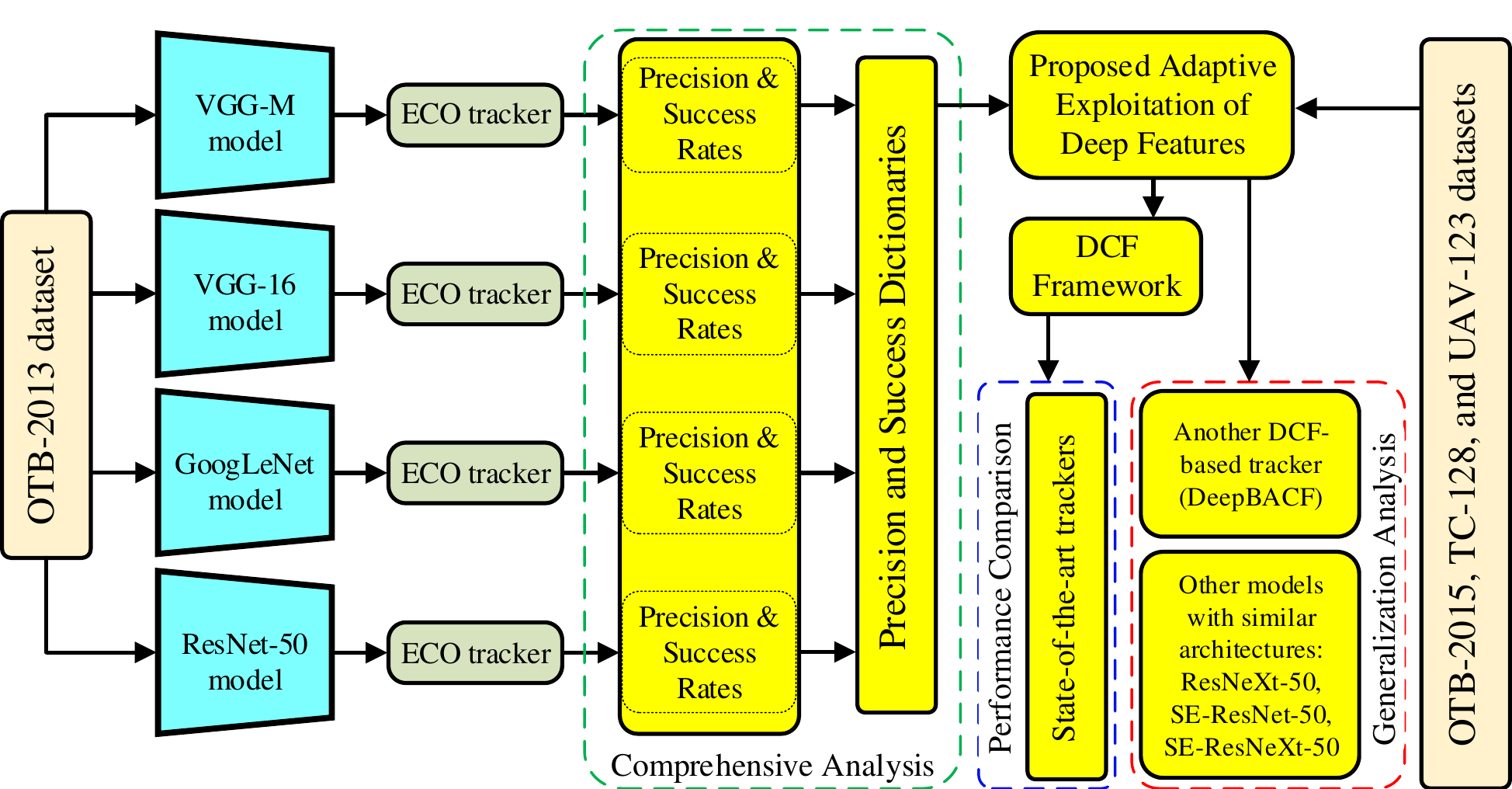}
\caption{A brief overview of this work for adaptive exploitation of deep features in DCF framework.}\label{Overview}
\end{figure}

Generic visual tracking is a fundamental task in computer vision, which aims to estimate the motion trajectory of an unknown target over time \cite{DL-Tracking-Survey,SurveyDeepTracking}. It is included in various practical applications such as automated surveillance and navigation systems, autonomous robots, and self-driving cars \cite{SurveyPedestrianTracking,COMET,Argoverse}. In recent years, \textit{discriminative correlation filters} (DCF) based trackers (e.g., \cite{DSST,ECO,BACF,RADSST}) have achieved great attention considering their robustness to the photometric/geometric variations and significant computational efficiency. The primary purpose of these trackers is to increase the discriminative power of correlation filters to distinguish a target from its background. However, their performance can be dramatically affected by practical scene attributes such severe occlusion, background clutter, deformation, viewpoint change, low resolution, fast camera motion, and heavy illumination variation. \\
\indent It is undeniable that feature extraction is a critical component of visual trackers to meaningfully represent an visual object or a part of it. Besides, the effective selection of features, considering scene information, plays a crucial role in the performance of the DCF-based trackers. Although some visual tracking methods typically use hand-crafted features (e.g., \textit{histogram of oriented gradients} (HOG), \textit{histogram of local intensities} (HOI), \textit{global color histogram} (GCH), and \textit{Color-Names} (CN)), deep features have been successfully employed for visual tracking purposes \cite{DL-Tracking-Survey,SurveyDeepTracking}. To provide unique features of a target and strength the robustness, recent visual tracking methods (e.g., \cite{IMM-DFT,ETDL,DeepTACF,STRCF,ECO,DeepHPFT}) generally exploit fixed number of feature maps extracted from CNNs. However, these trackers have not considered that adaptive utilization of high-dimensional deep features may result in higher learning accuracy (by removing redundant or noisy features), lower computational cost, and better model interpretation. By doing so, deep features can simultaneously provide descriptiveness and flexibility against challenging attributes. \\
\indent Roughly speaking, deep learning-based visual trackers can be categorized into the \textit{feature extraction networks} (FENs) and \textit{end-to-end networks} (EENs) \cite{SurveyDeepTracking}. The FENs are referred to the trackers that employ deep features extracted by pre-trained CNN models into the traditional frameworks such as DCFs. In contrast, the EENs directly evaluate target candidates by the fine-tuned/trained networks on visual tracking datasets. This work will be focused on the exploitation of FENs in the DCF framework. Although most of the recent DCF-based trackers have used deep features, they still utilize various CNN models and different layers of each model. For instance, these trackers widely employ AlexNet \cite{AlexNet} (e.g., \cite{GOTURN,DRT,GDT}) and VGG-Net \cite{VGGM,VGGNet} models, while deeper CNN models are not exhaustively investigated yet. Table~\ref{Table_Models} lists the most popular CNN models, which have been pre-trained on the ImageNet dataset \cite{ImageNet}. Coming to this end, the motivations of this work is to figure out about: 1) the best CNN model as well as the best feature maps of four models for visual tracking purposes, 2) the best combinations of deep features, 3) the robustness of feature maps related to scene attributes, and 4) the adaptive exploitation of best feature maps. Fig.~\ref{Overview} shows a brief overview of this work for DCF-based visual trackers, which employ FENs for feature extraction.\\
\indent The main contributions are as follows. First, as a pre-requisition to exploit various CNN models with different topologies, a modified efficient convolution operators tracker is proposed. Second, a comprehensive analysis of four popular pre-trained CNN models (namely, VGG-M \cite{VGGM}, VGG-16 \cite{VGGNet}, GoogLeNet \cite{GoogLeNet}, and ResNet-50 \cite{ResNet}, which have perceptible differences in terms of error rates) is provided. It ranks the best exploitations of features maps for visual tracking purposes. By the achieved results of the comprehensive analysis, attribute dictionaries are proposed for each model to exploit the best feature maps related to different situations of challenging scenarios. Hence, the first to the third aforementioned motivations are answered by a comprehensive analysis, appropriately. Then, based on the attribute dictionaries of each model, an adaptive exploitation method of deep features is proposed for answering to the fourth motivation. Furthermore, generalization of the proposed method into other DCF-based visual trackers is validated by the aid of the proposed \textit{deep background-aware correlation filters} (DeepBACF) method. Moreover, the generalization of attribute dictionaries is extensively investigated on the pre-trained ResNeXt-50 \cite{ResNeXt}, SE-ResNet-50 \cite{SE-Res-CVPR}, and SE-ResNeXt-50 \cite{SE-Res-CVPR} models, which have similar architectures as the ResNet-50 model. Finally, the performance of the best proposed adaptive method is extensively evaluated with the state-of-the-art trackers on well-known visual tracking datasets. To the best of our knowledge, this is the first work that comprehensively evaluates CNN models and their feature maps for visual tracking purposes. Moreover, this is the first proposed method that investigate adaptive exploitation of different convolutional layers depending on possible challenging attributes of video sequences for visual tracking. \\
\indent The rest of the paper is organized as follows. The overview of related work is described in Section~\ref{sec:2_RelatedWork}. In Section~\ref{sec:3_ProposedMethod}, the four CNN models are comprehensively analyzed, and then the proposed adaptive method for using the best CNN feature maps is presented. Extensive experimental results on visual tracking datasets are given in Section~\ref{sec:4_ExpResults}. Finally, the conclusion and future work are summarized in Section~\ref{sec:5_Conclusion}.
\begin{table}
\caption{Most popular CNN models, number of layers, and corresponding error rates for the classification task.} 
\centering 
\resizebox{\textwidth}{!}{
\begin{tabular}{c c c c c} \hline \hline 
CNN Model & Year & Number of Layers & Top-1 Error & Top-5 Error \\ \hline \hline 
AlexNet \cite{AlexNet} & 2012 & 8 & 41.8 & 19.2  \\ \hline 
VGG-M \cite{VGGM} & 2013 & 8 & 37.1 & 15.8  \\ \hline
VGG-16 (config. D) \cite{VGGNet} & 2014 & 16 & 28.5 & 9.9  \\ \hline 
VGG-19 \cite{VGGNet} & 2014 & 19 & 28.7 & 9.9  \\ \hline 
GoogLeNet \cite{GoogLeNet} & 2014 & 22 & 34.2 & 12.9   \\ \hline 
ResNet-50 \cite{ResNet} & 2015 & 50 & 24.6 & 7.7   \\ 
\hline
\end{tabular}}
\label{Table_Models}
\end{table}
\section{Related Work}
\label{sec:2_RelatedWork}
In this section, the diverse exploitation of CNN models and corresponding layers in recent visual trackers are highlighted. In fact, this brief review of the related work reveals the necessity of comprehensive analysis (Sec.~\ref{sec:3_1_analysis}) to use these CNN models in visual tracking. Related works are classified according to various CNN models. Moreover, the details of the employment of models, layers, and datasets are listed in Table~\ref{Table_Methods_models}.\\
\indent\textbf{VGG-M Model:} 
\textit{Spatially regularized discriminative correlation filters tracker} (DeepSRDCF) \cite{DeepSRDCF} aims to learn more discriminative appearance models on larger search regions. By introducing spatial regularization weights, its formulation penalizes unwanted boundary effects of standard DCF-based methods. To learn a target model in the continuous spatial domain, \textit{continuous convolution operator tracker} (C-COT) \cite{CCOT} employs multi-resolution deep feature maps and an implicit interpolation model for accurate sub-pixel localization of target. Also, \textit{efficient convolution operators tracker} (ECO) \cite{ECO} tackles the computational complexity and over-fitting problem of the C-COT by factorized convolutions, a compact model of training sample distribution, and conservative update strategy. Based on ECO, two trackers \textit{weighted ECO} (WECO) \cite{WECO} and VDSR-SRT \cite{VDSR-SRT} have been proposed. While the WECO tracker introduces a weighted sum operation and feature normalization, the VDSR-SRT tracker addresses the tracking in low-resolution images by a super-resolution algorithm. To exploit temporal information, \textit{spatial-temporal regularized correlation filters tracker} (STRCF) \cite{STRCF} utilizes a temporal regularization term as well as a spatial one to iteratively optimize its filters by the \textit{alternating direction method of multipliers algorithm} (ADMM) \cite{ADMM}. Also, \textit{weighted aggregation with enhancement filter tracker} (WAEF) \cite{WAEF} employs temporal Tikhonov regularization to provide better features and suppress unrelated frames. \textit{Region of interest (ROI) pooled correlation filters tracker} (RPCF) \cite{RPCF} aims to compress model size by utilizing smaller feature maps. Finally, \textit{target-aware correlation filters tracker} (TACF) \cite{DeepTACF} learns guided filters to prevent from background and distractors.
\\
\begin{table}
\caption{Exploited FENS in some visual tracking methods.} 
\centering
\resizebox{\textwidth}{!}{
\begin{tabular}{c c c c}
\hline\hline
Visual Tracking Method & Model & Pre-training Dataset & Name of Exploited Layer(s) \\
\hline \hline 
DeepSRDCF \cite{DeepSRDCF} & VGG-M & ImageNet & Conv1 \\
C-COT \cite{CCOT} & VGG-M & ImageNet & Conv1, Conv5 \\
ECO \cite{ECO} & VGG-M & ImageNet & Conv1, Conv5 \\
WECO \cite{WECO} & VGG-M &	ImageNet &	Conv1, Conv5 \\
VDSR-SRT \cite{VDSR-SRT} & VGG-M &	ImageNet &	Conv1, Conv5 \\
DeepSTRCF \cite{STRCF} & VGG-M & ImageNet & Conv3 \\
WAEF \cite{WAEF} & VGG-M &	ImageNet &	Conv1, Conv5 \\
RPCF \cite{RPCF} & VGG-M &	ImageNet &	Conv1, Conv5 \\
DeepTACF \cite{DeepTACF} & VGG-M &	ImageNet &	Conv1 \\ \hline
ETDL \cite{ETDL} & VGG-16 & ImageNet & Conv1-2 \\
FCNT \cite{FCNT} & VGG-16 & ImageNet & Conv4-3, Conv5-3 \\
DNT \cite{DNT} & VGG-16 & ImageNet & Conv4-3, Conv5-3 \\
CREST \cite{CREST} & VGG-16 & ImageNet &	Conv4-3 \\
CPT \cite{CPT} & VGG-16 & ImageNet & Conv5-1, Conv5-3 \\
DeepFWDCF \cite{DeepFWDCF} & VGG-16 & ImageNet & Conv4-3 \\
DTO \cite{DTO} & VGG-16, SSD &	ImageNet &	Conv3-3, Conv4-3, Conv5-3 \\ \hline
HCFT \cite{HCFT} & VGG-19 & ImageNet & Conv3-4, Conv4-4, Conv5-4 \\
HCFTs \cite{HCFTs} & VGG-19 & ImageNet & Conv3-4, Conv4-4, Conv5-4 \\
LCTdeep \cite{LCTdeep} & VGG-19 & ImageNet & Conv5-4 \\
HDT \cite{HDT} & VGG-19 & ImageNet & Conv4-2, Conv4-3, Conv4-4, Conv5-2, Conv5-3, Conv5-4 \\
IBCCF \cite{IBCCF} & VGG-19 & ImageNet &	Conv3-4, Conv4-4, Conv5-4 \\
DCPF \cite{DCPF} &	VGG-19 & ImageNet &	Conv3-4, Conv4-4, Conv5-4 \\
MCPF \cite{MCPF} &	VGG-19 & ImageNet &	Conv3-4, Conv4-4, Conv5-4 \\
MCCT \cite{MCCT} &	VGG-19 & ImageNet &	Conv4-4, Conv5-4 \\
ORHF \cite{ORHF} &	VGG-19 & ImageNet &	Conv3-4, Conv4-4, Conv5-4 \\
IMM-DFT \cite{IMM-DFT} & VGG-19 & ImageNet & Conv3-4, Conv4-4, Conv5-4 \\ \hline
DeepHPFT \cite{DeepHPFT} & VGG-16, VGG-19, and GoogLeNet &	ImageNet &	Conv5-3, Conv5-4, and icp6-out \\ 
\hline
\end{tabular}
}\label{Table_Methods_models}
\end{table}
\indent\textbf{VGG-16 Models:} 
\textit{Enhanced tracking and detection learning method} (ET -DL) \cite{ETDL} consists of adaptive multi-scale DCFs and a re-detection module to robustly track a target and find it after failures. To separate category detection and distraction determination, \textit{deep fully convolutional networks tracker} (FCNT) \cite{FCNT} uses distinct convolutional layers and a feature map selection method. Thereby, the computational burden can be reduced, and irrelevant features can be discarded. \textit{Dual network-based tracker} (DNT) \cite{DNT} embeds boundary and shape information into deep features to enjoy more effective features for visual tracking. Also, CREST tracker \cite{CREST} integrates the processes of learning DCFs with feature extraction to provide more appropriate features for visual tracking. Moreover, \textit{adaptive feature weighted DCF tracker} (FWDCF) \cite{DeepFWDCF} weights deep features by a segmentation model to suppress the background and distractors. To adaptively leverage low-dimensional features, \textit{channel pruning tracker} (CPT) \cite{CPT} provides a channel pruned VGG-16 model, average feature energy ratio method, and adaptive iterative strategy for target localization. In contrast to mentioned trackers, \textit{deep tracking with objectness method} \cite{DTO} assumes that the tracker is aware of object categories to investigate its effect on tracking performance.\\
\indent\textbf{VGG-19 Models:} 
\textit{Hierarchical correlation feature-based tracker} (HCFT) \cite{HCFT} learns multi-level correlation response maps on multiple convolutional layers to simultaneously alleviate appearance variation and precisely localize the target. Also, the modified HCFT (called HCFTs or HCFT*) \cite{HCFTs} partially compares the performance of three CNN models (i.e., AlexNet, VGG-19, and ResNet-152) and adds two region proposals and a classifier to HCFT for long-term visual tracking purposes. However, insufficient exploration of ResNet’s feature maps results in imperfect visual tracking. \textit{Deep long-term correlation tracker} (LCTdeep) \cite{LCTdeep} consists of distinctive DCFs, pyramidal features, short-term \& and long-term learning rates, and an incrementally learned detector to improve the tracking robustness in presence of significant appearance change and scale variation. By using an online decision-theoretical Hedge algorithm, \textit{hedged deep tracker} (HDT) \cite{HDT} aggregates weak CNN-based trackers for exploring advantages of hierarchical feature maps. To handle aspect ratio variation, \textit{1D Boundary and 2D Center CFs tracker} (IBCCF) provides a family of boundary CFs and optimizes the boundary and center correlation filters. By exploiting particle filters, the DCPF \cite{DCPF} tracker strengthens deep features to discriminate the target from its background. \textit{Multi-task correlation particle filter tracker} (MCPF) considers inter-dependencies among deep features to cover multiple modes in the posterior density of the target state. Besides, DeepHPFT tracker \cite{DeepHPFT} exploits hand-crafted and deep features in particle filter framework to improve the visual tracking performance. To decide based on reliable localization, MCCT tracker \cite{MCCT} constructs various DCFs that employ different features to learn different target models. To preserve computational complexity, ORHF tracker \cite{ORHF} validates the estimated confidence scores and selects effective deep features. Lastly, IMM-DFT tracker \cite{IMM-DFT} considers insufficiency of linear combination of deep features and provides adaptive hierarchical features for visual tracking. \\
\indent In contrast to existing FEN-based visual trackers, this work reviews all possible exploration of four widely used CNN models for visual tracking. By doing a comprehensive analysis, two attribute dictionaries for the CNN models are provided. These dictionaries do not follow concrete rules to employ into visual trackers. Thus, the dictionaries are considered as the keys to effectively select the most appropriate features regarding video characteristics or an estimation of them. To validate the analysis results, the generalization of the dictionaries are assessed on different visual tracking datasets, CNN models with similar architectures, and another DCF-based tracker. Owing to the analyses and dictionaries, effective adaptive exploitation of deep features will be possible. Therefore, an adaptive method is proposed, which can simply integrate into the DCF-based trackers to improve discrimination ability of target modeling.
\section{Proposed Visual Tracking Method}
\label{sec:3_ProposedMethod}
In this section, the architecture of the four most popular CNN models and the reason for choosing them in the comprehensive analysis is briefly mentioned. Then, the comprehensive analysis and main results of the CNN models are presented. By this analysis, the best features of each model related to the challenging attributes are provided. Finally, a method for adaptive exploitation of these models is proposed.
\subsection{Comprehensive Analysis of Pre-trained CNN Models}
\label{sec:3_1_analysis}
As shown in Table~\ref{Table_Models}, the most popular CNN models have just a slight difference in performance (e.g., the VGG-16 and VGG-19). As such, in this work, the model with lower complexity is selected. Also, the AlexNet has considerably less performance than others. Thus, four commonly used CNN models, namely VGG-M, VGG-16, GoogLeNet, and ResNet-50 are selected (for more details, see \cite{VGGM,VGGNet,GoogLeNet,ResNet}). Table~\ref{NetConfigs} shows the the configuration of models and test layers (denoted by D1 to D5) in this work. Regarding the topologies of these models, the architectures include either a simple multi-layer stack of non-linear layers (i.e., VGG-M and VGG-Net) or a directed acyclic graph topology (i.e., GoogLeNet and ResNet), which allows designing more complex designs with multiple inputs/outputs for layers. As a pre-requisition to compare the models, this paper proposes a modified ECO tracker which can exploit different CNN models using advanced deep learning modules (i.e., modifying the feature extraction process of ECO tracker by employing AutoNN, and McnExtraLayers modules in MatConvNet toolbox \cite{MatConvNet}). The ECO tracker \cite{ECO} fuses CNN and hand-crafted features, while it reduces the dimension of deep features by the \textit{principal component analysis} (PCA) and a down-sampling strategy. However, in order to have fair and meaningful comparisons, the modified ECO tracker (Sec.~\ref{sec:3_2_analysis_results}) and the proposed adaptive method (Sec.~\ref{sec:AdaptiveExplo}) do not fuse CNN features with hand-crafted ones and also do not apply down-sampling or dimensional reduction processes.\\
\begin{table}
\caption{Configuration of pre-trained CNN models. [Convolutional layers are denoted as “Conv$<$filter size$>$-$<$filter depth$>$”.]} 
\centering 
\resizebox{\textwidth}{!}{
\begin{tabular}{c|c|c|c|c|c|c|c|c} \hline \hline
\multicolumn{2}{c|}{VGG-Medium (8-layers)} & \multicolumn{2}{c|}{VGG-Net (16-layers, configuration D)} & \multicolumn{2}{c|}{GoogLeNet (22-layers)} & \multicolumn{3}{c}{ResNet-50 (50-layers)} \\ \hline 
Test Output & Layers & Test Output & Layers & Test Output & Layers & Test Output & \multicolumn{2}{c}{Layers} \\ \hline \hline
\multirow{2}{*}{D1} & \multirow{2}{*}{Conv7-96} & \multicolumn{2}{c|}{\hspace{2cm}Conv3-64} & \multirow{2}{*}{D1} & \multirow{2}{*}{Conv7-64} & \multirow{2}{*}{D1} & \multicolumn{2}{c}{\multirow{2}{*}{Conv7-64}} \\ \cline{3-4}
 &  & D1 & Conv3-64 &  &  &  & \multicolumn{2}{c}{} \\ \hline 
\multicolumn{2}{c|}{\hspace{2cm}LRN} & \multicolumn{2}{c|}{\hspace{2.1cm}max pool} & \multicolumn{2}{c|}{\hspace{2.1cm}max pool} & \multicolumn{3}{c}{\multirow{2}{*}{\hspace{2.1cm}max pool}} \\ \cline{1-6}
\multicolumn{2}{c|}{\multirow{2}{*}{\hspace{2.1cm}max pool}} & \multicolumn{2}{c|}{\hspace{2cm}Conv3-128} & \multirow{2}{*}{D2} & \multirow{2}{*}{Conv3-192} \\ \cline{3-4}\cline{7-9}
\multicolumn{2}{c|}{} & D2 & Conv3-128 &  & & \multicolumn{2}{c|}{\hspace{2.1cm}Conv1-64} & \multirow{3}{*}{$\times3$} \\ \cline{1-6}
\multirow{2}{*}{D2} & \multirow{2}{*}{Conv5-256} & \multicolumn{2}{c|}{\hspace{2.1cm}max pool} & \multicolumn{2}{c|}{\hspace{2.1cm}max pool} & \multicolumn{2}{c|}{\hspace{2.1cm}Conv3-64} &  \\ \cline{3-8}
 &  & \multicolumn{2}{c|}{} & \multicolumn{2}{c|}{\hspace{2.1cm}Inception (3a)} & D2 & Conv1-256 &  \\ \cline{1-2}\cline{5-9}
\multicolumn{2}{c|}{\multirow{2}{*}{\hspace{2.1cm}max pool}} & \multicolumn{2}{c|}{\hspace{2.1cm}Conv3-256} & D3 & Inception (3b) & \multicolumn{2}{c|}{\hspace{2.1cm}Conv1-128} & \multirow{4}{*}{$\times4$} \\ \cline{5-6}
\multicolumn{2}{c|}{} & \multicolumn{2}{c|}{\hspace{2.1cm}Conv3-256} & \multicolumn{2}{c|}{\hspace{2.1cm}max pool} & \multicolumn{2}{c|}{\hspace{2.1cm}Conv3-128} &  \\ \cline{1-8}
\multicolumn{2}{c|}{\hspace{2.1cm}Conv3-512} & D3 & Conv3-256 & \multicolumn{2}{c|}{} & \multirow{2}{*}{D3} & \multirow{2}{*}{Conv1-512} &  \\ \cline{3-4} 
\multicolumn{2}{c|}{\hspace{2.1cm}Conv3-512} & \multicolumn{2}{c|}{\hspace{2.1cm}max pool} & \multicolumn{2}{c|}{} &  &  &  \\ \cline{1-4}\cline{7-9}
\multirow{2}{*}{D3} & \multirow{2}{*}{Conv3-512} & \multicolumn{2}{c|}{\hspace{2.1cm}Conv3-512} & \multicolumn{2}{c|}{\hspace{2.1cm}Inception (4a)} & \multicolumn{2}{c|}{\hspace{2.1cm}Conv1-256} & \multirow{4}{*}{$\times6$} \\ 
 &  & \multicolumn{2}{c|}{\hspace{2.1cm}Conv3-512} & \multicolumn{2}{c|}{\hspace{2.1cm}Inception (4b)} & \multicolumn{2}{c|}{\hspace{2.1cm}Conv3-256} &  \\ \cline{1-4} \cline{7-8}
\multicolumn{2}{c|}{\multirow{3}{*}{\hspace{2.1cm}max pool}} & \multirow{2}{*}{D4} & \multirow{2}{*}{Conv3-512} & \multicolumn{2}{c|}{\hspace{2.1cm}Inception (4c)} & \multirow{2}{*}{D4} & \multirow{2}{*}{Conv1-1024} &  \\ 
\multicolumn{2}{c|}{} & & & \multicolumn{2}{c|}{\hspace{2.1cm}Inception (4d)} &  & & \\ \cline{3-4}\cline{7-9}
\multicolumn{2}{c|}{} & \multicolumn{2}{c|}{\hspace{2.1cm}max pool} & \multicolumn{2}{c|}{} & \multicolumn{2}{c|}{\hspace{2.1cm}Conv1-512} & \multirow{4}{*}{$\times3$} \\ \cline{1-6}
\multicolumn{2}{c|}{\multirow{3}{*}{\hspace{2.1cm}FC-4096}} & \multicolumn{2}{c|}{} & D4 & Inception (4e) & \multicolumn{2}{c|}{\hspace{2.1cm}Conv3-512} &  \\ \cline{5-8}
\multicolumn{2}{c|}{} & \multicolumn{2}{c|}{\hspace{2.1cm}Conv3-512} & \multicolumn{2}{c|}{\hspace{2.1cm}max pool} & \multirow{2}{*}{D5} & \multirow{2}{*}{Conv1-2048} &  \\ \cline{5-6} 
\multicolumn{2}{c|}{} & \multicolumn{2}{c|}{\hspace{2.1cm}Conv3-512} & \multicolumn{2}{c|}{\hspace{2.1cm}Inception (5a)} &  &  &  \\ \cline{1-2}\cline{5-9}
\multicolumn{2}{c|}{\multirow{3}{*}{\hspace{2.1cm}FC-4096}} & \multicolumn{2}{c|}{} & D5 & Inception (5b) & \multicolumn{3}{c}{\multirow{3}{*}{\hspace{2.1cm}average pool}} \\ \cline{3-6}
\multicolumn{2}{c|}{} & D5 & Conv3-512 & \multicolumn{2}{c|}{\multirow{2}{*}{\hspace{2.1cm}average pool}} & \multicolumn{3}{c}{} \\ \cline{3-4}
\multicolumn{2}{c|}{} & \multicolumn{2}{c|}{\hspace{2.1cm}max pool} & \multicolumn{2}{c|}{} & \multicolumn{3}{c}{} \\ \hline 
\multicolumn{2}{c|}{\multirow{2}{*}{\hspace{2.1cm}FC-1000}} & \multicolumn{2}{c|}{\hspace{2.1cm}FC-4096} & \multicolumn{2}{c|}{\hspace{2.1cm}Dropout (40\%)} & \multicolumn{3}{c}{\multirow{2}{*}{\hspace{2.1cm}FC-1000}} \\ \cline{3-6} 
\multicolumn{2}{c|}{} & \multicolumn{2}{c|}{\hspace{2.1cm}FC-4096} & \multicolumn{2}{c|}{\hspace{2.1cm}FC-1000} & \multicolumn{3}{c}{} \\ \hline 
\multicolumn{2}{c|}{\multirow{2}{*}{\hspace{2.1cm}Soft-max}} & \multicolumn{2}{c|}{\hspace{2.1cm}FC-1000} & \multicolumn{2}{c|}{\multirow{2}{*}{\hspace{2.1cm}Soft-max}} & \multicolumn{3}{c}{\multirow{2}{*}{\hspace{2.1cm}Soft-max}} \\ \cline{3-4} 
\multicolumn{2}{c|}{} & \multicolumn{2}{c|}{\hspace{2.1cm}Soft-max} & \multicolumn{2}{c|}{} & \multicolumn{3}{c}{} \\ \hline 
\end{tabular}}
\label{NetConfigs}
\end{table}
\indent The comprehensive analyses of convolutional layers for each model are listed in Table~\ref{SuccessDic} and Table~\ref{PrecisionDic}. To improve the evaluation speeds, all subsequent layers after the last test layers (i.e., the D3 or D5 output) are removed. In contrast to other works, all of the single layers and also all possible combinations of layers of the models are investigated in this paper. Also, the best (First to third) and the worst feature maps for visual tracking purposes are ranked in these tables. Furthermore, the challenging attributes are categorized based on the factors related to object, camera, and environment. This categorization facilitates exploring these results for the proposed adaptive method (Sec.~\ref{sec:AdaptiveExplo}).\\
\indent In this paper, all evaluations are based on the well-known precision and success metrics. The overlap success metric is the percentage of frames that their overlap score of estimated and ground-truth bounding boxes is more than a specific threshold while the distance precision metric is defined as the percentage of frames that their estimated location error with the ground-truth location is smaller than a particular threshold. Note that the default thresholds of standard benchmarks (i.e., 50\% overlap and 20 pixels) are used for the evaluations \cite{OTB2015}. The success and precision results of the comprehensive analyses of the CNN models on the OTB-2013 dataset \cite{OTB2013}, their resolution, and the number of feature maps are listed in Table~\ref{SuccessDic} and Table~\ref{PrecisionDic}. These tables present the results on more than 29000 frames for each evaluated case of CNN models. In this work, visual tracking datasets \cite{OTB2013,OTB2015,TC128} have common challenging attributes including \textit{illumination variation} (IV), \textit{out-of-plane rotation} (OPR), \textit{scale variation} (SV), \textit{occlusion} (OCC), \textit{deformation} (DEF), \textit{motion blur} (MB), \textit{fast motion} (FM), \textit{in-plane rotation} (IPR), \textit{out-of-view} (OV), \textit{background clutter} (BC), and \textit{low resolution} (LR). Note that due to extensive evaluations, only the best and worst CNN layers are presented (see Appendix~\ref{appendix} for more details). These tables provide an excellent perspective to use these CNN models and their features for visual tracking. Besides, the pros and cons of the models, layers, and combinations regarding each attribute are clarified. These analyses encourage other visual trackers to propose more sophisticated adaptive methods but also employ precise CNN layers for improving their performance in the presence of specific destructive attributes.
\begin{table}
\caption{Success analysis results for pre-trained CNN models on OTB-2013 dataset. [The first to third best layers and the worst layer in each case are shown with green, blue, yellow, and red color, respectively. The multi-resolution feature maps are abbreviated by “MR”.]} 
\centering 
\resizebox{\textwidth}{!}{
\begin{tabular}{c |c |c |c |c |c |c |c |c |c |c |c |c |c |c} \hline\hline
\multirow{3}{*}{Model} & \multirow{3}{*}{Layers} & \multirow{3}{*}{Features: Resolution/Depth} & \multirow{3}{*}{Overall} & \multicolumn{11}{c}{Attributes } \\ \cline{5-15}
 &  &  &  & \multicolumn{4}{c|}{Object} & \multicolumn{4}{c|}{Camera} & \multicolumn{3}{c}{Environment} \\ \cline{5-15}
 &  &  &  & SV & DEF  & OPR & IPR & FM & MB & LR & OV & BC & OCC & IV \\ \hline \hline 
\multirow{7}{*}{VGG-M} & D1 & 109x109 / 96 & 0.797 & 0.727 & 0.831 & 0.759 & 0.713 & 0.741 & 0.754 & 0.598 & 0.840 & 0.719 & 0.807 & \cellcolor{yellow}0.752 \\ \cline{2-15}
 & D2 & 26x26 / 256 & \cellcolor{green}0.827 & \cellcolor{yellow}0.752 & \cellcolor{green}0.891 & \cellcolor{green}0.798 & \cellcolor{yellow}0.727 & 0.787 & 0.799 & 0.659 & 0.901 & \cellcolor{green}0.752 & \cellcolor{blue!25}0.856 & \cellcolor{yellow}0.752 \\ \cline{2-15}
 & D3 & 13x13 / 512 & \cellcolor{red!50}0.661 & \cellcolor{red!50}0.645 & \cellcolor{red!50}0.643 & \cellcolor{red!50}0.644 & \cellcolor{red!50}0.610 & \cellcolor{red!50}0.639 & \cellcolor{red!50}0.687 & \cellcolor{red!50}0.383 & \cellcolor{red!50}0.599 & \cellcolor{red!50}0.620 & \cellcolor{red!50}0.700 & \cellcolor{red!50}0.591 \\ \cline{2-15}
 & D1, D2 & MR / 352 & 0.802 & 0.737 & \cellcolor{blue!25}0.879 & 0.774 & 0.697 & \cellcolor{yellow}0.797 & \cellcolor{blue!25}0.824 & \cellcolor{blue!25}0.694 & 0.858 & 0.728 & \cellcolor{yellow}0.844 & \cellcolor{blue!25}0.763 \\ \cline{2-15}
 & D1, D3 & MR / 608 & 0.813 & \cellcolor{blue!25}0.756 & 0.841 & 0.780 & \cellcolor{green}0.732 & \cellcolor{green}0.801 & \cellcolor{green}0.829 & \cellcolor{green}0.701 & \cellcolor{green}0.931 & \cellcolor{blue!25}0.749 & 0.840 & \cellcolor{green}0.783 \\ \cline{2-15}
 & D2, D3 & MR / 768 & \cellcolor{yellow}0.816 & \cellcolor{green}0.766 & 0.826 & \cellcolor{yellow}0.784 & 0.718 & \cellcolor{blue!25}0.799 & \cellcolor{yellow}0.807 & \cellcolor{yellow}0.672 & \cellcolor{blue!25}0.913 & \cellcolor{yellow}0.746 & 0.831 & 0.742 \\ \cline{2-15}
 & D1, D2, D3 & MR / 864 & \cellcolor{blue!25}0.823 & 0.751 & \cellcolor{yellow}0.866 & \cellcolor{blue!25}0.793 & \cellcolor{blue!25}0.729 & 0.787 & 0.806 & 0.660 & \cellcolor{yellow}0.905 & 0.736 & \cellcolor{green}0.871 & 0.738 \\ \hline\hline 
\multirow{12}{*}{VGG-16} & D4 & 28x28 / 512 & \cellcolor{green}0.849 & \cellcolor{blue!25}0.792 & \cellcolor{green}0.893 & \cellcolor{green}0.836 & \cellcolor{blue!25}0.778 & \cellcolor{green}0.822 & \cellcolor{green}0.818 & 0.671 & \cellcolor{green}0.924 & \cellcolor{green}0.775 & \cellcolor{green}0.889 & \cellcolor{green}0.772 \\ \cline{2-15}
 & D5 & 14x14 / 512 & \cellcolor{red!50}0.649 & \cellcolor{red!50}0.637 & \cellcolor{red!50}0.609 & \cellcolor{red!50}0.639 & 0.619 & \cellcolor{red!50}0.653 & \cellcolor{red!50}0.635 & \cellcolor{red!50}0.398 & \cellcolor{red!50}0.609 & \cellcolor{red!50}0.542 & \cellcolor{red!50}0.647 & 0.668 \\ \cline{2-15}
 & D1, D2 & MR / 192 & 0.754 & 0.679 & 0.789 & 0.705 & 0.647 & 0.685 & 0.680 & 0.560 & 0.743 & 0.652 & 0.779 & \cellcolor{red!50}0.663 \\ \cline{2-15}
 & D2, D5 & MR / 640 & 0.741 & 0.700 & 0.747 & 0.691 & \cellcolor{red!50}0.612 & 0.688 & 0.685 & 0.591 & 0.734 & 0.678 & 0.756 & 0.677 \\ \cline{2-15}
 & D3, D4 & MR / 768 & 0.801 & 0.730 & \cellcolor{yellow}0.854 & 0.764 & 0.723 & 0.739 & 0.795 & 0.710 & 0.775 & \cellcolor{yellow}0.744 & 0.818 & 0.715 \\ \cline{2-15}
 & D3, D5 & MR / 768 & 0.791 & 0.739 & 0.824 & 0.755 & 0.689 & 0.760 & 0.769 & 0.673 & 0.824 & 0.716 & 0.830 & 0.731 \\ \cline{2-15}
 & D4, D5 & MR / 1024 & \cellcolor{blue!25}0.833 & \cellcolor{green}0.793 & 0.846 & \cellcolor{blue!25}0.816 & \cellcolor{green}0.779 & 0.822 & \cellcolor{blue!25}0.807 & 0.644 & \cellcolor{blue!25}0.913 & 0.731 & \cellcolor{yellow}0.869 & 0.739 \\ \cline{2-15}
 & D1, D4, D5 & MR / 1078 & \cellcolor{yellow}0.825 & \cellcolor{yellow}0.766 & 0.845 & \cellcolor{yellow}0.796 & 0.733 & \cellcolor{blue!25}0.805 & \cellcolor{yellow}0.797 & 0.704 & \cellcolor{yellow}0.894 & \cellcolor{blue!25}0.752 & \cellcolor{blue!25}0.886 & \cellcolor{yellow}0.762 \\ \cline{2-15}
 & D2, D3, D4 & MR / 896 & 0.798 & 0.747 & 0.836 & 0.760 & 0.706 & 0.772 & 0.778 & \cellcolor{blue!25}0.713 & 0.886 & 0.707 & 0.830 & 0.730 \\ \cline{2-15}
 & D3, D4, D5 & MR / 1280 & 0.821 & 0.755 & \cellcolor{blue!25}0.856 & 0.790 & \cellcolor{yellow}0.735 & 0.770 & 0.770 & \cellcolor{blue!25}0.713 & 0.875 & 0.743 & 0.856 & 0.747 \\ \cline{2-15}
 & D2, D3, D4, D5 & MR / 1408 & 0.804 & 0.759 & 0.803 & 0.756 & 0.686 & \cellcolor{yellow}0.790 & 0.791 & \cellcolor{yellow}0.711 & 0.878 & 0.741 & 0.830 & \cellcolor{blue!25}0.763 \\ \cline{2-15}
 & D1, D2, D3, D4, D5 & MR / 1472 & 0.792 & 0.757 & 0.795 & 0.752 & 0.689 & 0.788 & 0.788 & \cellcolor{green}0.714 & 0.871 & 0.735 & 0.821 & 0.757 \\ \hline \hline 
\multirow{14}{*}{GoogLeNet} & D3 & 28x28 / 256 & 0.818 & 0.767 & \cellcolor{green}0.879 & 0.786 & 0.711 & 0.762 & 0.756 & 0.526 & 0.778 & 0.715 & 0.843 & 0.732 \\ \cline{2-15}
 & D4 & 14x14 / 528 & 0.774 & 0.730 & 0.875 & 0.779 & 0.704 & 0.732 & 0.691 & 0.519 & 0.799 & \cellcolor{blue!25}0.785 & 0.831 & 0.726 \\ \cline{2-15}
 & D5 & 7x7 / 832 & \cellcolor{red!50}0.395 & \cellcolor{red!50}0.333 & \cellcolor{red!50}0.332 & \cellcolor{red!50}0.395 & \cellcolor{red!50}0.419 & \cellcolor{red!50}0.398 & \cellcolor{red!50}0.362 & \cellcolor{red!50}0.299 & \cellcolor{red!50}0.354 & \cellcolor{red!50}0.404 & \cellcolor{red!50}0.373 & \cellcolor{red!50}0.422 \\ \cline{2-15}
 & D2, D3 & MR / 448 & \cellcolor{yellow}0.822 & 0.764 & 0.865 & \cellcolor{yellow}0.792 & 0.726 & \cellcolor{blue!25}0.785 & \cellcolor{green}0.811 & 0.705 & \cellcolor{green}0.889 & 0.761 & \cellcolor{yellow}0.856 & 0.752 \\ \cline{2-15}
 & D2, D4 & MR / 720 & 0.811 & 0.745 & 0.864 & 0.778 & 0.710 & 0.764 & 0.785 & 0.702 & \cellcolor{yellow}0.881 & 0.748 & 0.839 & 0.737 \\ \cline{2-15}
 & D3, D4 & MR / 784 & \cellcolor{yellow}0.822 & 0.765 & 0.875 & \cellcolor{yellow}0.792 & \cellcolor{yellow}0.746 & 0.744 & 0.791 & 0.701 & 0.774 & 0.762 & 0.855 & 0.716 \\ \cline{2-15}
 & D3, D5 & MR / 1088 & 0.820 & 0.770 & 0.876 & 0.789 & 0.718 & 0.761 & 0.737 & 0.539 & 0.790 & 0.712 & 0.845 & 0.730 \\ \cline{2-15}
 & D4, D5 & MR / 1360 & 0.791 & 0.759 & \cellcolor{blue!25}0.877 & 0.791 & \cellcolor{green}0.760 & 0.759 & 0.713 & 0.621 & 0.758 & \cellcolor{green}0.858 & 0.809 & \cellcolor{green}0.781 \\ \cline{2-15}
 & D1, D2, D3 & MR / 512 & 0.819 & 0.760 & 0.866 & 0.788 & 0.720 & \cellcolor{yellow}0.784 & \cellcolor{yellow}0.804 & 0.693 & 0.874 & 0.761 & 0.854 & \cellcolor{blue!25}0.754 \\ \cline{2-15} 
 & D1, D2, D4 & MR / 784 & 0.801 & 0.747 & 0.860 & 0.764 & 0.694 & 0.752 & 0.773 & \cellcolor{yellow}0.705 & 0.873 & 0.723 & 0.837 & 0.737 \\ \cline{2-15}
 & D1, D3, D4 & MR / 848 & 0.774 & 0.750 & 0.768 & 0.730 & 0.660 & 0.763 & 0.778 & \cellcolor{blue!25}0.707 & 0.840 & 0.725 & 0.809 & 0.716 \\ \cline{2-15} 
 & D2, D3, D4 & MR / 976 & 0.798 & 0.754 & 0.816 & 0.761 & 0.686 & 0.769 & 0.785 & \cellcolor{green}0.711 & \cellcolor{blue!25}0.885 & 0.753 & 0.812 & 0.742 \\ \cline{2-15}
 & D2, D3, D5 & MR / 1280 & \cellcolor{blue!25}0.827 & \cellcolor{blue!25}0.771 & 0.870 & \cellcolor{blue!25}0.798 & 0.731 & 0.784 & \cellcolor{blue!25}0.807 & 0.693 & 0.878 & 0.764 & \cellcolor{blue!25}0.869 & \cellcolor{yellow}0.753 \\ \cline{2-15}
 & D3, D4, D5 & MR / 1616 & \cellcolor{green}0.840 & \cellcolor{green}0.793 & 0.876 & \cellcolor{green}0.815 & \cellcolor{blue!25}0.752 & \cellcolor{green}0.789 & 0.798 & 0.676 & 0.870 & \cellcolor{yellow}0.766 & \cellcolor{green}0.881 & 0.745 \\ \hline\hline 
\multirow{11}{*}{ResNet-50} & D1 & 112x112 / 64 & 0.789 & 0.741 & 0.799 & 0.759 & \cellcolor{yellow}0.719 & 0.704 & 0.707 & 0.597 & 0.707 & 0.691 & 0.787 & 0.725 \\ \cline{2-15}
 & D3 & 28x28 / 512 & \cellcolor{blue!25}0.825 & 0.760 & \cellcolor{blue!25}0.874 & \cellcolor{blue!25}0.796 & \cellcolor{blue!25}0.726 & 0.807 & \cellcolor{yellow}0.823 & 0.696 & \cellcolor{yellow}0.918 & \cellcolor{green}0.775 & \cellcolor{green}0.871 & \cellcolor{blue!25}0.766 \\ \cline{2-15} 
 & D4 & 14x14 / 1024 & 0.734 & 0.709 & 0.767 & 0.736 & 0.681 & 0.732 & 0.696 & 0.389 & 0.711 & 0.700 & 0.761 & 0.605 \\ \cline{2-15} 
 & D5 & 7x7 / 2048 & \cellcolor{red!50}0.478 & \cellcolor{red!50}0.453 & \cellcolor{red!50}0.380 & \cellcolor{red!50}0.502 & \cellcolor{red!50}0.500 & \cellcolor{red!50}0.423 & \cellcolor{red!50}0.451 & \cellcolor{red!50}0.178 & \cellcolor{red!50}0.356 & \cellcolor{red!50}0.322 & \cellcolor{red!50}0.465 & \cellcolor{red!50}0.432 \\ \cline{2-15}
 & D1, D3 & MR / 576 & 0.811 & \cellcolor{yellow}0.765 & 0.858 & 0.778 & 0.712 & 0.765 & 0.783 & 0.676 & 0.816 & 0.727 & \cellcolor{yellow}0.856 & 0.746 \\ \cline{2-15} 
 & D3, D4 & MR / 1536 & 0.811 & \cellcolor{blue!25}0.766 & 0.833 & 0.777 & 0.706 & \cellcolor{blue!25}0.818 & 0.822 & 0.691 & 0.904 & \cellcolor{yellow}0.764 & 0.839 & \cellcolor{blue!25}0.766 \\ \cline{2-15}
 & D3, D5 & MR / 2560 & \cellcolor{green}0.829 & \cellcolor{blue!25}0.766 & \cellcolor{green}0.883 & \cellcolor{green}0.801 & \cellcolor{green}0.735 & \cellcolor{yellow}0.816 & \cellcolor{blue!25}0.830 & \cellcolor{yellow}0.704 & \cellcolor{blue!25}0.930 & 0.762 & \cellcolor{green}0.871 & \cellcolor{yellow}0.763 \\ \cline{2-15}
 & D1, D3, D4 & MR / 1600 & \cellcolor{yellow}0.815 & 0.762 & \cellcolor{yellow}0.864 & \cellcolor{yellow}0.783 & 0.717 & 0.777 & 0.771 & 0.688 & 0.848 & 0.723 & \cellcolor{blue!25}0.864 & 0.751 \\ \cline{2-15} 
 & D1, D3, D5 & MR / 2624 & 0.814 & 0.756 & 0.862 & 0.782 & 0.716 & 0.781 & 0.789 & \cellcolor{blue!25}0.708 & 0.858 & 0.733 & \cellcolor{blue!25}0.864 & 0.754 \\ \cline{2-15}
 & D3, D4, D5 & MR / 3584 & \cellcolor{yellow}0.815 & \cellcolor{green}0.769 & 0.837 & 0.782 & 0.710 & \cellcolor{green}0.829 & \cellcolor{green}0.843 & \cellcolor{blue!25}0.708 & \cellcolor{green}0.940 & \cellcolor{blue!25}0.772 & 0.847 & \cellcolor{green}0.777 \\ \cline{2-15}
 & D1, D3, D4, D5 & MR / 3648 & 0.810 & 0.754 & 0.850 & 0.777 & \cellcolor{yellow}0.719 & 0.780 & 0.777 & \cellcolor{green}0.714 & 0.860 & 0.707 & \cellcolor{yellow}0.856 & 0.742 \\ \hline
\end{tabular}
}
\label{SuccessDic}
\end{table}
\begin{table}
\caption{Precision analysis results for pre-trained CNN models on OTB-2013 dataset. [The first to third best layers and the worst layer in each case are shown with green, blue, yellow, and red color, respectively. The multi-resolution feature maps are abbreviated by “MR”.]} 
\centering 
\resizebox{\textwidth}{!}{
\begin{tabular}{c |c |c |c |c |c |c |c |c |c |c |c |c |c |c} \hline \hline 
\multirow{3}{*}{Model} & \multirow{3}{*}{Layers} & \multirow{3}{*}{Features: Resolution/Depth} & \multirow{3}{*}{Overall} & \multicolumn{11}{c}{Attributes } \\ \cline{5-15}
 &  &  &  & \multicolumn{4}{c|}{Object} & \multicolumn{4}{c|}{Camera} & \multicolumn{3}{c}{Environment} \\ \cline{5-15}
 &  &  &  & SV & DEF & OPR & IPR & FM & MB & LR & OV & BC & OCC & IV \\ \hline \hline 
\multirow{7}{*}{VGG-M} & D1 & 109x109 / 96 & \cellcolor{yellow}0.895 & \cellcolor{yellow}0.860 & 0.903 & \cellcolor{yellow}0.891 & \cellcolor{blue!25}0.846 & 0.804 & \cellcolor{yellow}0.808 & 0.594 & 0.848 & \cellcolor{yellow}0.815 & 0.898 & \cellcolor{yellow}0.837 \\ \cline{2-15}
 & D2 & 26x26 / 256 & 0.884 & 0.841 & \cellcolor{blue!25}0.916 & 0.874 & 0.814 & \cellcolor{yellow}0.816 & 0.783 & \cellcolor{yellow}0.673 & 0.894 & 0.785 & 0.913 & 0.812 \\ \cline{2-15}
 & D3 & 13x13 / 512 & \cellcolor{red!50}0.752 & \cellcolor{red!50}0.738 & \cellcolor{red!50}0.724 & \cellcolor{red!50}0.764 & \cellcolor{red!50}0.734 & \cellcolor{red!50}0.677 & \cellcolor{red!50}0.731 & \cellcolor{red!50}0.377 & \cellcolor{red!50}0.561 & \cellcolor{red!50}0.697 & \cellcolor{red!50}0.762 & \cellcolor{red!50}0.705 \\ \cline{2-15}
 & D1, D2 & MR / 352 & \cellcolor{blue!25}0.897 & \cellcolor{blue!25}0.874 & \cellcolor{green}0.919 & \cellcolor{blue!25}0.901 & \cellcolor{yellow}0.837 & \cellcolor{green}0.855 & \cellcolor{blue!25}0.847 & \cellcolor{blue!25}0.702 & 0.860 & \cellcolor{blue!25}0.827 & \cellcolor{green}0.939 & \cellcolor{blue!25}0.845 \\ \cline{2-15}
 & D1, D3 & MR / 608 & \cellcolor{green}0.906 & \cellcolor{green}0.884 & \cellcolor{yellow}0.905 & \cellcolor{green}0.905 & \cellcolor{green}0.862 & \cellcolor{green}0.855 & \cellcolor{green}0.855 & \cellcolor{green}0.711 & \cellcolor{green}0.927 & \cellcolor{green}0.850 & \cellcolor{blue!25}0.928 & \cellcolor{green}0.872 \\ \cline{2-15}
 & D2, D3 & MR / 768 & 0.870 & 0.848 & 0.871 & 0.856 & 0.789 & \cellcolor{blue!25}0.823 & 0.792 & 0.669 & \cellcolor{blue!25}0.897 & 0.788 & 0.884 & 0.813 \\ \cline{2-15}
 & D1, D2, D3 & MR / 864 & 0.873 & 0.818 & 0.872 & 0.861 & 0.797 & 0.814 & 0.788 & 0.670 & \cellcolor{yellow}0.896 & 0.793 & \cellcolor{yellow}0.924 & 0.778 \\ \hline \hline 
\multirow{12}{*}{VGG-16} & D3 & 56x56 / 256 & 0.886 & 0.861 & 0.886 & 0.879 & 0.819 & 0.817 & 0.788 & 0.676 & 0.842 & 0.824 & 0.915 & 0.822 \\ \cline{2-15}
 & D4 & 28x28 / 512 & \cellcolor{green}0.905 & 0.877 & \cellcolor{green}0.921 & \cellcolor{green}0.905 & \cellcolor{blue!25}0.853 & \cellcolor{green}0.861 & \cellcolor{green}0.855 & \cellcolor{yellow}0.729 & \cellcolor{green}0.934 & 0.833 & \cellcolor{blue!25}0.944 & \cellcolor{green}0.850 \\ \cline{2-15}
 & D5 & 14x14 / 512 & \cellcolor{red!50}0.746 & \cellcolor{red!50}0.752 & \cellcolor{red!50}0.701 & \cellcolor{red!50}0.750 & \cellcolor{red!50}0.715 & \cellcolor{red!50}0.710 & 0.691 & \cellcolor{red!50}0.453 & \cellcolor{red!50}0.593 & \cellcolor{red!50}0.608 & \cellcolor{red!50}0.731 & \cellcolor{red!50}0.665 \\ \cline{2-15}
 & D1, D2 & MR. / 192 & 0.835 & 0.772 & 0.826 & 0.813 & 0.747 & 0.761 & 0.685 & 0.559 & 0.750 & 0.754 & 0.863 & 0.727 \\ \cline{2-15}
 & D1, D4 & MR / 576 & \cellcolor{yellow}0.894 & 0.862 & 0.904 & 0.887 & \cellcolor{yellow}0.851 & 0.808 & 0.829 & 0.723 & 0.772 & \cellcolor{blue!25}0.839 & 0.927 & 0.816 \\ \cline{2-15}
 & D3, D4 & MR / 768 & 0.885 & 0.849 & \cellcolor{blue!25}0.906 & 0.879 & 0.841 & 0.796 & 0.815 & \cellcolor{green}0.732 & 0.783 & 0.824 & 0.917 & 0.808 \\ \cline{2-15}
 & D4, D5 & MR / 1024 & \cellcolor{blue!25}0.898 & \cellcolor{yellow}0.879 & 0.898 & \cellcolor{blue!25}0.895 & \cellcolor{green}0.855 & \cellcolor{blue!25}0.855 & \cellcolor{blue!25}0.839 & 0.704 & \cellcolor{blue!25}0.921 & 0.808 & \cellcolor{yellow}0.932 & 0.829 \\ \cline{2-15}
 & D1, D3, D4 & MR / 832 & 0.891 & 0.859 & \cellcolor{yellow}0.905 & 0.886 & 0828 & 0.812 & 0.780 & 0.720 & 0.882 & 0.806 & 0.925 & 0.819 \\ \cline{2-15}
 & D1, D4, D5 & MR / 1078 & 0.889 & 0.853 & 0.861 & 0.881 & 0.822 & \cellcolor{yellow}0.853 & \cellcolor{yellow}0.838 & 0.725 & \cellcolor{yellow}0.899 & 0.835 & \cellcolor{green}0.954 & 0.809 \\ \cline{2-15}
 & D2, D3, D4 & MR / 896 & 0.893 & 0.862 & 0.904 & \cellcolor{yellow}0.888 & 0.831 & 0.813 & 0.781 & \cellcolor{blue!25}0.731 & 0.888 & 0.808 & 0.928 & 0.820 \\ \cline{2-15}
 & D2, D3, D4, D5 & MR / 1408 & 0.884 & \cellcolor{blue!25}0.882 & 0.851 & 0.877 & 0.817 & 0.846 & 0.830 & 0.726 & 0.877 & \cellcolor{yellow}0.837 & 0.913 & \cellcolor{yellow}0.843 \\ \cline{2-15}
 & D1, D2, D3, D4, D5 & MR / 1472 & 0.886 & \cellcolor{green}0.886 & 0.851 & 0.878 & 0.819 & 0.849 & 0.832 & 0.728 & 0.879 & \cellcolor{green}0.841 & 0.915 & \cellcolor{blue!25}0.844 \\ \hline \hline 
\multirow{15}{*}{GoogLeNet} & D2 & 56x56 / 192 & 0.890 & 0.861 & 0.901 & 0.885 & 0.827 & 0.838 & 0.824 & 0.702 & \cellcolor{yellow}0.876 & 0.829 & 0.927 & 0.836 \\ \cline{2-15}
 & D3 & 28x28 / 256 & 0.870 & 0.857 & \cellcolor{yellow}0.903 & 0.858 & 0.793 & 0.803 & 0.768 & 0.521 & 0.755 & 0.759 & 0.888 & 0.805 \\ \cline{2-15}
 & D5 & 7x7 / 832 & \cellcolor{red!50}0.556 & \cellcolor{red!50}0.474 & \cellcolor{red!50}0.539 & \cellcolor{red!50}0.559 & \cellcolor{red!50}0.553 & \cellcolor{red!50}0.445 & \cellcolor{red!50}0.443 & \cellcolor{red!50}0.367 & \cellcolor{red!50}0.383 & \cellcolor{red!50}0.539 & \cellcolor{red!50}0.511 & \cellcolor{red!50}0.565 \\ \cline{2-15}
 & D2, D3 & MR / 448 & \cellcolor{blue!25}0.904 & \cellcolor{blue!25}0.886 & 0.901 & \cellcolor{green}0.903 & \cellcolor{blue!25}0.850 & \cellcolor{blue!25}0.847 & \cellcolor{blue!25}0.830 & 0.715 & \cellcolor{green}0.885 & \cellcolor{yellow}0.837 & \cellcolor{green}0.947 & \cellcolor{blue!25}0.842 \\ \cline{2-15}
 & D2, D4 & MR / 720 & 0.891 & 0.860 & 0.901 & 0.885 & 0.828 & 0.805 & 0.773 & 0.717 & \cellcolor{blue!25}0.879 & 0.803 & 0.923 & 0.814 \\ \cline{2-15}
 & D2, D5 & MR / 1024 & 0.870 & 0.823 & 0.859 & 0.859 & 0.794 & 0.800 & 0.769 & 0.700 & \cellcolor{yellow}0.876 & 0.798 & 0.920 & 0.779 \\ \cline{2-15}
 & D3, D4 & MR / 784 & 0.879 & 0.841 & \cellcolor{blue!25}0.904 & 0.868 & 0.827 & 0.769 & 0.781 & 0.719 & 0.774 & 0.805 & 0.898 & 0.783 \\ \cline{2-15}
 & D3, D5 & MR / 1088 & 0.861 & 0.840 & \cellcolor{yellow}0.903 & 0.845 & 0.777 & 0.768 & 0.717 & 0.538 & 0.770 & 0.731 & 0.868 & 0.781 \\ \cline{2-15}
 & D4, D5 & MR / 1360 & 0.866 & 0.859 & 0.895 & 0.887 & \cellcolor{green}0.879 & 0.841 & 0.807 & \cellcolor{green}0.787 & 0.791 & \cellcolor{green}0.914 & 0.846 & \cellcolor{green}0.855 \\ \cline{2-15}
 & D1, D2, D3 & MR / 512 & \cellcolor{yellow}0.903 & \cellcolor{yellow}0.884 & 0.898 & \cellcolor{yellow}0.901 & 0.847 & 0.841 & 0.821 & 0.706 & 0.870 & \cellcolor{yellow}0.837 & 0.944 & 0.838 \\ \cline{2-15}
 & D1, D3, D4 & MR / 848 & 0.848 & 0.830 & 0.786 & 0.826 & 0.755 & 0.831 & 0.808 & 0.719 & 0.848 & \cellcolor{blue!25}0.841 & 0.884 & 0.789 \\ \cline{2-15} 
 & D2, D3, D4 & MR / 976 & 0.872 & 0.861 & 0.848 & 0.861 & 0.797 & 0.808 & 0.773 & \cellcolor{blue!25}0.728 & 0.885 & 0.807 & 0.891 & 0.816 \\ \cline{2-15}
 & D2, D3, D5 & MR / 1280 & 0.902 & 0.882 & 0.900 & 0.900 & 0.846 & \cellcolor{yellow}0.845 & \cellcolor{yellow}0.827 & 0.706 & 0.876 & 0.836 & \cellcolor{blue!25}0.946 & \cellcolor{yellow}0.841 \\ \cline{2-15}
 & D3, D4, D5 & MR / 1616 & \cellcolor{green}0.906 & \cellcolor{green}0.888 & \cellcolor{green}0.907 & \cellcolor{blue!25}0.902 & \cellcolor{yellow}0.849 & \cellcolor{green}0.849 & \cellcolor{green}0.834 & 0.699 & 0.876 & 0.833 & 0.945 & 0.835 \\ \cline{2-15}
 & D1, D3, D4, D5 & MR / 1680 & 0.843 & 0.820 & 0.787 & 0.820 & 0.747 & 0.813 & 0.789 & \cellcolor{yellow}0.725 & 0.854 & 0.825 & 0.874 & 0.776 \\ \hline \hline 
\multirow{9}{*}{ResNet-50} & D3 & 28x28 / 512 & \cellcolor{blue!25}0.900 & 0.869 & \cellcolor{blue!25}0.913 & \cellcolor{blue!25}0.896 & \cellcolor{blue!25}0.839 & 0.863 & \cellcolor{yellow}0.847 & 0.716 & \cellcolor{yellow}0.925 & \cellcolor{blue!25}0.835 & \cellcolor{green}0.943 & \cellcolor{blue!25}0.851 \\ \cline{2-15}
 & D5 & 7x7 / 2048 & \cellcolor{red!50}0.606 & \cellcolor{red!50}0.621 & \cellcolor{red!50}0.565 & \cellcolor{red!50}0.623 & \cellcolor{red!50}0.593 & \cellcolor{red!50}0.408 & \cellcolor{red!50}0.447 & \cellcolor{red!50}0.217 & \cellcolor{red!50}0.307 & \cellcolor{red!50}0.466 & \cellcolor{red!50}0.586 & \cellcolor{red!50}0.551 \\ \cline{2-15}
 & D3, D4 & MR / 1536 & 0.885 & \cellcolor{blue!25}0.876 & 0.867 & 0.876 & 0.813 & \cellcolor{blue!25}0.874 & \cellcolor{blue!25}0.852 & 0.700 & 0.921 & \cellcolor{green}0.836 & 0.914 & \cellcolor{blue!25}0.851 \\ \cline{2-15}
 & D3, D5 & MR / 2560 & \cellcolor{green}0.901 & \cellcolor{yellow}0.870 & \cellcolor{green}0.918 & \cellcolor{green}0.897 & \cellcolor{green}0.840 & \cellcolor{yellow}0.866 & \cellcolor{blue!25}0.852 & 0.719 & \cellcolor{blue!25}0.933 & \cellcolor{yellow}0.829 & \cellcolor{blue!25}0.941 & \cellcolor{yellow}0.846 \\ \cline{2-15}
 & D1, D3, D4 & MR / 1600 & 0.884 & 0.848 & \cellcolor{yellow}0.903 & 0.874 & 0.814 & 0.815 & 0.781 & 0.720 & 0.879 & 0.811 & 0.913 & 0.816 \\ \cline{2-15}
 & D1, D3, D5 & MR / 2624 & 0.886 & 0.851 & 0.899 & 0.877 & 0.819 & 0.827 & 0.834 & \cellcolor{green}0.734 & 0.887 & 0.819 & 0.918 & 0.823 \\ \cline{2-15}
 & D2, D3, D5 & MR / 2816 & 0.865 & 0.862 & 0.832 & 0.850 & 0.787 & 0.827 & 0.804 & 0.708 & 0.849 & \cellcolor{blue!25}0.835 & 0.881 & 0.821 \\ \cline{2-15}
 & D3, D4, D5 & MR / 3584 & \cellcolor{yellow}0.893 & \cellcolor{green}0.889 & 0.870 & \cellcolor{yellow}0.887 & \cellcolor{yellow}0.827 & \cellcolor{green}0.879 & \cellcolor{green}0.866 & \cellcolor{yellow}0.731 & \cellcolor{green}0.951 & \cellcolor{green}0.836 & \cellcolor{yellow}0.929 & \cellcolor{green}0.854 \\ \cline{2-15}
 & D1, D3, D4, D5 & MR / 3648 & 0.878 & 0.837 & 0.888 & 0.866 & 0.813 & 0.812 & 0.782 & \cellcolor{blue!25}0.732 & 0.885 & 0.794 & 0.901 & 0.803 \\ \hline 
\end{tabular}
}
\label{PrecisionDic}
\end{table}
\subsubsection{Comprehensive Analysis Results}
\label{sec:3_2_analysis_results}
In the following, the fundamental remarks of comprehensive analysis are presented. \\
\indent\textbf{\textit{Remark 1:}} It is evident from the results that the last convolutional layer of CNN models has the worst tracking performance. The reason is that these models have been trained to classify objects in the last layer. Due to strategies of dimension reduction (e.g., pooling layers), the last convolutional layers have a low spatial resolution such that the accurate localization of the target is not possible. This observation has already been demonstrated by \cite{CCOT,FCNT,HCFT,HCFTs} as the last convolutional layer captures the semantic object category while suffering from a coarse resolution for accurate localization.\\
\indent\textbf{\textit{Remark 2:}} Most of deep learning-based trackers \cite{FCNT,DeepMotionFeatures,DNT,HCFTs,SurveyDeepTracking,DL-Tracking-Survey} have mentioned that the multi-level feature maps (shallow \& deep convolutional layers) enhances the performance of visual trackers. For example, these multi-level features mostly improve scale estimation process of visual trackers. However, there is not any specific rule on how to combine the CNN feature maps to achieve the best visual tracking performance. For example, deep visual trackers in \cite{FCNT,DNT} utilize the combination of D4 and D5 layers from the VGG-16 model. This combination provides 1024 feature maps and leads to the second rank in performance evaluation while the single D4 layer from this model has 512 feature maps and achieves the best performance regarding the success and precision metrics. Depending on the desired precision, success, and computational complexity (more feature maps, more complexity), Table~\ref{SuccessDic} and Table~\ref{PrecisionDic} indicate the most reasonable features and combinations for visual tracking. \\
\indent\textbf{\textit{Remark 3:}} The increase in the number of feature maps does not always improve the tracking performance but also considerably increases the computational complexity. However, it may enhance the performance in the presence of specific attributes. For instance, the combination of D3, D4, and D5 of ResNet-50 improves the tracking performance against the SV, FM, MB, OV, and IV attributes. Note that it generally adds redundant feature maps that are not properly involving to discriminate the target from its background. Hence, blindly increase the number of feature maps may significantly reduce both the tracking speed and the performance. For example, this observation has been employed in FCNT tracker \cite{FCNT} such that a feature map selection process is performed on the D4 and D5 layers of the VGG-16 model to avoid over-fitting on noisy feature maps.\\
\indent\textbf{\textit{Remark 4:}} The most destructive impact on performance is related to the LR. This problem is originated from the limited number of pixels that represent target information. Recently, this issues has been investigated in various computer vision tasks \cite{SmallDetReview,SCRDet,COMET}. According to the achieved results, employing shallow and deep convolutional layers could alleviate this deficiency.    \\
\indent\textbf{\textit{Remark 5:}} Although the use of a fixed number of layers brings simplicity, adaptive exploitation of deep features grants flexibility to visual tracking methods. Considering analysis results, deep features provide distinct responses to the attributes. Thereby, fixed features possibly reduce the tracking performance in challenging scenarios and also limits the robustness of trackers. Therefore, visual trackers can select different CNN layers based on their application or aims to enhance the accuracy, robustness, or a trade-off between accuracy and robustness. Moreover, the feature maps do not equally respond to the attributes (each layer might be sensitive to some of them) due to different parameters and architectures of CNN models. For instance, the D4 layer in the VGG-16 model has the most acceptable results against the most attributes while low-resolution targets dramatically impact on its performance. Moreover, the efficiency level for each layer is related to the objective of each visual tracker. As such, based on the precision, success, or both, the layer(s) selection options may differ. This important property was in fact the primary motivation of this paper to adaptively exploit the CNN feature maps for visual tracking. \\
\indent To integrate all the benefits into an adaptive visual tracking method, the following proposed adaptive method exploits the results of the comprehensive analysis (i.e., Table~\ref{SuccessDic} and Table~\ref{PrecisionDic}) as the attribute dictionaries of the CNN models, referred as precision and success dictionaries. These attribute dictionaries include apparent and latent characteristics of models, which are effective for visual tracking.
\subsection{Proposed Adaptive Exploitation of Deep Features}
\label{sec:AdaptiveExplo}
The proposed method composed of determination of an attribute vector, integration of attribute dictionaries into the DCF formulations, and a DCF-based tracker. Furthermore, the proposed method can be more sophistically incorporated into other DCF-based tracking methods considering their specific characteristics.
\subsubsection{Attribute Vector Determination}
\label{sec:AttributeDeter}
Visual attributes can be roughly categorized according to the related characteristics of object, camera, and environment. As a result, visual trackers can use such categorized attributes to create an attribute vector for their applications. Some of these attributes can be effortlessly specified from the initial bounding box of a target in the first frame. For instance, an object recognition process can specify whether the object is rigid or non-rigid; Or, target resolution can be determined by counting its number of pixels. Moreover, visual tracking methods can achieve valuable information about visual attributes based on specific applications; as an example, different options that can be adjusted by a user. Visual attribute detection methods \cite{AttributeDetection1,AttributeDetection2,AttributeDetection3,AttributeDetection4} also can be incorporated with visual trackers for estimating an attribute vector for each frame. Moreover, the visual tracking methods can estimate visual attributes based on their definitions in visual tracking datasets; For instance, the definition of IV, SV, BC, MB, DEF, object motion, camera motion, aspect-ratio change, scene complexity, and absolute motion in \cite{VOT-2015}. However, this section focuses on the investigation of adaptive exploitation of deep features and its effects on tracking performance. Thus, employing approaches for visual attribute detection are beyond the objectives of this section and will be studied in future works. But, the per-frame estimation of attribute vectors is still an open problem in visual tracking. \\
\indent It is assumed that an attribute vector (i.e., a full or incomplete vector) is provided for the visual tracker hereafter. For each application, the attribute vector will be an eleven-component vector such that each component is specified by zero or one (binary values). If there is a probability of occurring each attribute, the corresponding component will be set to one. Hence, the proposed method can adaptively select the best features according to the specified attribute vector, which can represent all the joint combinations of challenging attributes. It is evident that the proposed method selects the best overall feature layers if all components of the attribute vector are zero or one. In this work, the attribute vector of each video sequence is exploited which is provided by visual tracking datasets to figure out the effect of the proposed adaptive method. Also, the generalization of attribute dictionaries will be validated regarding different attribute vectors of the UAV-123 dataset \cite{UAV123}, which can be considered as imprecise attribute vectors.
\subsubsection{Revisited Formulation of DCF-based Visual Trackers}
\label{sec:RevisitFormul}
Generally, DCF-based visual tracking methods aim to learn a set of convolution filters by minimizing the objective function as
\begin{equation}\label{Eq.1}
{\mathrm{arg} {\mathop{\mathrm{min}}_{h} \frac{1}{2}\ }{\left\|\sum^K_{k=1}{x^k_j*h^k}-y\right\|}^2_2\ }+\frac{1}{2}\sum^K_{k=1}{{\left\|w\cdot h^k\right\|}^2_2}
\end{equation}
in which $x_j$, $h$, $y$, and $w$ are the $j^{th}$ training sample, multi-channel convolution filters, desired Gaussian response, spatial regularization matrix, respectively. Also, $K$, and $*$ represent a fixed number of feature channels and convolution operator, respectively. By defining additional terms, DCF-based trackers form various expressions such that the filters will have been learned via closed-form solutions or iterative algorithms (e.g., \cite{SRDCF,STRCF,BACF}). \\
\indent The proposed method can integrate into any form of current DCF-based trackers that use CNN models. It adaptively selects the best convolutional layers for visual tracking applications. Given attribute dictionaries of CNN models and attribute vector of tracking, the proposed method selects the best trade-off between the precision and success metrics, which ensure the best accuracy and robustness for tracking. The proposed method defines an ordered multi-label set $\mathcal{S}=\{{\mathcal{\zeta}}^1,{\mathcal{\zeta}}^2,\cdots,{\mathcal{\zeta}}^N\}$
, in which ${\mathcal{\zeta}}^i = \left\{{\mathcal{L}}^i_1,{\mathcal{L}}^i_2,\cdots ,{\mathcal{L}}^i_{L}\right\}$ indicates the available configurations of models in Table~\ref{SuccessDic} and Table~\ref{PrecisionDic}. The configurations can be defined by ${\mathcal{L}}^i_j\in \{D1,D2,\cdots ,D5\}$. Also, $N$ and $L$ are the maximum number of configurations (i.e., number of single and combined layers) and the maximum number of test output for each model, respectively. For instance, ${\mathcal{\zeta}}^1$ and ${\mathcal{\zeta}}^7$ comprises $\{D1\}$ and $\{D1, D2, D3\}$ for the success dictionary of VGG-M model, respectively. \\ 
\indent For each CNN model, the proposed method defines an ordered pair 
$\mathcal{C}=\{\{a_1,b_1\},\{a_2,b_2\},\cdots ,\{a_L,b_L\}\}$
, in which $a_i$ and $b_i$ indicate the test output (according to Table~\ref{NetConfigs}, and corresponding feature channels, respectively. For example, we have $\mathcal{C}=\{\{a_1,b_1\}=\{D1,96\},\{a_2,b_2\}=\{D2,256\},\{a_3,b_3\}=\{D3,512\}\}$ for the VGG-M model. The attribute vector is denoted as $z$ with the length of $M=11$ (i.e., the number of attributes). The proposed objective function is defined as 
\begin{equation}\label{Eq.2}
{\mathcal{\zeta}}^i{{:=}}{\mathrm{arg} {\mathop{\mathrm{max}}_{\mathcal{S}} \left(\frac{1}{2}(\left(z^T.P_1\right)+\left(z^T.P_2\right))\right)\ }\ }
\end{equation}
where the precision and success dictionaries are indicated by $P_1$ and $P_2$ matrices with $M\times N$ dimension, respectively. As mentioned in Sec.~\ref{sec:3_1_analysis}, Table~\ref{SuccessDic} and Table~\ref{PrecisionDic} just represent the best and worst feature maps, and the completed analyses are provided in Appendix~\ref{appendix}. Note that to provide consistency to select network configurations, all the precision and success results of best settings are used in experimental evaluations. It means that for each dictionary there are 51 configurations, which 36 configurations are common in Table~\ref{SuccessDic} \& Table~\ref{PrecisionDic}, and the others are completed by the corresponding ones in Appendix~\ref{appendix}. Based on the objective function, the proposed method selects the best convolutional feature maps, which result in the best trade-off between the accuracy and robustness for a tracking application. Then, it computes 
\begin{equation}\label{Eq.3}
K=\sum^L_{n=1}{b_n} \ \ \ \ \ \ \ s.t.\left({\mathcal{L}}^i_n,{b}_n\right)\in\mathcal{C}
\end{equation}
to adaptively determine the number of feature channels for DCF-based visual tracking methods. In fact, thanks to having the attribute dictionaries from comprehensive analysis and also the attribute vector for each application, the proposed method can automatically and quickly select the best CNN feature maps which are robust to realistic challenges and are applicable in DCF-based visual tracking methods. \\
\indent To demonstrate the effectiveness, the proposed method is integrated into two well-known DCF-based trackers, namely ECO \cite{ECO} and BACF \cite{BACF}, which are properly modified to exploit deep features extracted by CNN models with various topologies. Since the dimension reduction of deep features has been removed for fair comparisons, the proposed ECO-based tracker aims to minimize the following loss function \cite{ECO}
\begin{equation}\label{Eq.eco_loss}
E\left(h\right)=\mathbb{E}\left\{{\left\|\sum^\kappa_{n=1}({h^n*V_k\{x^n}\})-y\right\|}^2_2+\sum^K_{k=1}{{\left\|w\cdot h^k\right\|}^2_2}\right\}\
\end{equation}
in which $\mathbb{E}$,  $G_h\{x\}={h^k*V_k\{x^k}\}$, and $\kappa$ represent the mathematical expectation (i.e., expected value), detection score of the target, and the number of training samples, respectively. Also, it performs convolutions in the continuous domain as well as directly predicts detection scores by the interpolation operator $V$. The loss function~(\ref{Eq.eco_loss}) presents a quadratic problem with the closed-form solution $(\Lambda^H\Phi \Lambda+W^HW)\hat{h}=\Lambda^H\Phi\hat{y}$, which $\hat{h}$ and $\hat{y}$ are the vectorized Fourier coefficients of $h$ and $y$, respectively. Also, the Hermitian operator is denoted by $^H$; and $\Lambda$, $\Phi$, and $W$ represent the matrices of interpolated target samples, sample weights, and regularization, respectively. At last, it is iteratively optimized by the Gauss-Newton and Conjugate Gradient methods to achieve convolution filters $h$ (for more details, see \cite{ECO}). \\
\indent Furthermore, the generalization of proposed method is investigated by the proposed DeepBACF tracker that aims to model foreground and background of target by the following objective function \cite{BACF}
\begin{equation}\label{Eq.bacf_loss}
E\left(h\right)=\frac{1}{2}\sum^\kappa_{n=1}{\left\|y(n)-  \sum^K_{k=1}{{h^T_k \Pi x_k[\Delta\tau_j]}^2_2}\right\|}^2_2  + \frac{\lambda}{2} \sum^K_{k=1}{{\left\|h^k\right\|}^2_2}\
\end{equation}
where $\Pi$, $[\Delta\tau_j]$, $\lambda$, and $^T$ are cropping operator, circular shift operator, regularization term, and conjugate transpose operator, respectively. The corresponding filters in frequency domain can be expressed as
\begin{equation}\label{Eq.bacf_freq}
\begin{split}
E & \left(\bold{h},\bold{\hat{g}}\right) =\frac{1}{2}{\left\|\bold{\hat{y}}-  \bold{\hat{X}}\bold{\hat{g}}\right\|}^2_2  + \frac{\lambda}{2} {\left\|\bold{h}\right\|}^2_2 \\
& \text{s.t.} \ \ \ \bold{\hat{g}}=\sqrt{T}(\bold{F}\bold{\Pi}^T\otimes\bold{I}_K)\bold{h}
\end{split}
\end{equation}
in which $\bold{\hat{g}}$, $\bold{F}$, $\bold{I}_K$, and $\otimes$ indicate an auxiliary variable, orthonormal matrix of complex Fourier basis vectors, identity matrix, and Kronecker product, respectively. Finally, the loss function~(\ref{Eq.bacf_freq}) is iteratively optimized following the \textit{alternating direction method of multipliers} (ADMM) \cite{ADMM}, which breaks the augmented Lagrangian form of Eq.~(\ref{Eq.bacf_freq}) into three sub-problems and optimizes one at each step (see \cite{BACF} for more details). Algorithm~\ref{alg:PropMethods} shows the the process of proposed method, which is integrated into DCF-based trackers for adaptive exploitation of deep features.

\begin{algorithm}
\caption{Proposed Adaptive Exploitation of Deep Features for DCF Framework}
\label{alg:PropMethods}
\textbf{Input:} Pre-trained CNN models (VGG-M \cite{VGGM}, VGG-16 \cite{VGGNet}, GoogLeNet \cite{GoogLeNet}, ResNet-50 \cite{ResNet}), A DCF-based tracker, Sequence frames, Initial \textit{bounding box} (BB) of target (i.e., target region), Attribute vector of sequence \leavevmode \\ 

\textbf{Prerequisite:} Comprehensive analysis of FENs on OTB-2013 dataset \\
  - Specify configurations of models   \\
  \For{each CNN model}{
    Evaluate the tracker on the single layers, independently \leavevmode \\
    Evaluate the tracker on all combination of layers 
  }
  - Validate generalization of results on another DCF-based tracker  \\
  - Validate generalization of model dictionaries on other models with similar architecture  \\
\textbf{Analysis Output:} Attribute dictionaries of CNN models ($P_1$: Precision dictionary, $P_2$: Success dictionary) \\

\For{A Video sequence \& CNN model}{
    Define ordered multi-label set $\mathcal{S}$  \leavevmode \\
    Define ordered pair $\mathcal{C}$  \leavevmode \\
    Select the best feature maps (${\mathcal{\zeta}}^i$) by Eq.~(\ref{Eq.2}) \leavevmode \\
    Compute the number of channels ($K$) by Eq.~(\ref{Eq.3}) \leavevmode \\
    \textbf{Output:} Best feature maps (single or combined convolutional layers), Number of channels  \leavevmode \\
    Set ${\mathcal{\zeta}}^i$ \& $K$ for a DCF-based tracker  \leavevmode \\
    \For{Sequence frames}{
        Extract deep features \leavevmode \\
        Model target appearance by Eq.~(\ref{Eq.eco_loss}) or Eq.~(\ref{Eq.bacf_freq}) \leavevmode \\
        Optimize correlation/convolution filters by iterative algorithms \leavevmode \\
        Tracking-by-detection \leavevmode \\
        Update target model 
    }
}
 \textbf{Output:} Location and scale of a visual target  
\end{algorithm}
\section{Experimental Results}
\label{sec:4_ExpResults}
In this section, the implementation details and experimental analysis are presented. For the experiments, first, the proposed method is validated with the baseline trackers, which employ a fixed number of feature channels. Then, the generalization of proposed method and analysis results are investigated by another DCF-based tracker and the models with similar architectures, respectively. Finally, the proposed method is extensively evaluated compared the state-of-the-art visual tracking methods.
\subsection{Implementation Details}
\label{sec:4_1_implement_details}
For fair and meaningful comparisons, in baseline comparison, the proposed adaptive method is compared with the baseline trackers (i.e., modified ECO, and modified BACF trackers), which exploit a fixed number of deep features from any CNN models. Note that the number of ADMM’s iterations for the modified BACF is set to 15, such that it can efficiently learn the background-aware correlation filters. However, all other parameters of these trackers are set the same as the baseline ones \cite{ECO,BACF}, and kept fixed through all experiments. Although different settings could provide a better performance, the reported results demonstrate the effectiveness of the proposed method even without any hyper-parameter tuning. The implementations are performed on an Intel I7-6800K 3.40 GHz CPU with 64 GB RAM with the aid of advanced MatConvNet toolbox, which uses an NVIDIA GeForce GTX 1080 GPU for its computations. The qualitative evaluations are conducted as the \textit{one-pass evaluations} (OPEs) on the OTB-2015 \cite{OTB2015}, TC-128 \cite{TC128}, and UAV-123 \cite{UAV123} datasets. Slightly different from previously mentioned attributes, the videos of the UAV-123 dataset also have been labeled by \textit{aspect ratio change} (ARC), \textit{camera motion} (CM), \textit{full occlusion} (FOC), \textit{partial occlusion} (POC), \textit{similar object} (SOB), and \textit{viewpoint change} (VC). Table~\ref{Datasets} illustrates the details of tracking datasets that are used in this work. In addition to the VGG-M, VGG-16, GoogLeNet, and ResNet-50 models in Sec.~\ref{sec:3_1_analysis}, the generalization of attribute dictionaries of the ResNet-50 model is explored by the ResNeXt-50 \cite{ResNeXt}, SE-ResNet-50 \cite{SE-Res-CVPR}, and SE-ResNeXt-50 \cite{SE-Res-CVPR}.

\begin{table}
\caption{Exploited visual tracking datasets in this work [NoV: number of videos, NoF: number of frames].} 
\centering 
\resizebox{\textwidth}{!}{
\begin{tabular}{c|c|c|c|c|c|c|c|c|c|c|c|c|c|c} \hline \hline
\multirow{2}{*}{Dataset} & \multirow{2}{*}{NoV} & \multirow{2}{*}{NoF} & \multicolumn{12}{c}{NoV Per Attribute} \\ \cline{4-15}
 &  &  & IV & OPR & SV & \multicolumn{2}{c|}{OCC} & DEF & MB & FM & IPR & OV & BC & LR \\ \hline \hline 
OTB-2013 \cite{OTB2013} & 51 & 29491 & 25 & 39 & 28 & \multicolumn{2}{c|}{29} & 19 & 12 & 17 & 31 & 6 & 21 & 4 \\ \hline 
OTB-2015 \cite{OTB2015} & 100 & 59040 & 38 & 63 & 65 & \multicolumn{2}{c|}{49} & 44 & 31 & 40 & 53 & 14 & 31 & 10 \\ \hline 
TC-128 \cite{TC128} & 129 & 55346 & 37 & 73 & 66 & \multicolumn{2}{c|}{64} & 38 & 35 & 53 & 59 & 16 & 46 & 21 \\ \hline \hline 
\multirow{2}{*}{UAV-123 \cite{UAV123}} & \multirow{2}{*}{123} & \multirow{2}{*}{112578} & IV & CM & SV & POC & FOC & SOB & ARC & FM & VC & OV & BC & LR \\ \cline{4-15}
 &  &  & 31 & 70 & 109 & 73 & 33 & 39 & 68 & 28 & 60 & 30 & 21 & 48 \\ \hline 
\end{tabular}
}
\label{Datasets}
\end{table}
\subsection{Baseline Comparison}
\label{sec:4_2_baseline_comp}
\begin{figure}
\justify
\subfigure{\includegraphics[width=5.8cm, height=4.5cm]{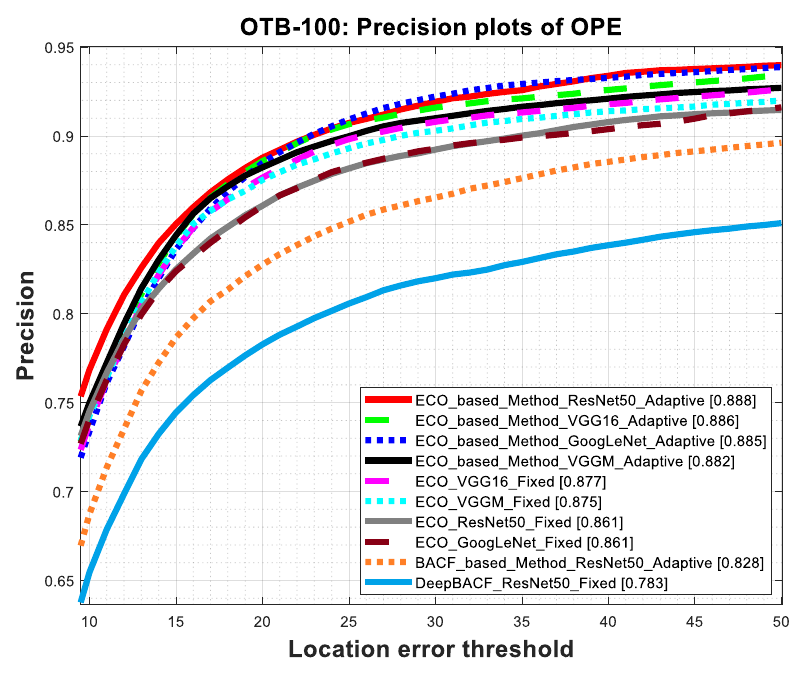}} 
\hspace{0mm}
\subfigure{\includegraphics[width=5.8cm, height=4.5cm]{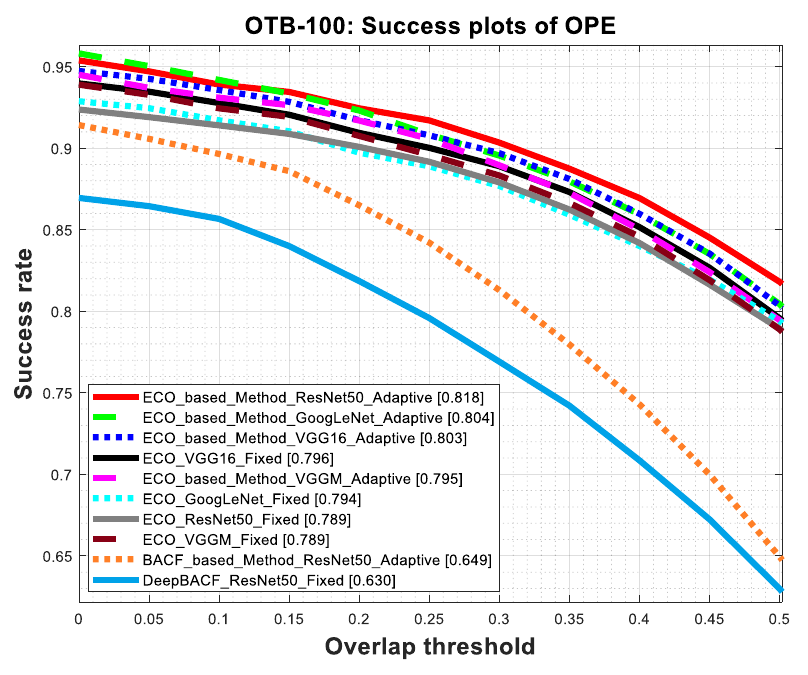}}
\vspace{-4mm}
\justify
\subfigure{\includegraphics[width=5.8cm, height=4.5cm]{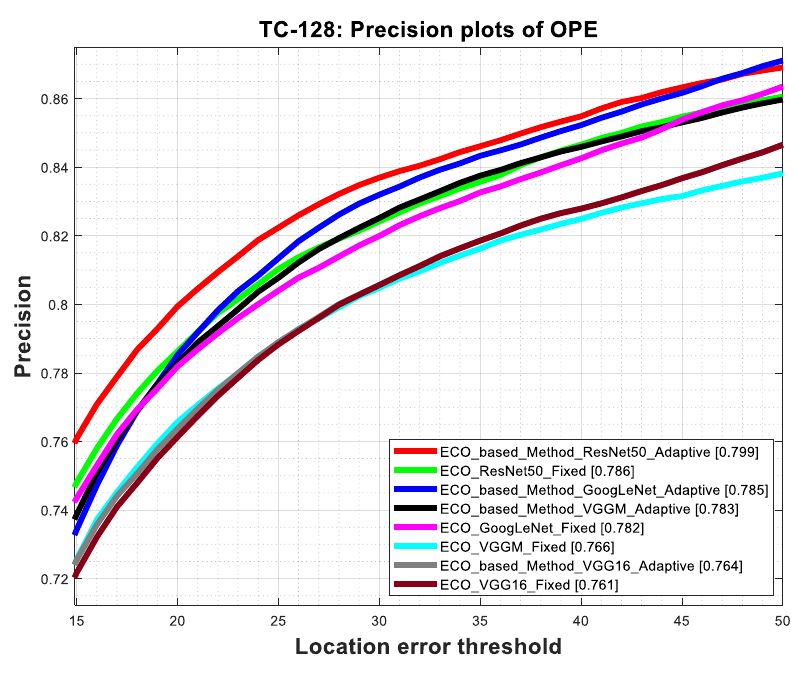}} 
\hspace{0mm}
\subfigure{\includegraphics[width=5.8cm, height=4.5cm]{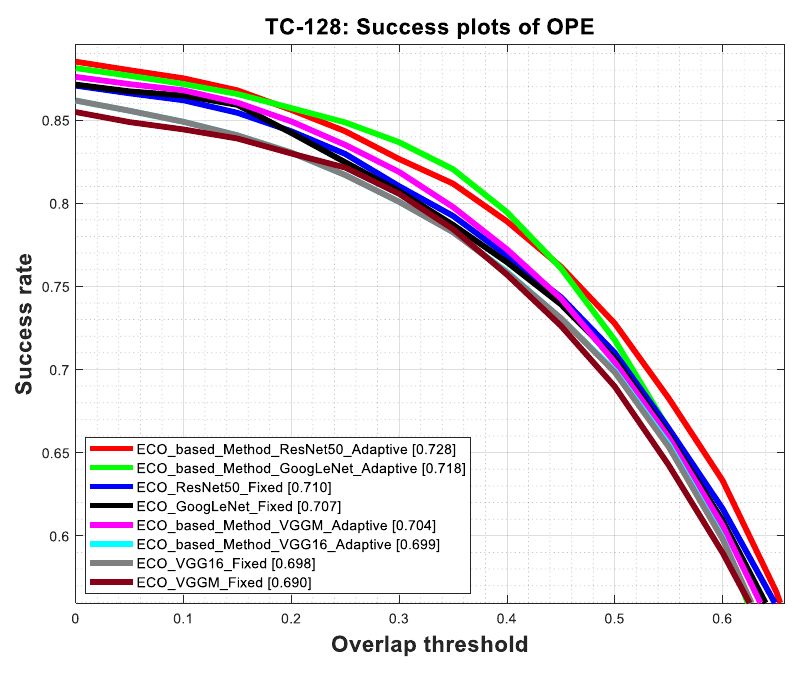}}
\vspace{-4mm}
\justify
\subfigure{\includegraphics[width=5.8cm, height=4.5cm]{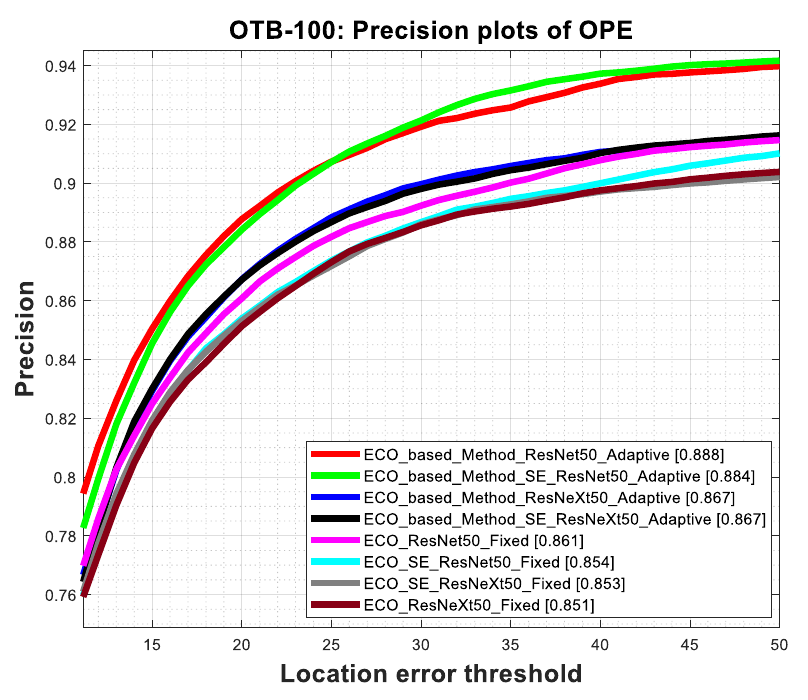}} 
\hspace{0mm}
\subfigure{\includegraphics[width=5.8cm, height=4.5cm]{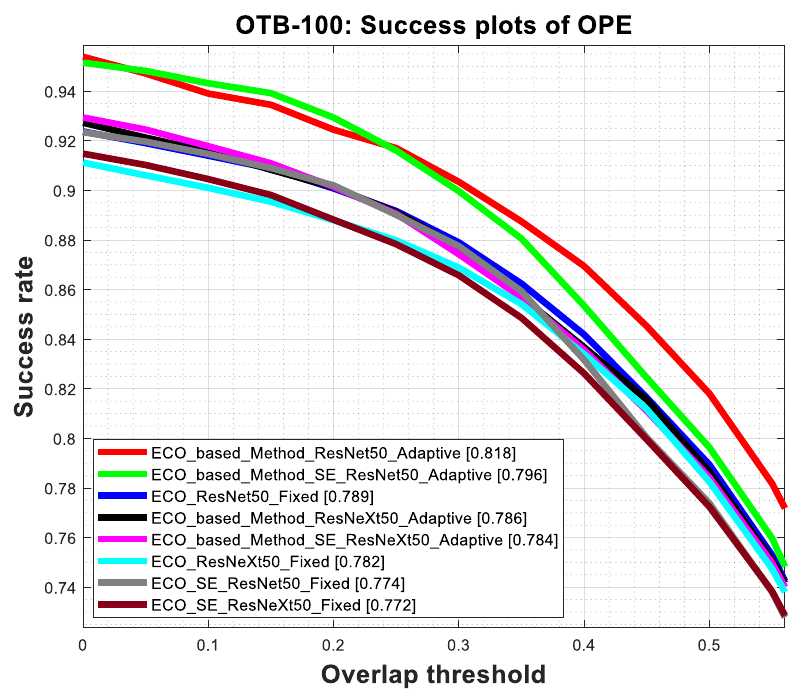}}
\vspace{-2mm}
\caption{Overall precision and success evaluations on the OTB-2015 and TC-128 visual tracking datasets. (top row:) Baseline comparison of proposed adaptive method with na\"ive feature selection and its generalization into the DeepBACF method. (middle row:) Generalization of the baseline comparison on different visual tracking dataset. (bottom row:) Generalization of attribute dictionaries on ResNeXt-50, SE-ResNet-50, and SE-ResNeXt-50 models which have the same architecture with ResNet-50 model.} \label{fig.BaselineComp}
\vspace{-4mm}
\end{figure}

The baseline comparison supports five main aims as follows. First, it confirms the effectiveness of the proposed adaptive method compared with na\"ive feature selection for visual tracking purposes. Second, the best CNN model for the adaptive selection of feature maps is selected. Third, the generalization of the proposed method is investigated on different visual tracking datasets. Fourth, the generalization of the proposed method is explored by integrating it into another DCF-based visual tracker. Finally, the generalization of attribute dictionaries is evaluated by other CNN models with similar architectures. \\
\indent The baseline comparisons are performed on the OTB-2013 and TC-128 datasets. Fig.~\ref{fig.BaselineComp} presents the achieved results by the modified ECO-based tracker, which either employs a fixed number of CNN features or uses the proposed method to utilize adaptive deep features. Note that the proposed adaptive method exploits the results of Table~\ref{SuccessDic} and Table~\ref{PrecisionDic} corresponding to video characteristics, while the best average result (i.e., the average of precision \& success metrics) for each model is considered for setting the fixed features. For instance, the D3 and D5 layers are selected as the fixed configuration of the ResNet-50 model. Based on the results on the OTB-2015 and TC-128 datasets (see Fig.~\ref{fig.BaselineComp} (top \& middle rows)), the proposed method outperforms the average precision and success rates up to 2.7\% and 2.9\% compared with the na\"ive feature selection methods, respectively. Moreover, based on these results and generalization of the results on TC-128 dataset, the ResNet-50 is the best model for visual tracking purposes. It provides more representational power of in the primary and middle layers, which is beneficial for visual tracking. The achieved considerable margin of performances indicates the advantages of the performed comprehensive analysis and adaptive exploitation of deep features. \\
\indent To investigate the generalization ability of the proposed method, it is integrated into another well-known DCF-based tracker, namely BACF \cite{BACF}. The experiments are conducted on the proposed DeepBACF tracker, which is able to employ deep features from various CNN models. Fig.~\ref{fig.BaselineComp} (top row) shows the results of the DeepBACF with either fixed features or adaptive features of the ResNet-50 model on the OTB-2015 dataset. According to it, the proposed method improves the average precision and success rates of the DeepBACF up to 4.5\% and 1.9\%, respectively. \\
\indent The generalization of the attribute dictionaries of the ResNet-50 is extensively evaluated by the pre-trained ResNet-50, ResNeXt-50, SE-ResNet-50, and SE-ResNeXt-50 models which have similar architectures. As shown in Fig.~\ref{fig.BaselineComp} (bottom row), the proposed method has gained up to 1.6\%, 3\%, and 1.4\% in average precision rate, and 0.4\%, 2.2\%, and 1.2\% in average success rate compared with the na\"ive feature selection of ResNeXt-50, SE-ResNet-50, and SE-ResNeXt-50 models, respectively. However, these models have been trained differently (for more details, please refer to \cite{ResNeXt,SE-Res-CVPR}). For instance, these models utilize various building blocks (e.g., split-transform-merge paradigm) to facilitate the training procedure under the restricted complexity. \\
\indent Finally, the generalization of the proposed adaptive method has evaluated by the DeepBACF tracker with the ResNet-50 model on the OTB-2015 dataset (see Fig. 1(a)). The results clearly show that the proposed adaptive method improves the average precision and success rates of the DeepBACF up to 4.5\% and 1.9\%, respectively.

\subsection{Performance Comparison}
\label{sec:4_3_perf_comp}
\begin{figure}
\justify
\subfigure{\includegraphics[width=5.8cm, height=4.5cm]{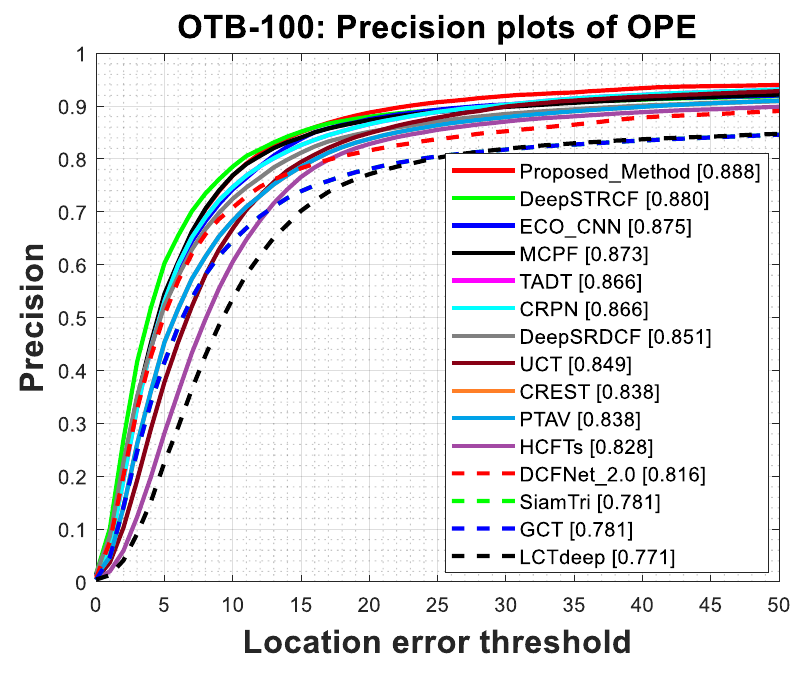}} 
\hspace{0mm}
\subfigure{\includegraphics[width=5.8cm, height=4.5cm]{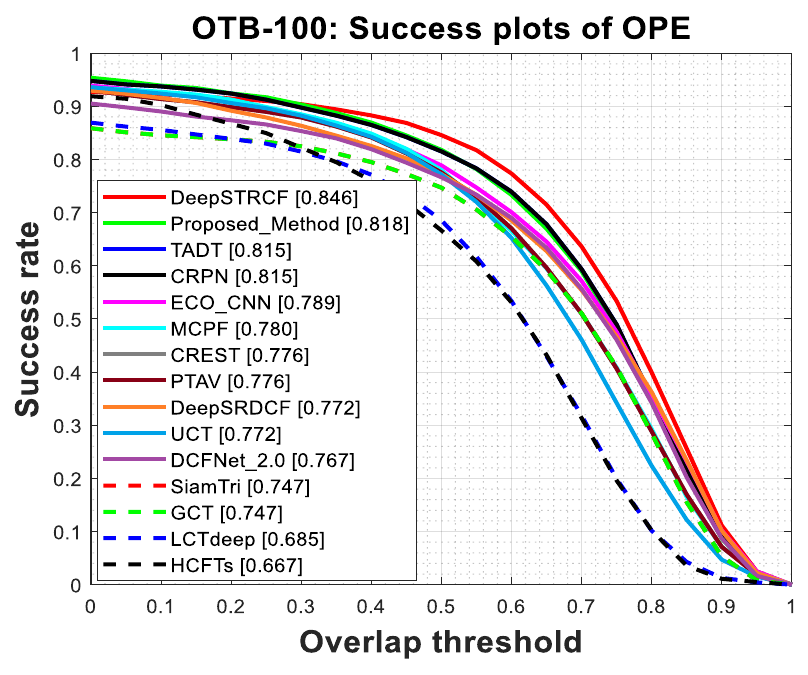}}
\vspace{-4mm}
\justify
\subfigure{\includegraphics[width=5.8cm, height=4.5cm]{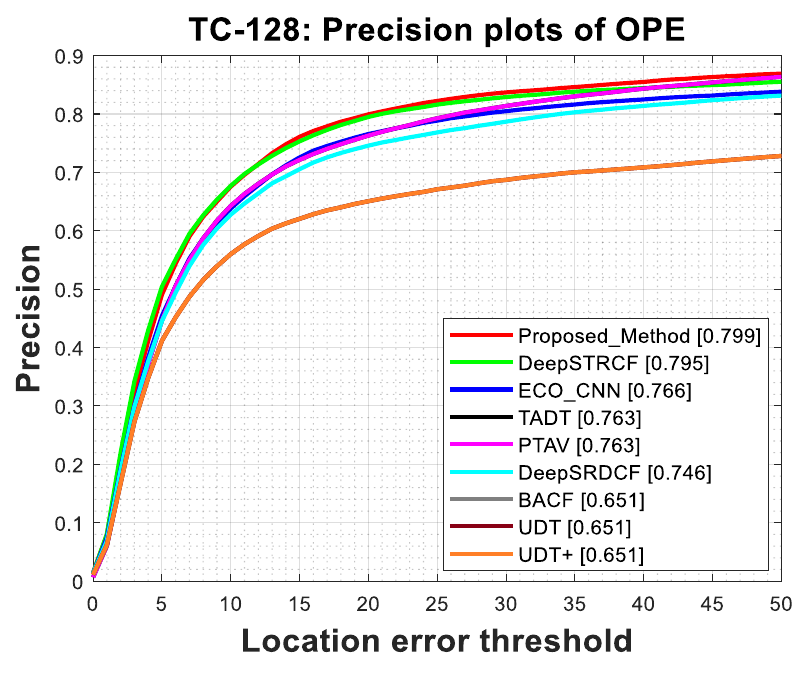}} 
\hspace{0mm}
\subfigure{\includegraphics[width=5.8cm, height=4.5cm]{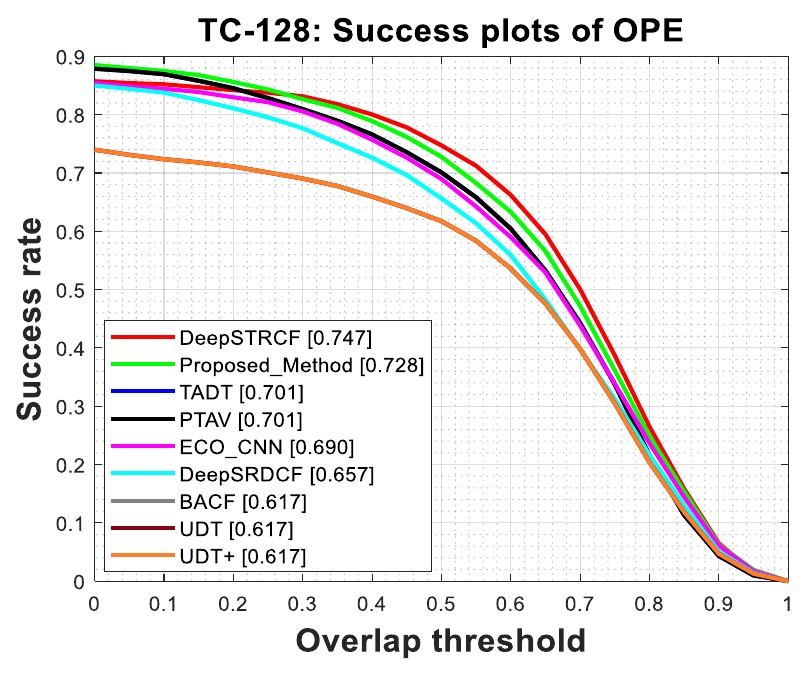}}
\vspace{-4mm}
\justify
\subfigure{\includegraphics[width=5.8cm, height=4.5cm]{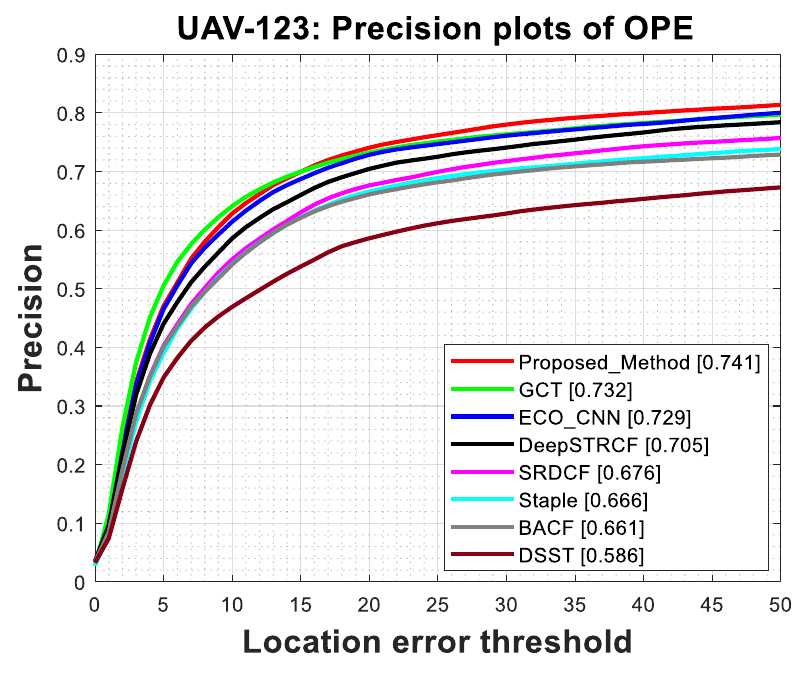}} 
\hspace{0mm}
\subfigure{\includegraphics[width=5.8cm, height=4.5cm]{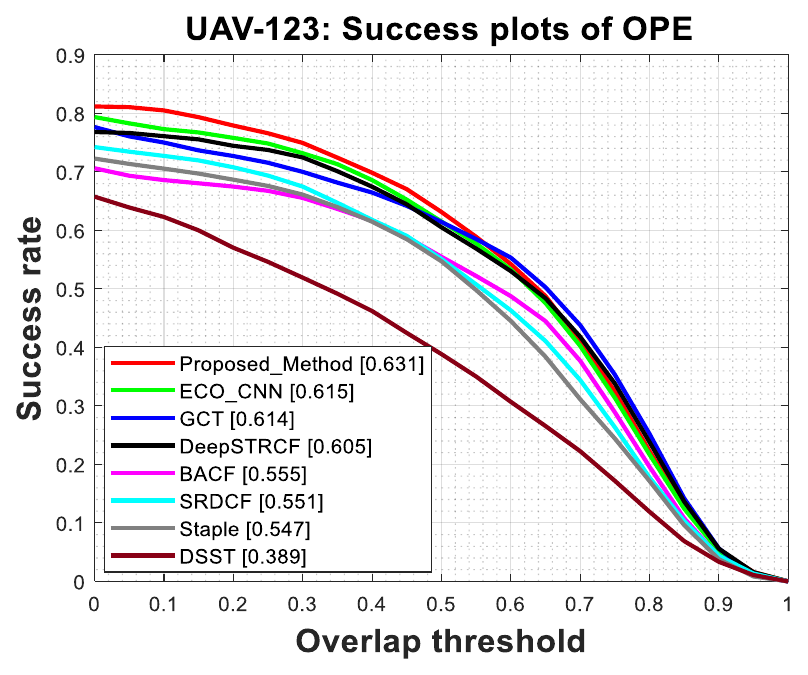}}
\vspace{-2mm}
\caption{Overall precision and success evaluations on OTB-2015, TC-128, and UAV-123 datasets. [Proposed adaptive method using pre-trained ResNet-50 model is compared with the state-of-the-art visual trackers.]} \label{fig.PerfComp}
\vspace{-4mm}
\end{figure}

To quantitatively compare the proposed method with the state-of-the-art trackers, the proposed ResNet-based tracker is selected. It is compared with 14, 8, and 6 state-of-the-art visual trackers (which their benchmark results have been publicly available) on the OTB-2015 \cite{OTB2015}, TC-128 \cite{TC128}, and UAV-123 \cite{UAV123} datasets, respectively. Note that the several attributes of the UAV-123 dataset do not exist in attribute dictionaries. Thus, the experiments on the UAV-123 dataset will indicate the effectiveness of the proposed tracker when the attribute vector is an incomplete or erroneous vector. The proposed tracker is compared with ECO \cite{ECO}, DeepSTRCF \cite{STRCF}, MCPF \cite{MCPF}, TADT \cite{TADT}, CRPN \cite{CRPN}, DeepSRDCF \cite{DeepSRDCF}, UCT \cite{UCT}, CREST \cite{CREST}, PTAV \cite{PTAV}, HCFTs \cite{HCFTs}, DCFNet-2 \cite{DCFNet}, SiamTri \cite{Tripletloss}, GCT \cite{GCT}, LCTdeep \cite{LCTdeep}, BACF \cite{BACF}, UDT \cite{UDT}, UDT$+$ \cite{UDT}, DSST \cite{DSST}, and Staple \cite{Staple}. In addition to the visual trackers that exploit FENs, the proposed method is also compared with the EEN-based trackers, which have been extensively trained on various datasets. Fig.~\ref{fig.PerfComp} shows the overall performance comparisons of visual trackers. \\
\indent According to the results in Fig.~\ref{fig.PerfComp}, the proposed adaptive method outperforms the baseline tracker \cite{ECO} up to 1.9\% and 2.7\% in average of precision and success rates on all datasets. To compare tracking speed, the proposed and baseline \cite{ECO} trackers run at $\sim$6 \& $\sim$10 \textit{frame-per-second} (FPS) on the machine, as mentioned in Sec. \ref{sec:4_1_implement_details}. It means the effective selection of deep features not only improves the tracking performance but also provides an acceptable speed. Since the DeepSTRCF tracker employs the combination of hand-crated and deep features, it has achieved the best success rates on the OTB-2015 and TC-128 dataset. However, the proposed method has gained up to 1.6\% improvement in average precision rate compared with the DeepSTRCF on all datasets. Also, the proposed method can provide more flexibility to another tracking applications compared with other DCF-based trackers such as \cite{DeepSRDCF,STRCF,ECO,BACF}. For example, the average value of precision \& success rates of the proposed method gains up to 1.3\%, 1.4\%, and 3.1\% compared with the GCT, ECO, and DeepSTRCF, respectively. Furthermore, the proposed method has achieved better performances in challenging scenarios comparing with EEN-based trackers \cite{TADT,CRPN,GCT,Tripletloss,UDT,UCT}. As an instance, the proposed tracker outperforms the TADT \cite{TADT}, GCT \cite{GCT}, and UCT \cite{UCT} up to 1.2\%, 4.2\%, and 8.9\% in terms of average precision and success rates on the OTB-2015 dataset, respectively. \\
\begin{figure}
\centering
\subfigure{\includegraphics[width=2.82cm,height=2.8cm]{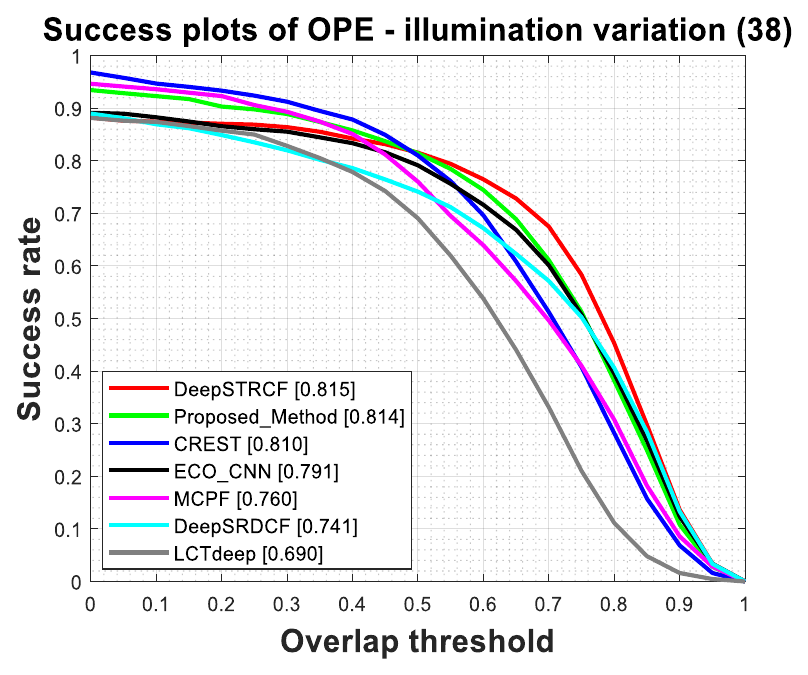}}
\hspace{0mm}
\subfigure{\includegraphics[width=2.82cm,height=2.8cm]{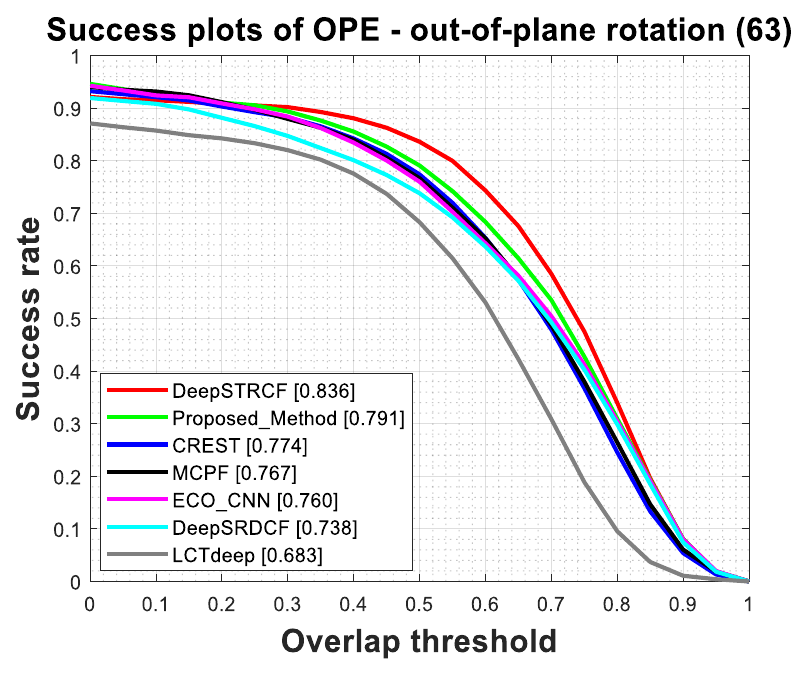}}
\hspace{0mm}
\subfigure{\includegraphics[width=2.82cm,height=2.8cm]{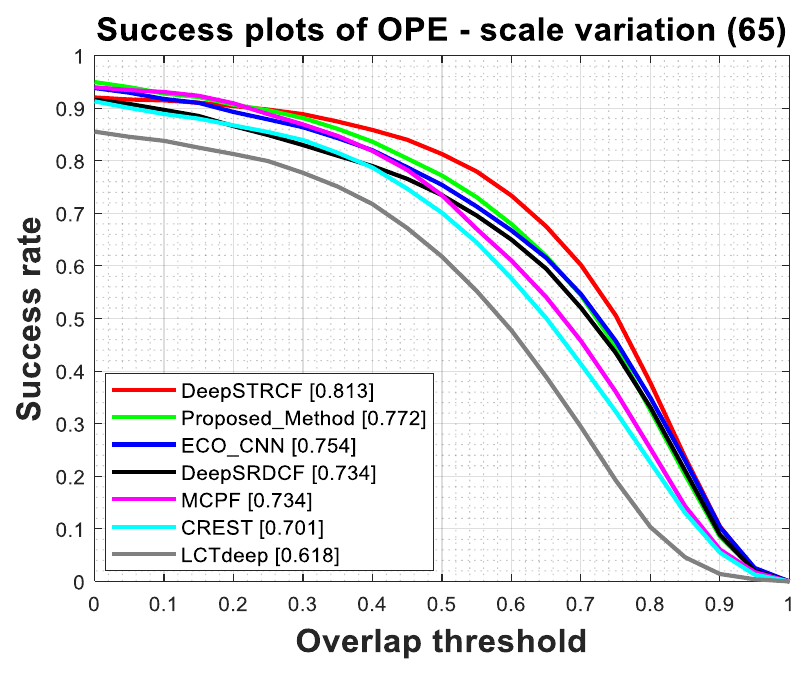}}
\hspace{0mm}
\subfigure{\includegraphics[width=2.82cm,height=2.8cm]{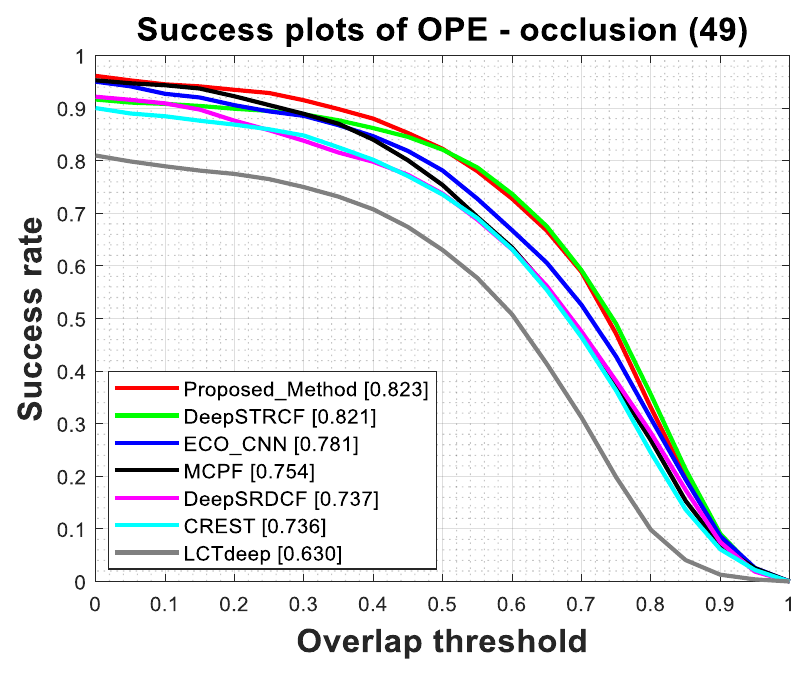}}
\vspace{0mm}
\subfigure{\includegraphics[width=2.82cm,height=2.8cm]{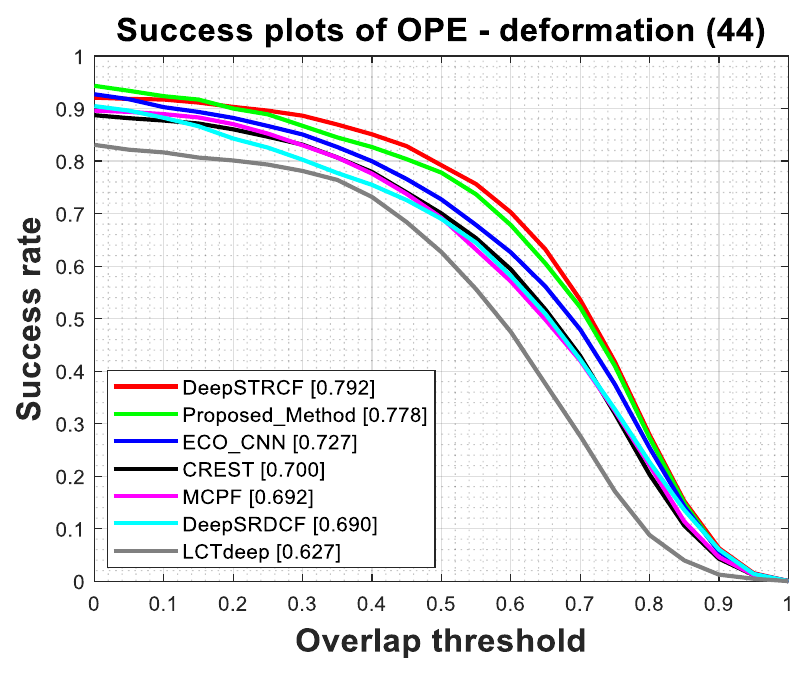}}
\hspace{0mm}
\subfigure{\includegraphics[width=2.82cm,height=2.8cm]{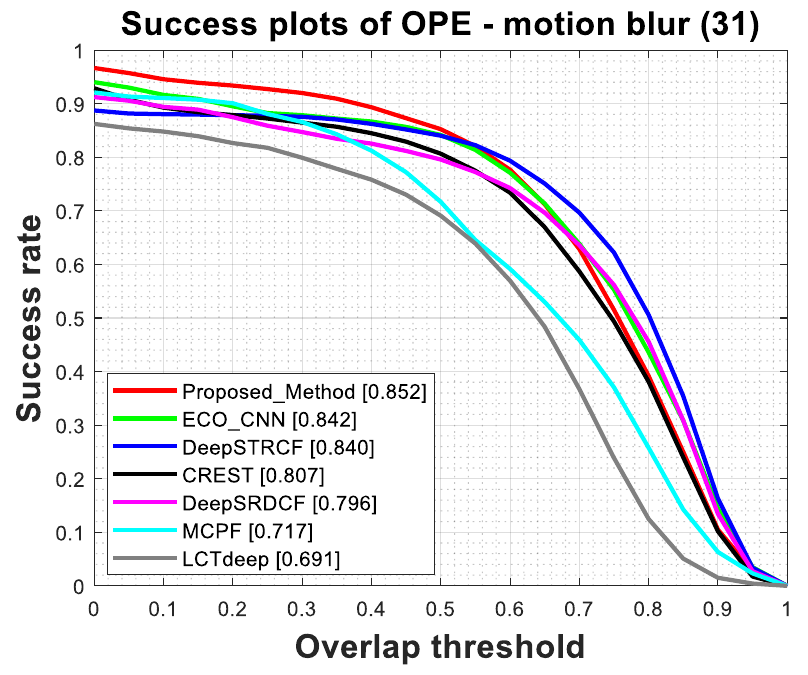}}%
\hspace{0mm}
\subfigure{\includegraphics[width=2.82cm,height=2.8cm]{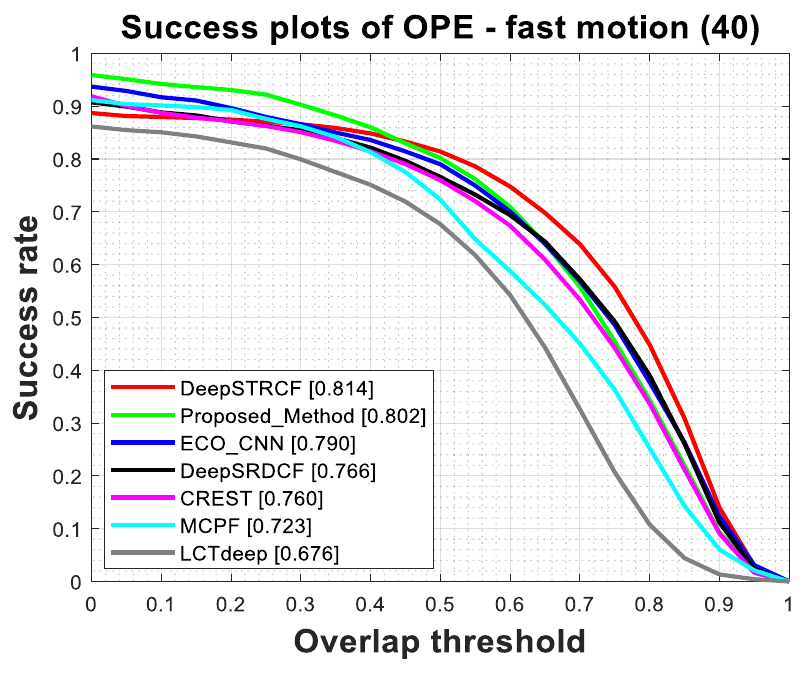}}
\hspace{0mm}
\subfigure{\includegraphics[width=2.82cm,height=2.8cm]{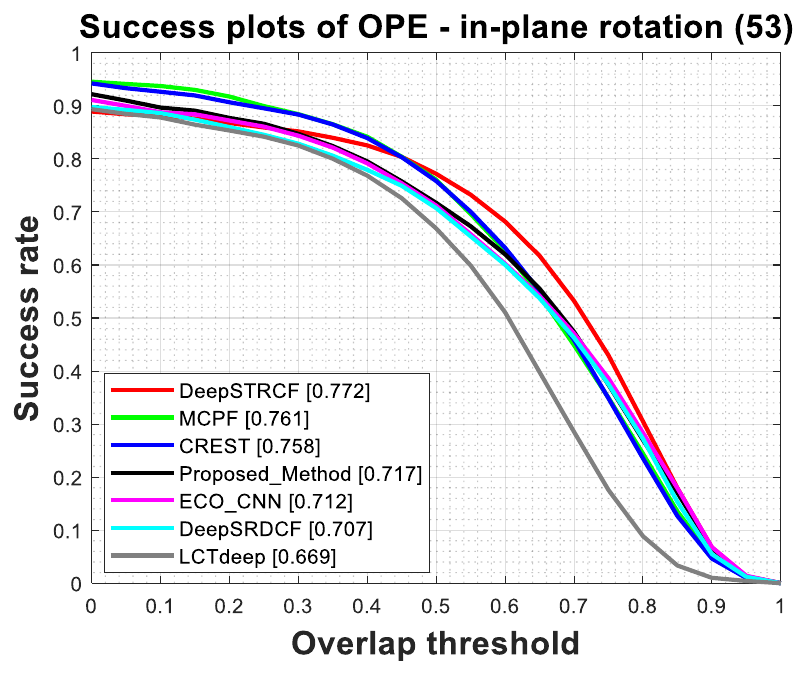}}
\vspace{-2mm}
\subfigure{\includegraphics[width=2.82cm,height=2.8cm]{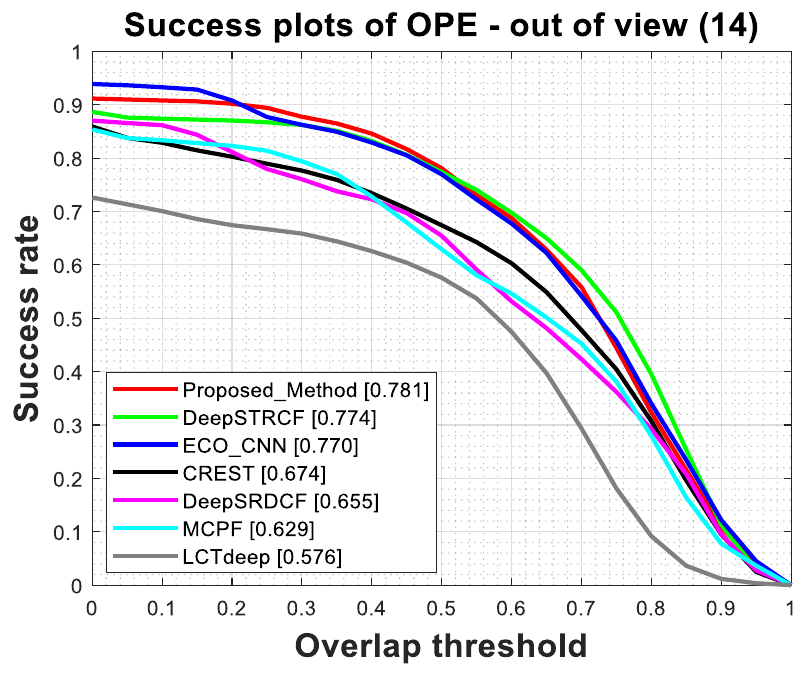}}
\hspace{0mm}
\subfigure{\includegraphics[width=2.82cm,height=2.8cm]{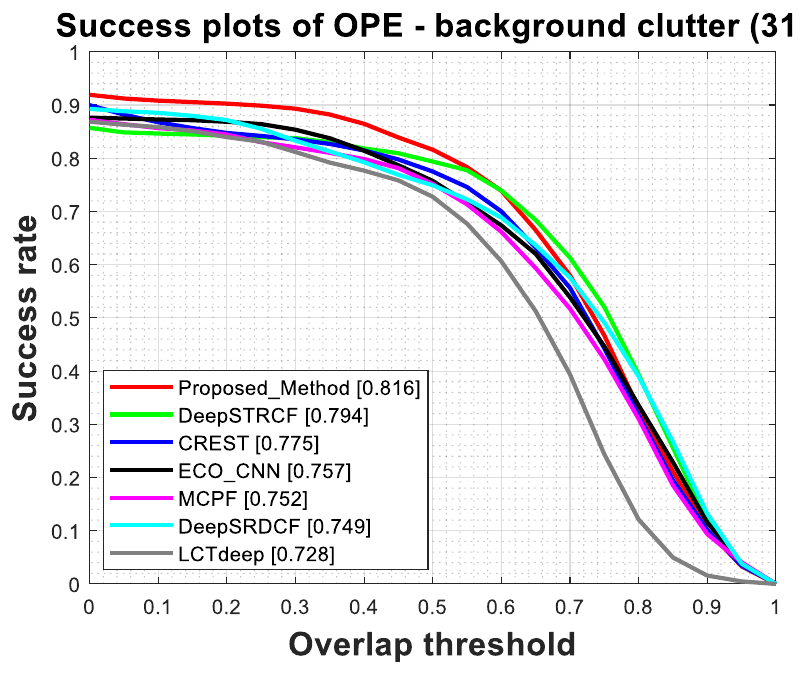}}%
\hspace{0mm}
\subfigure{\includegraphics[width=2.82cm,height=2.8cm]{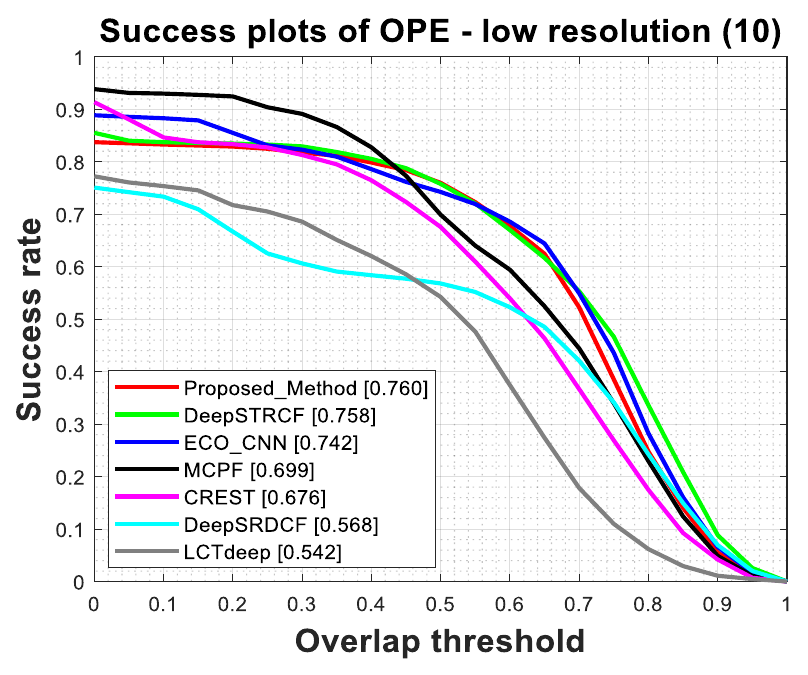}}
\caption{Attribute-based comparison of the proposed method with DCF-based trackers in terms of success rates on OTB-2015 dataset.}
\label{fig:AttComp}
\end{figure}
\indent To investigate the strengths and limitations of the proposed method, the attribute-based comparisons of DCF-based trackers are shown in Fig.~\ref{fig:AttComp}. According to this figure, the proposed method has improved the baseline tracker \cite{ECO} up to 2.3\%, 3.1\%, 1.8\%, 4.2\%, 5.1\%, 1\%, 1.2\%, 0.5\%, 1.1\%, 5.9\%, and 1.8\% on the IV, OPR, SV, OCC, DEF, MB, FM, IPR, OV, BC, and LR, respectively. These results demonstrate the proposed method can provide considerable improvements to the performance of DCF-based trackers. While the proposed tracker has moderately alleviated the baseline tracker \cite{ECO} on the IPR, MB, and OV attributes (from 0.5\% to 1.1\%), it considerably improves tracking performance against the challenging OCC, DEF, and BC attributes (from 4.2\% to 5.9\%). Compared to other DCF-based trackers, the proposed method has achieved the best performance in the presence of the challenging attributes of OCC, BC, OV, MB, and LR. For instance, the proposed method outperforms the DeepSTRCF up to 2.2\%, 0.7\%, 0.2\%, 1.2\%, and 0.2\% on the BC, OV, LR, MB, and OCC attributes, respectively. Although the proposed method significantly outperforms the baseline tracker, its performance still can be improved on the IV, OPR, SV, DEF, and FM attributes. These deficiencies comes from the inherent limitations of baseline tracker. For example, the proposed method and baseline tracker \cite{ECO} could not handle the IPR attribute and provide close results. However, they can be addressed by better representation of target through the video sequences by exploring temporal information or feature fusion strategies. \\ 
\indent The excellent performance of the proposed method arises from three primary reasons. First, a deeper insight into the knowledge about the efficient deep features for visual tracking by the comprehensive analysis. Second, simultaneous utilization of both dictionaries motives the method to improve the robustness in presence of challenging attributes but also provide an accurate localization of the target. Third, the adaptive exploitation of feature maps helps the visual tracker to have a better perception of the target and possible conditions. Hence, the appearance model of a target can be adaptively modeled by different combinations of features. It can considerably improve the dicriminative power of DCF-based methods for visual tracking. \\
\begin{figure}
\justify
\includegraphics[width=11.5cm, height=8cm]{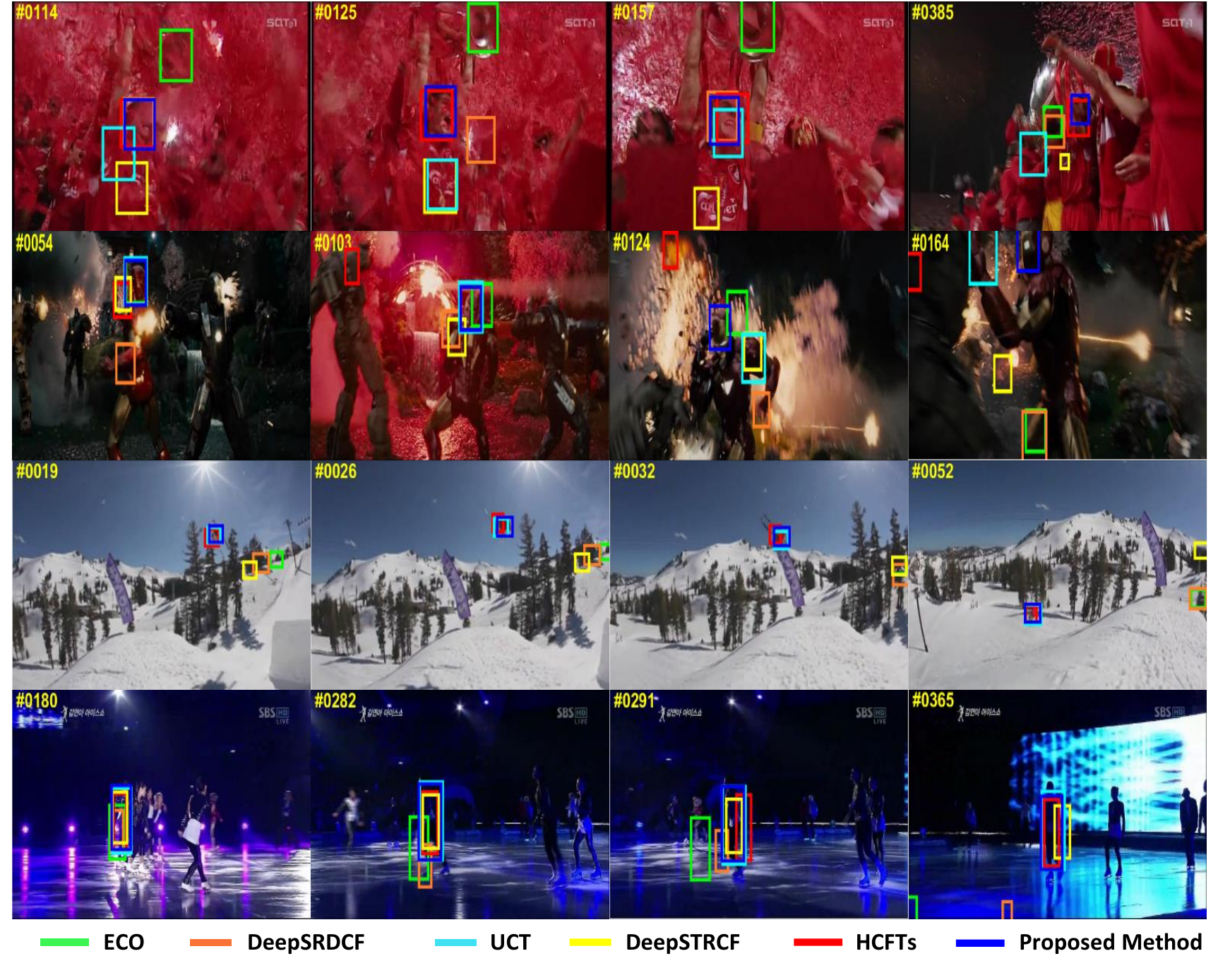}
\vspace{-2mm}
\caption{Qualitative evaluations of ECO, DeepSRDCF, UCT, HCFTs, DeepSTRCF trackers, and proposed adaptive tracker using pre-trained ResNet-50 on four challenging video sequences on the OTB-2015 dataset (namely: Soccer, Ironman, Skiing, and Skating1; from top to bottom row, respectively).}\label{fig.QualitativeComparison}
\vspace{-4mm}
\end{figure}
\indent Finally, qualitative comparisons on four challenging video sequences of the OTB-2015 dataset are shown in Fig.~\ref{fig.QualitativeComparison}. These videos include broad range of challenging attributes including the SV, OCC, FM, IPR, OPR, BC, OV, MB, IV, and LR. Also, the proposed method is compared with various visual trackers, namely ECO \cite{ECO}, DeepSTRCF \cite{STRCF}, DeepSRDCF \cite{DeepSRDCF}, HCFTs \cite{HCFTs}, and UCT \cite{UCT}. As shown in Fig. 3, the proposed adaptive method can provide both robustness and accuracy to the DCF tracking framework in the presence of real-world scenarios. However, it still can be improved by estimation of frame-based attributes and providing an approach to exploit various deep features during online tracking. 
\section{Conclusion and Future Work}
\label{sec:5_Conclusion}
The performance of four state-of-the-art pre-trained CNN models for visual tracking was analyzed. The comprehensive analysis was performed for all single and combined CNN feature maps of the CNN-based models on the well-known OTB-2013 dataset. The analysis results were used as the attribute dictionaries to adaptively select the best feature maps of CNN models in challenging scenarios. Extensive quantitative and qualitative experiments on the OTB-2015 and TC-128 visual tracking datasets demonstrated the effectiveness and generalization of the proposed method on different trackers, datasets, and models (with similar architectures) to employ the best set of deep features. \\
\indent In future work, to estimate a per-frame attribute vector, the integration of visual attribute detection methods will be explored, which can efficiently improve the robustness of visual trackers in an online manner. Although the proposed method adaptively selects the best feature maps (based on possible challenging applications), it employs a fixed set of feature maps throughout a video sequence. At the subsequent research, the proposed method will be extended on the deep learning-based methods that exploit variable deep features to construct robust appearance models of the target. This idea can effectively prevent the drift problem of visual trackers, which is caused by the contamination of a target model with background information. \\

\noindent\textbf{Acknowledgement:} This work was partly supported by a grant (No. $96013046$) from \textit{Iran National Science Foundation} (INSF).\\

\noindent\textbf{Conflict of interest} All authors declare that they have no conflict of interest. 
\vspace{-4mm}

\bibliographystyle{plainnat}
\bibliography{ref}
\newpage
\begin{appendices}
\section{Comprehensive analyses of VGG-M, VGG-16, GoogLeNet, and ResNet-50 models on the OTB-2013 dataset}\label{appendix}
In the following, the results of proposed comprehensive analysis of four pre-trained CNN models are presented. In fact, Table~\ref{SuccessDic} and Table~\ref{PrecisionDic} are summarized the best overall and attribute-based analyses of the ones in this appendix. In the following, Fig.~6 compares the overall precision and success rates of modified ECO tracker, which employs the VGG-M \cite{VGGM}, VGG-16 \cite{VGGNet}, GoogLeNet \cite{GoogLeNet}, and ResNet-50 \cite{ResNet} for feature extraction. According to Fig.~\ref{Overview}, two attribute dictionaries are formed, which allow the DCF-based trackers to exploit deep features, adaptively.
\vspace{-4mm}
\begin{center}
\begin{figure}[b!]
\justify
\includegraphics[width=0.98\linewidth]{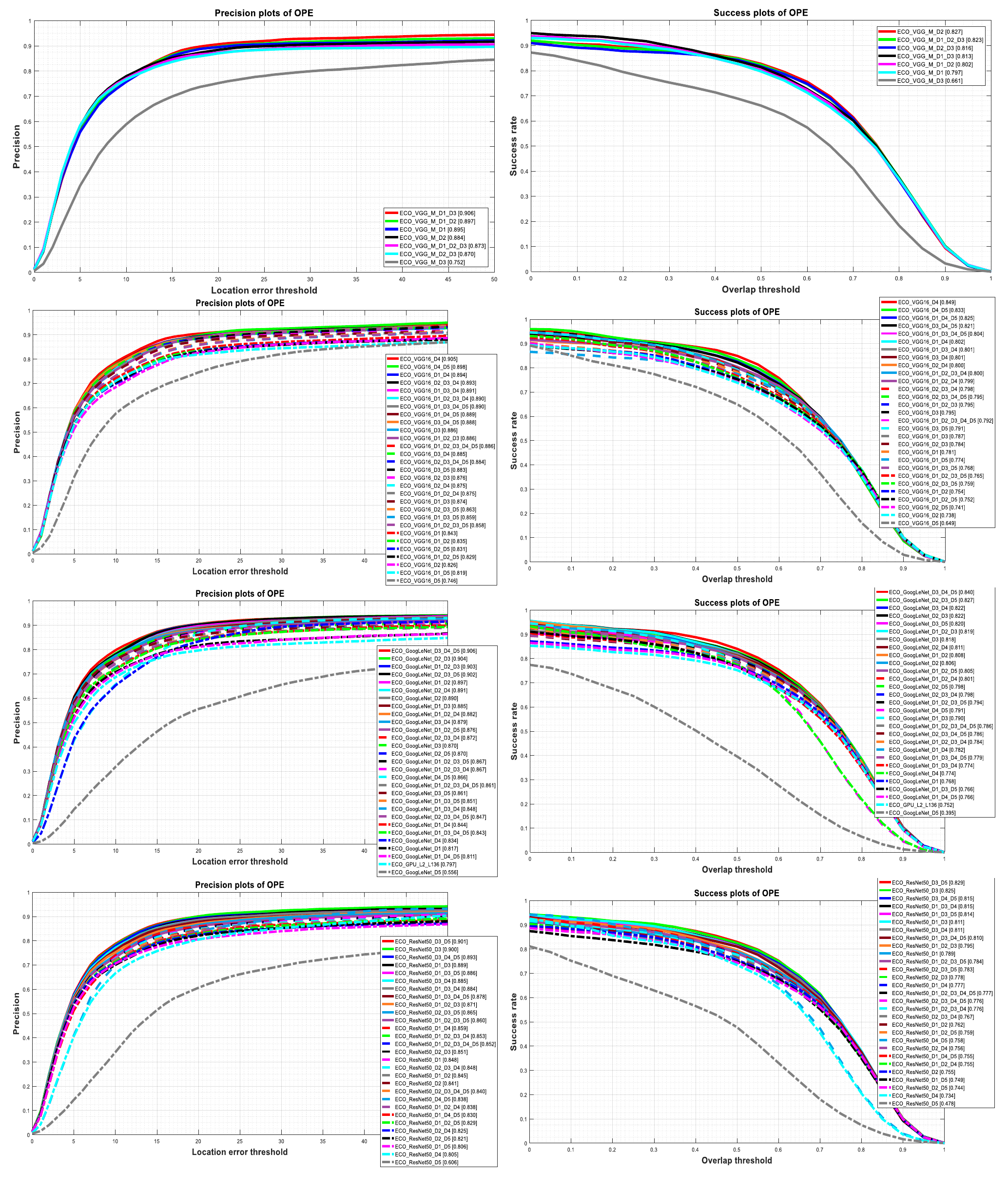}
\vspace{-2mm}
\caption{Overall precision \& success plots of VGG-M, VGG-16, GoogLeNet, and ResNet-50 models on the OTB-2013 dataset.}\label{fig.A.overall}
\vspace{-4mm}
\end{figure}
\begin{figure}
\justify
\includegraphics[width=11.5cm, height=15cm]{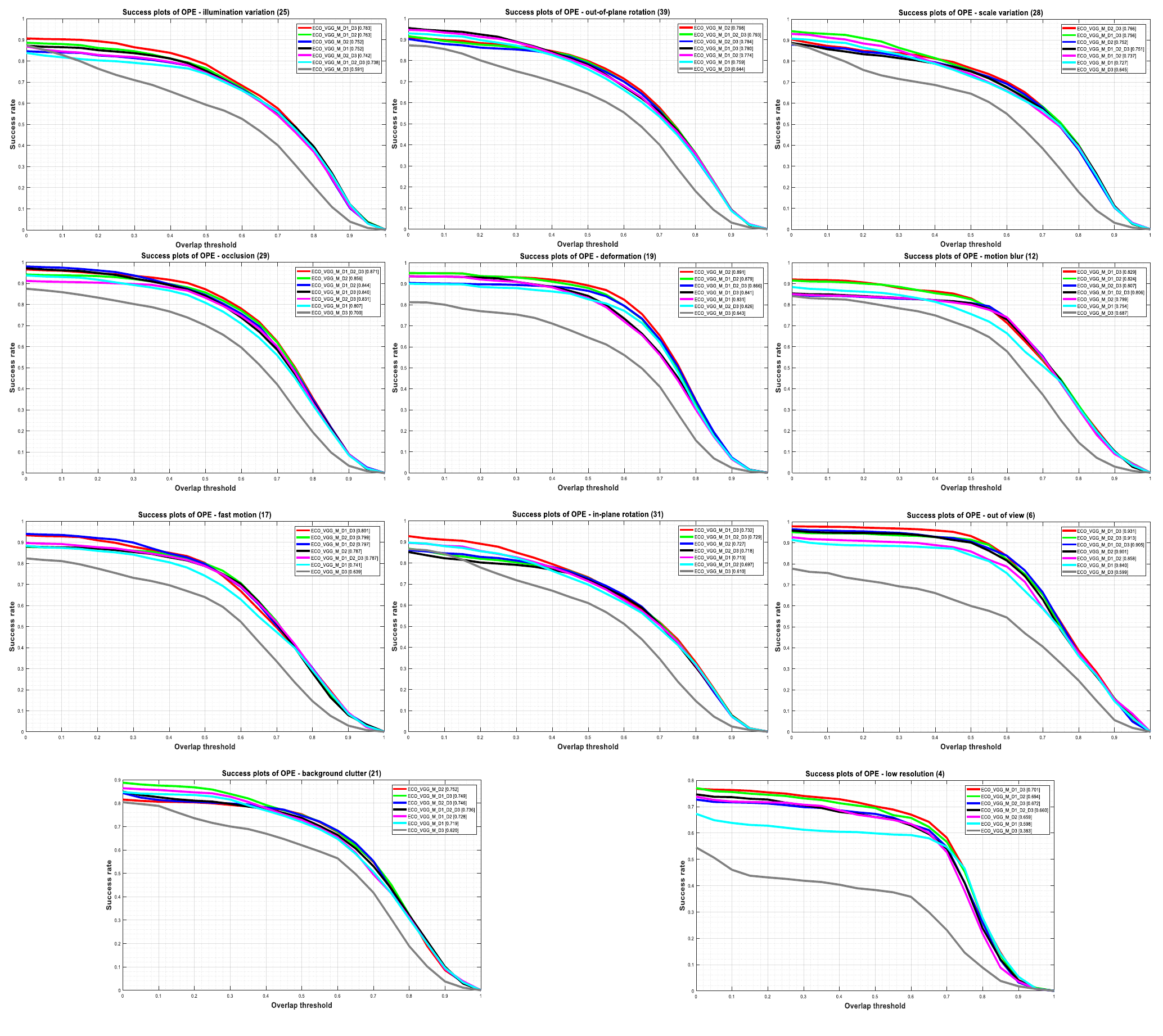}
\vspace{-2mm}
\caption{Attribute-based success plots of VGG-M model on OTB-2013 dataset.}\label{fig.A.vggm_suc}
\vspace{-4mm}
\end{figure}
\begin{table}
\caption{Success analysis results for pre-trained VGG-M model on OTB-2013 dataset.} 
\centering 
\resizebox{\textwidth}{!}{
\begin{tabular}{c |c |c |c |c |c |c |c |c |c |c |c |c} \hline \hline 
\multirow{3}{*}{Layers} & \multirow{3}{*}{Features: Resolution/Depth} & \multicolumn{11}{c}{Attributes } \\ \cline{3-13}
&  & \multicolumn{4}{c|}{Object} & \multicolumn{4}{c|}{Camera} & \multicolumn{3}{c}{Environment} \\ \cline{3-13}
&  & SV & DEF & OPR & IPR & FM & MB & LR & OV & BC & OCC & IV \\ \hline \hline
D1 & 109x109 / 96 & 0.727 & 0.831 & 0.759 & 0.713 & 0.741 & 0.754 & 0.598 & 0.840 & 0.719 & 0.807 & 0.752 \\ \hline 
D2 & 26x26 / 256 & 0.752 & 0.891 & 0.798 & 0.727 & 0.787 & 0.799 & 0.659 & 0.901 & 0.752 & 0.856 & 0.752 \\ \hline 
D3 & 13x13 / 512 & 0.645 & 0.643 & 0.644 & 0.610 & 0.639 & 0.687 & 0.383 & 0.599 & 0.620 & 0.700 & 0.591 \\ \hline 
D1, D2 & MR / 352 & 0.737 & 0.879 & 0.774 & 0.697 & 0.797 & 0.824 & 0.694 & 0.858 & 0.728 & 0.844 & 0.763 \\ \hline 
D1, D3 & MR / 608 & 0.756 & 0.841 & 0.780 & 0.732 & 0.801 & 0.829 & 0.701 & 0.931 & 0.749 & 0.840 & 0.783 \\ \hline 
D2, D3 & MR / 768 & 0.766 & 0.826 & 0.784 & 0.718 & 0.799 & 0.807 & 0.672 & 0.913 & 0.746 & 0.831 & 0.742 \\ \hline 
D1, D2, D3 & MR / 864 & 0.751 & 0.866 & 0.793 & 0.729 & 0.787 & 0.806 & 0.660 & 0.905 & 0.736 & 0.871 & 0.738 \\ \hline 
\end{tabular}
}
\label{Table.A.SucDic_VGGM}
\end{table}
\newpage
\begin{figure}
\justify
\includegraphics[width=11.5cm, height=15cm]{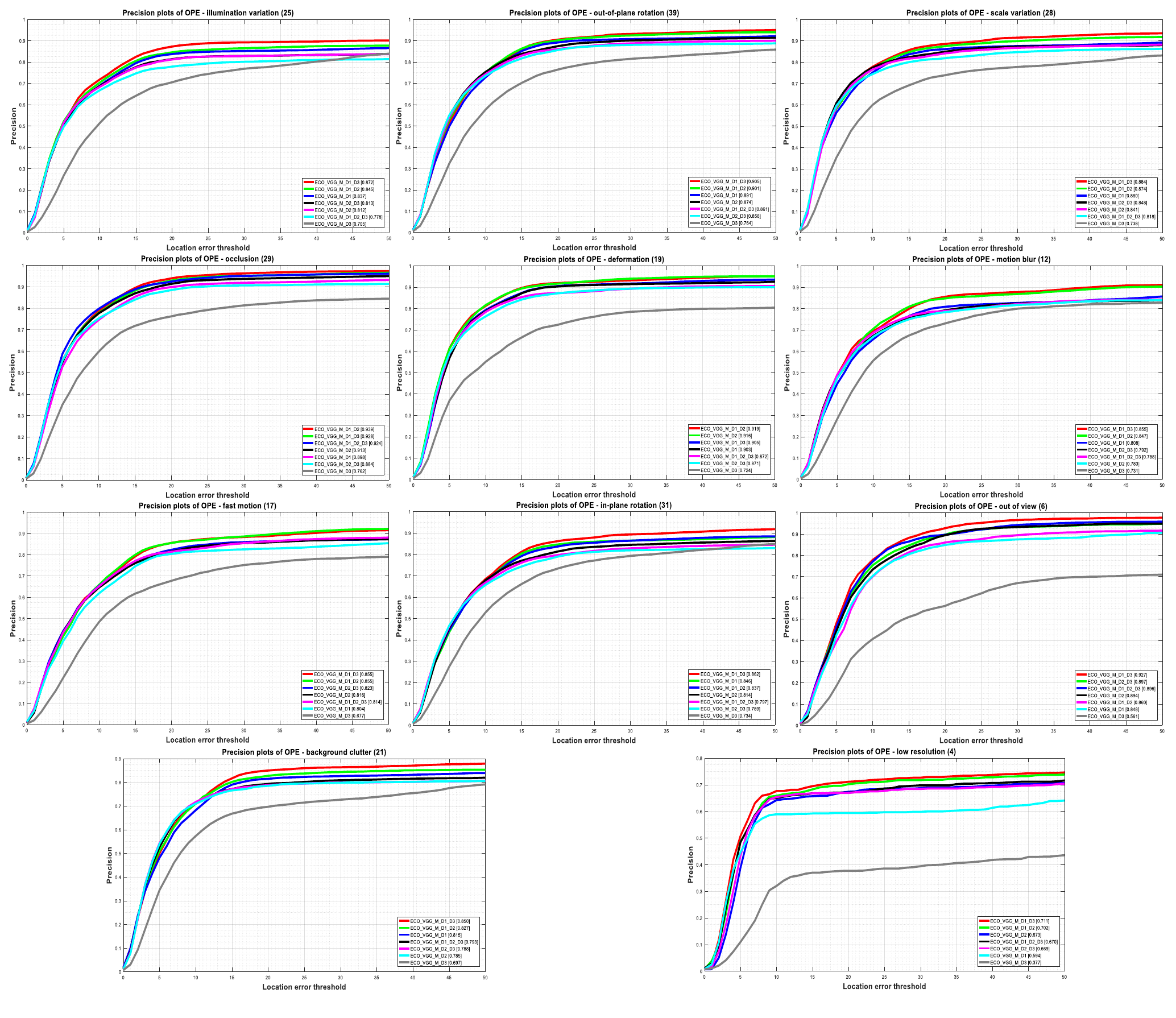}
\vspace{-2mm}
\caption{Attribute-based precision plots of VGG-M model on OTB-2013 dataset.}\label{fig.A.vggm_pre}
\vspace{-4mm}
\end{figure}
\begin{table}
\caption{Precision analysis results for pre-trained VGG-M model on OTB-2013 dataset.} 
\centering 
\resizebox{\textwidth}{!}{
\begin{tabular}{c |c |c |c |c |c |c |c |c |c |c |c |c} \hline \hline 
\multirow{3}{*}{Layers} & \multirow{3}{*}{Features: Resolution/Depth} & \multicolumn{11}{c}{Attributes } \\ \cline{3-13}
&  & \multicolumn{4}{c|}{Object} & \multicolumn{4}{c|}{Camera} & \multicolumn{3}{c}{Environment} \\ \cline{3-13}
&  & SV & DEF & OPR & IPR & FM & MB & LR & OV & BC & OCC & IV \\ \hline \hline
D1 & 109x109 / 96 & 0.860 & 0.903 & 0.891 & 0.846 & 0.804 & 0.808 & 0.594 & 0.848 & 0.815 & 0.898 & 0.837 \\ \hline 
D2 & 26x26 / 256 & 0.841 & 0.916 & 0.874 & 0.814 & 0.816 & 0.783 & 0.673 & 0.894 & 0.785 & 0.913 & 0.812 \\ \hline 
D3 & 13x13 / 512 & 0.738 & 0.724 & 0.764 & 0.734 & 0.677 & 0.731 & 0.377 & 0.561 & 0.697 & 0.762 & 0.705 \\ \hline 
D1, D2 & MR / 352 & 0.874 & 0.919 & 0.901 & 0.837 & 0.855 & 0.847 & 0.702 & 0.860 & 0.827 & 0.939 & 0.845 \\ \hline 
D1, D3 & MR / 608 & 0.884 & 0.905 & 0.905 & 0.862 & 0.855 & 0.855 & 0.711 & 0.927 & 0.850 & 0.928 & 0.872 \\ \hline 
D2, D3 & MR / 768 & 0.848 & 0.871 & 0.856 & 0.789 & 0.823 & 0.792 & 0.669 & 0.897 & 0.788 & 0.884 & 0.813 \\ \hline 
D1, D2, D3 & MR / 864 & 0.818 & 0.872 & 0.861 & 0.797 & 0.814 & 0.788 & 0.670 & 0.896 & 0.793 & 0.924 & 0.778 \\ \hline 
\end{tabular}
}
\label{Table.A.PreDic_VGGM}
\end{table}
\begin{figure}
\justify
\includegraphics[width=0.98\linewidth]{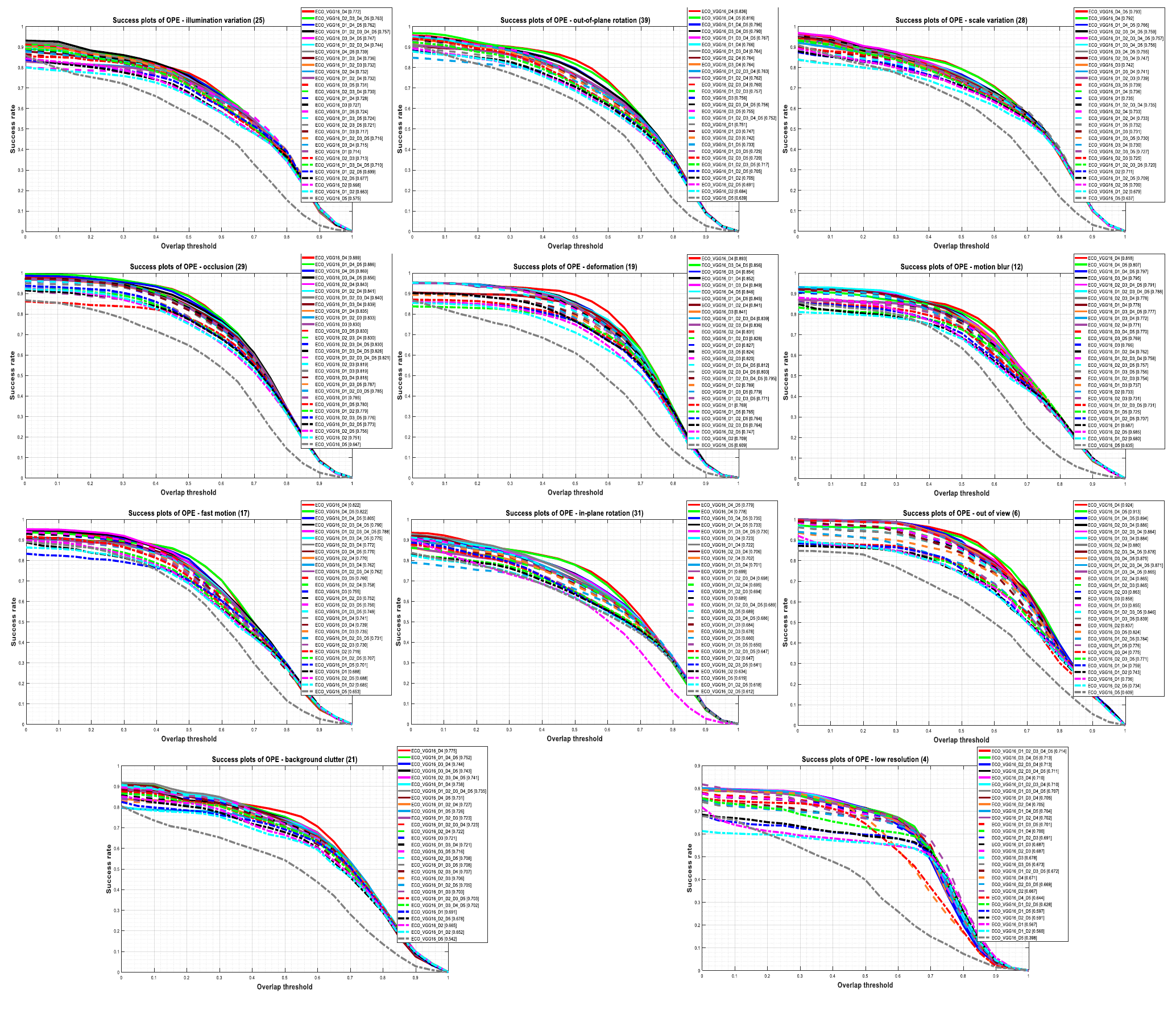}
\vspace{-2mm}
\caption{Attribute-based success plots of VGG-16 model on OTB-2013 dataset.}\label{fig.A.vgg16_suc}
\vspace{-4mm}
\end{figure}
\begin{table}
\caption{Success analysis results for pre-trained VGG-16 model on OTB-2013 dataset.} 
\centering 
\resizebox{\textwidth}{!}{
\begin{tabular}{c |c |c |c |c |c |c |c |c |c |c |c |c} \hline \hline 
\multirow{3}{*}{Layers} & \multirow{3}{*}{Features: Resolution/Depth} & \multicolumn{11}{c}{Attributes } \\ \cline{3-13}
&  & \multicolumn{4}{c|}{Object} & \multicolumn{4}{c|}{Camera} & \multicolumn{3}{c}{Environment} \\ \cline{3-13}
&  & SV & DEF & OPR & IPR & FM & MB & LR & OV & BC & OCC & IV \\ \hline \hline
D1 & 224x224 / 64 & 0.735 & 0.769 & 0.751 & 0.699 & 0.688 & 0.687 & 0.567 & 0.736 & 0.691 & 0.785 & 0.714 \\ \hline 
D2 & 112x112 / 128 & 0.711 & 0.709 & 0.684 & 0.634 & 0.719 & 0.733 & 0.667 & 0.837 & 0.665 & 0.751 & 0.668 \\ \hline 
D3 & 56x56 / 256 & 0.742 & 0.841 & 0.756 & 0.689 & 0.755 & 0.766 & 0.678 & 0.858 & 0.721 & 0.830 & 0.727 \\ \hline 
D4 & 28x28 / 512 & 0.792 & 0.893 & 0.836 & 0.778 & 0.822 & 0.818 & 0.671 & 0.924 & 0.775 & 0.889 & 0.772 \\ \hline 
D5 & 14x14 / 512 & 0.637 & 0.609 & 0.639 & 0.619 & 0.653 & 0.635 & 0.398 & 0.609 & 0.542 & 0.647 & 0.668 \\ \hline 
D1, D2 & MR / 192 & 0.679 & 0.789 & 0.705 & 0.647 & 0.685 & 0.680 & 0.560 & 0.743 & 0.652 & 0.779 & 0.663 \\ \hline 
D1, D3 & MR / 320 & 0.731 & 0.827 & 0.747 & 0.684 & 0.735 & 0.737 & 0.687 & 0.855 & 0.703 & 0.819 & 0.717 \\ \hline 
D1, D4 & MR / 576 & 0.736 & 0.852 & 0.766 & 0.722 & 0.741 & 0.778 & 0.700 & 0.759 & 0.738 & 0.835 & 0.728 \\ \hline 
D1, D5 & MR / 576 & 0.732 & 0.765 & 0.733 & 0.660 & 0.701 & 0.725 & 0.597 & 0.776 & 0.726 & 0.780 & 0.724 \\ \hline 
D2, D3 & MR / 384 & 0.725 & 0.820 & 0.742 & 0.678 & 0.730 & 0.731 & 0.687 & 0.863 & 0.706 & 0.819 & 0.713 \\ \hline 
D2, D4 & MR / 640 & 0.733 & 0.831 & 0.764 & 0.702 & 0.770 & 0.771 & 0.705 & 0.880 & 0.722 & 0.843 & 0.732 \\ \hline 
D2, D5 & MR / 640 & 0.700 & 0.747 & 0.691 & 0.612 & 0.688 & 0.685 & 0.591 & 0.734 & 0.678 & 0.756 & 0.677 \\ \hline 
D3, D4 & MR / 768 & 0.730 & 0.854 & 0.764 & 0.723 & 0.739 & 0.795 & 0.710 & 0.775 & 0.744 & 0.818 & 0.715 \\ \hline 
D3, D5 & MR / 768 & 0.739 & 0.824 & 0.755 & 0.689 & 0.760 & 0.769 & 0.673 & 0.824 & 0.716 & 0.830 & 0.731 \\ \hline 
D4, D5 & MR / 1024 & 0.793 & 0.846 & 0.816 & 0.779 & 0.822 & 0.807 & 0.644 & 0.913 & 0.731 & 0.869 & 0.739 \\ \hline 
D1, D2, D3 & MR / 448 & 0.739 & 0.828 & 0.757 & 0.694 & 0.752 & 0.754 & 0.691 & 0.865 & 0.723 & 0.833 & 0.732 \\ \hline 
D1, D2, D4 & MR / 704 & 0.733 & 0.841 & 0.762 & 0.695 & 0.758 & 0.762 & 0.702 & 0.865 & 0.727 & 0.841 & 0.732 \\ \hline 
D1, D2, D5 & MR / 704 & 0.709 & 0.764 & 0.705 & 0.618 & 0.707 & 0.707 & 0.628 & 0.784 & 0.705 & 0.773 & 0.699 \\ \hline 
D1, D3, D4 & MR / 832 & 0.741 & 0.849 & 0.764 & 0.701 & 0.762 & 0.772 & 0.705 & 0.884 & 0.721 & 0.839 & 0.736 \\ \hline 
D1, D3, D5 & MR / 832 & 0.730 & 0.779 & 0.725 & 0.650 & 0.749 & 0.756 & 0.701 & 0.839 & 0.708 & 0.787 & 0.724 \\ \hline 
D1, D4, D5 & MR / 1078 & 0.766 & 0.845 & 0.796 & 0.733 & 0.805 & 0.797 & 0.704 & 0.894 & 0.752 & 0.886 & 0.762 \\ \hline 
D2, D3, D4 & MR / 896 & 0.747 & 0.836 & 0.760 & 0.706 & 0.772 & 0.778 & 0.713 & 0.886 & 0.707 & 0.830 & 0.730 \\ \hline 
D2, D3, D5 & MR / 896 & 0.727 & 0.764 & 0.720 & 0.641 & 0.750 & 0.757 & 0.669 & 0.771 & 0.708 & 0.776 & 0.721 \\ \hline 
D3, D4, D5 & MR / 1280 & 0.755 & 0.856 & 0.790 & 0.735 & 0.770 & 0.770 & 0.713 & 0.875 & 0.743 & 0.856 & 0.747 \\ \hline 
D1, D2, D3, D4 & MR / 960 & 0.735 & 0.839 & 0.763 & 0.698 & 0.762 & 0.758 & 0.710 & 0.884 & 0.723 & 0.840 & 0.744 \\ \hline 
D1, D2, D3, D5 & MR / 960 & 0.720 & 0.771 & 0.717 & 0.647 & 0.731 & 0.731 & 0.672 & 0.846 & 0.703 & 0.785 & 0.716 \\ \hline 
D2, D3, D4, D5 & MR / 1408 & 0.759 & 0.803 & 0.756 & 0.686 & 0.790 & 0.791 & 0.711 & 0.878 & 0.741 & 0.830 & 0.763 \\ \hline 
D1, D3, D4, D5 & MR / 1344 & 0.756 & 0.812 & 0.767 & 0.730 & 0.775 & 0.777 & 0.707 & 0.865 & 0.702 & 0.828 & 0.710 \\ \hline 
D1, D2, D3, D4, D5 & MR / 1472 & 0.757 & 0.795 & 0.752 & 0.689 & 0.788 & 0.788 & 0.714 & 0.871 & 0.735 & 0.821 & 0.757 \\ \hline 
\end{tabular}
}
\label{Table.A.SucDic_VGG16}
\end{table}
\begin{figure}
\justify
\includegraphics[width=0.98\linewidth]{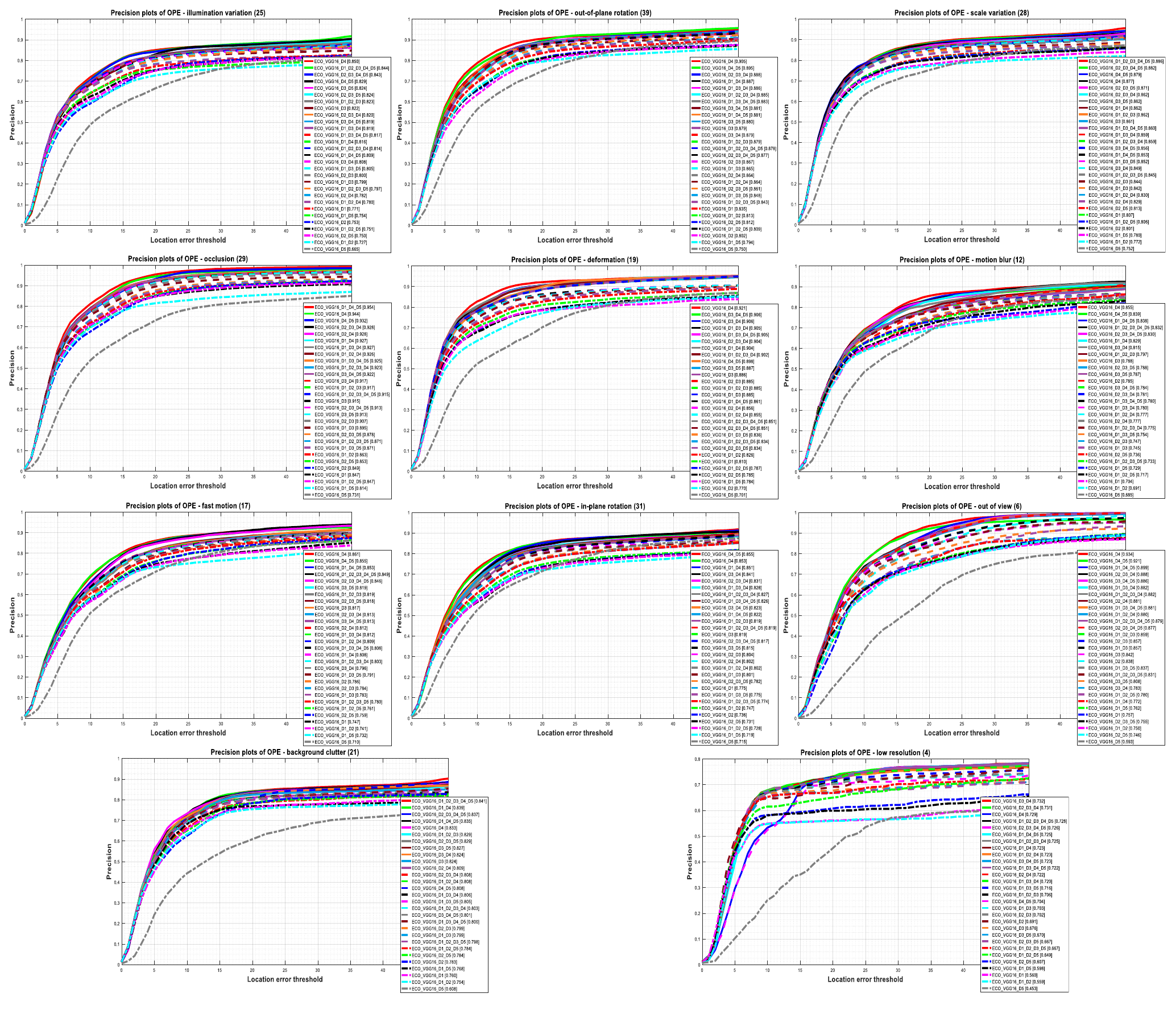}
\vspace{-2mm}
\caption{Attribute-based precision plots of VGG-16 model on OTB-2013 dataset.}\label{fig.A.vgg16_pre}
\vspace{-4mm}
\end{figure}
\begin{table}
\caption{Precision analysis results for pre-trained VGG-16 model on OTB-2013 dataset.} 
\centering 
\resizebox{\textwidth}{!}{
\begin{tabular}{c |c |c |c |c |c |c |c |c |c |c |c |c} \hline \hline 
\multirow{3}{*}{Layers} & \multirow{3}{*}{Features: Resolution/Depth} & \multicolumn{11}{c}{Attributes } \\ \cline{3-13}
&  & \multicolumn{4}{c|}{Object} & \multicolumn{4}{c|}{Camera} & \multicolumn{3}{c}{Environment} \\ \cline{3-13}
&  & SV & DEF & OPR & IPR & FM & MB & LR & OV & BC & OCC & IV \\ \hline \hline
D1 & 224x224 / 64 & 0.807 & 0.810 & 0.835 & 0.775 & 0.747 & 0.704 & 0.560 & 0.757 & 0.760 & 0.847 & 0.771 \\ \hline 
D2 & 112x112 / 128 & 0.801 & 0.770 & 0.802 & 0.736 & 0.786 & 0.785 & 0.691 & 0.838 & 0.783 & 0.849 & 0.753 \\ \hline 
D3 & 56x56 / 256 & 0.861 & 0.886 & 0.879 & 0.819 & 0.817 & 0.788 & 0.676 & 0.842 & 0.824 & 0.915 & 0.822 \\ \hline 
D4 & 28x28 / 512 & 0.877 & 0.921 & 0.905 & 0.853 & 0.861 & 0.855 & 0.729 & 0.934 & 0.833 & 0.944 & 0.850 \\ \hline 
D5 & 14x14 / 512 & 0.752 & 0.701 & 0.750 & 0.715 & 0.710 & 0.691 & 0.453 & 0.593 & 0.608 & 0.731 & 0.665 \\ \hline 
D1, D2 & MR / 192 & 0.772 & 0.826 & 0.813 & 0.747 & 0.761 & 0.685 & 0.559 & 0.750 & 0.754 & 0.863 & 0.727 \\ \hline 
D1, D3 & MR / 320 & 0.842 & 0.885 & 0.865 & 0.801 & 0.783 & 0.745 & 0.703 & 0.857 & 0.799 & 0.895 & 0.799 \\ \hline 
D1, D4 & MR / 576 & 0.862 & 0.904 & 0.887 & 0.851 & 0.808 & 0.829 & 0.723 & 0.772 & 0.839 & 0.927 & 0.816 \\ \hline 
D1, D5 & MR / 576 & 0.780 & 0.784 & 0.794 & 0.719 & 0.732 & 0.729 & 0.598 & 0.762 & 0.768 & 0.814 & 0.754 \\ \hline 
D2, D3 & MR / 384 & 0.844 & 0.885 & 0.867 & 0.804 & 0.784 & 0.747 & 0.702 & 0.857 & 0.799 & 0.907 & 0.800 \\ \hline 
D2, D4 & MR / 640 & 0.829 & 0.856 & 0.864 & 0.802 & 0.812 & 0.777 & 0.722 & 0.881 & 0.809 & 0.928 & 0.782 \\ \hline 
D2, D5 & MR / 640 & 0.813 & 0.785 & 0.812 & 0.731 & 0.759 & 0.736 & 0.607 & 0.746 & 0.784 & 0.853 & 0.750 \\ \hline 
D3, D4 & MR / 768 & 0.849 & 0.906 & 0.879 & 0.841 & 0.796 & 0.815 & 0.732 & 0.783 & 0.824 & 0.917 & 0.808 \\ \hline 
D3, D5 & MR / 768 & 0.862 & 0.887 & 0.880 & 0.815 & 0.819 & 0.787 & 0.670 & 0.808 & 0.827 & 0.913 & 0.824 \\ \hline 
D4, D5 & MR / 1024 & 0.879 & 0.898 & 0.895 & 0.855 & 0.855 & 0.839 & 0.704 & 0.921 & 0.808 & 0.932 & 0.829 \\ \hline 
D1, D2, D3 & MR / 448 & 0.862 & 0.885 & 0.879 & 0.819 & 0.819 & 0.797 & 0.706 & 0.859 & 0.829 & 0.917 & 0.823 \\ \hline 
D1, D2, D4 & MR / 704 & 0.830 & 0.855 & 0.864 & 0.802 & 0.809 & 0.777 & 0.723 & 0.880 & 0.808 & 0.926 & 0.780 \\ \hline 
D1, D2, D5 & MR / 704 & 0.806 & 0.787 & 0.809 & 0.728 & 0.761 & 0.717 & 0.649 & 0.780 & 0.784 & 0.847 & 0.751 \\ \hline 
D1, D3, D4 & MR / 832 & 0.859 & 0.905 & 0.886 & 0828 & 0.812 & 0.780 & 0.720 & 0.882 & 0.806 & 0.925 & 0.819 \\ \hline 
D1, D3, D5 & MR / 832 & 0.852 & 0.836 & 0.848 & 0.775 & 0.791 & 0.754 & 0.716 & 0.837 & 0.805 & 0.871 & 0.805 \\ \hline 
D1, D4, D5 & MR / 1078 & 0.853 & 0.861 & 0.881 & 0.822 & 0.853 & 0.838 & 0.725 & 0.899 & 0.835 & 0.954 & 0.809 \\ \hline 
D2, D3, D4 & MR / 896 & 0.862 & 0.904 & 0.888 & 0.831 & 0.813 & 0.781 & 0.731 & 0.888 & 0.808 & 0.928 & 0.820 \\ \hline 
D2, D3, D5 & MR / 896 & 0.871 & 0.834 & 0.861 & 0.782 & 0.818 & 0.788 & 0.667 & 0.755 & 0.829 & 0.878 & 0.824 \\ \hline 
D3, D4, D5 & MR / 1280 & 0.856 & 0.906 & 0.881 & 0.823 & 0.813 & 0.784 & 0.723 & 0.886 & 0.801 & 0.922 & 0.819 \\ \hline 
D1, D2, D3, D4 & MR / 960 & 0.859 & 0.902 & 0.885 & 0.827 & 0.803 & 0.775 & 0.725 & 0.882 & 0.803 & 0.923 & 0.814 \\ \hline 
D1, D2, D3, D5 & MR / 960 & 0.845 & 0.834 & 0.843 & 0.774 & 0.780 & 0.733 & 0.667 & 0.831 & 0.798 & 0.871 & 0.797 \\ \hline 
D2, D3, D4, D5 & MR / 1408 & 0.882 & 0.851 & 0.877 & 0.817 & 0.846 & 0.830 & 0.726 & 0.877 & 0.837 & 0.913 & 0.843 \\ \hline 
D1, D3, D4, D5 & MR / 1344 & 0.860 & 0.905 & 0.883 & 0.826 & 0.808 & 0.780 & 0.722 & 0.881 & 0.800 & 0.925 & 0.817 \\ \hline 
D1, D2, D3, D4, D5 & MR / 1472 & 0.886 & 0.851 & 0.878 & 0.819 & 0.849 & 0.832 & 0.728 & 0.879 & 0.841 & 0.915 & 0.844 \\ \hline 
\end{tabular}
}
\label{Table.A.PreDic_VGG16}
\end{table}
\begin{figure}
\justify
\includegraphics[width=0.98\linewidth]{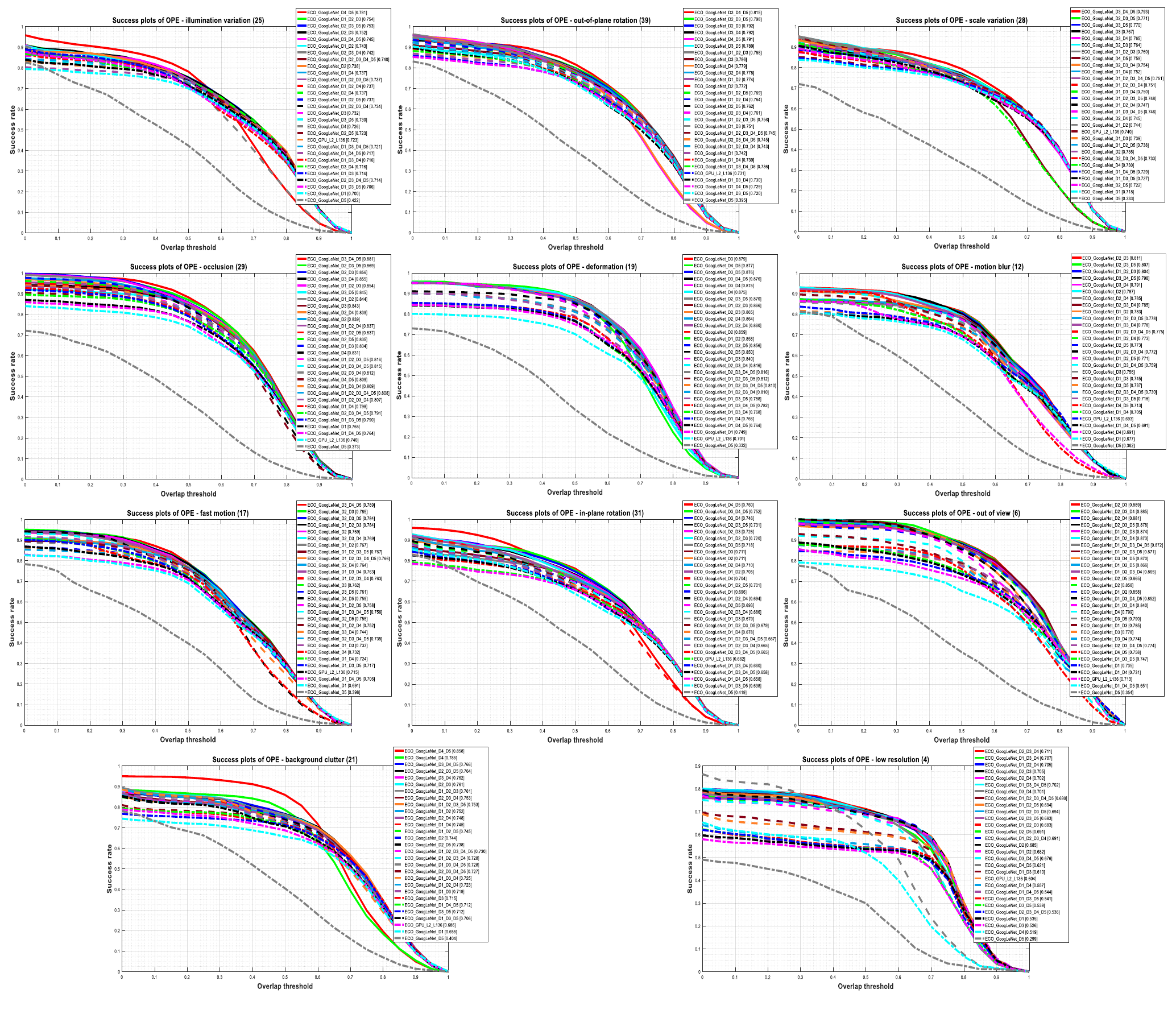}
\vspace{-2mm}
\caption{Attribute-based success plots of GoogLeNet model on OTB-2013 dataset.}\label{fig.A.googlenet_suc}
\vspace{-4mm}
\end{figure}
\begin{table}
\caption{Success analysis results for pre-trained GoogLeNet model on OTB-2013 dataset.} 
\centering 
\resizebox{\textwidth}{!}{
\begin{tabular}{c |c |c |c |c |c |c |c |c |c |c |c |c} \hline \hline 
\multirow{3}{*}{Layers} & \multirow{3}{*}{Features: Resolution/Depth} & \multicolumn{11}{c}{Attributes } \\ \cline{3-13}
&  & \multicolumn{4}{c|}{Object} & \multicolumn{4}{c|}{Camera} & \multicolumn{3}{c}{Environment} \\ \cline{3-13}
&  & SV & DEF & OPR & IPR & FM & MB & LR & OV & BC & OCC & IV \\ \hline \hline
D1 & 112x112 / 64 & 0.718 & 0.749 & 0.742 & 0.696 & 0.691 & 0.677 & 0.535 & 0.735 & 0.655 & 0.765 & 0.700 \\ \hline 
D2 & 56x56 / 192 & 0.735 & 0.859 & 0.772 & 0.711 & 0.769 & 0.787 & 0.685 & 0.858 & 0.744 & 0.839 & 0.738 \\ \hline 
D3 & 28x28 / 256 & 0.767 & 0.879 & 0.786 & 0.711 & 0.762 & 0.756 & 0.526 & 0.778 & 0.715 & 0.843 & 0.732 \\ \hline 
D4 & 14x14 / 528 & 0.730 & 0.875 & 0.779 & 0.704 & 0.732 & 0.691 & 0.519 & 0.799 & 0.785 & 0.831 & 0.726 \\ \hline 
D5 & 7x7 / 832 & 0.333 & 0.332 & 0.395 & 0.419 & 0.398 & 0.362 & 0.299 & 0.354 & 0.404 & 0.373 & 0.422 \\ \hline 
D1, D2 & MR / 256 & 0.744 & 0.858 & 0.774 & 0.705 & 0.767 & 0.783 & 0.682 & 0.858 & 0.752 & 0.844 & 0.743 \\ \hline 
D1, D3 & MR / 320 & 0.739 & 0.840 & 0.751 & 0.679 & 0.733 & 0.745 & 0.610 & 0.785 & 0.719 & 0.834 & 0.714 \\ \hline 
D1, D4 & MR / 592 & 0.752 & 0.766 & 0.739 & 0.678 & 0.724 & 0.705 & 0.557 & 0.731 & 0.745 & 0.798 & 0.737 \\ \hline 
D1, D5 & MR / 896 & 0.740 & 0.701 & 0.731 & 0.662 & 0.715 & 0.693 & 0.604 & 0.713 & 0.686 & 0.740 & 0.723 \\ \hline 
D2, D3 & MR / 448 & 0.764 & 0.865 & 0.792 & 0.726 & 0.785 & 0.811 & 0.705 & 0.889 & 0.761 & 0.856 & 0.752 \\ \hline 
D2, D4 & MR / 720 & 0.745 & 0.864 & 0.778 & 0.710 & 0.764 & 0.785 & 0.702 & 0.881 & 0.748 & 0.839 & 0.737 \\ \hline 
D2, D5 & MR / 1024 & 0.722 & 0.850 & 0.762 & 0.693 & 0.755 & 0.773 & 0.691 & 0.865 & 0.738 & 0.835 & 0.723 \\ \hline 
D3, D4 & MR / 784 & 0.765 & 0.875 & 0.792 & 0.746 & 0.744 & 0.791 & 0.701 & 0.774 & 0.762 & 0.855 & 0.716 \\ \hline 
D3, D5 & MR / 1088 & 0.770 & 0.876 & 0.789 & 0.718 & 0.761 & 0.737 & 0.539 & 0.790 & 0.712 & 0.845 & 0.730 \\ \hline 
D4, D5 & MR / 1360 & 0.759 & 0.877 & 0.791 & 0.760 & 0.759 & 0.713 & 0.621 & 0.758 & 0.858 & 0.809 & 0.781 \\ \hline 
D1, D2, D3 & MR / 512 & 0.760 & 0.866 & 0.788 & 0.720 & 0.784 & 0.804 & 0.693 & 0.874 & 0.761 & 0.854 & 0.754 \\ \hline 
D1, D2, D4 & MR / 784 & 0.747 & 0.860 & 0.764 & 0.694 & 0.752 & 0.773 & 0.705 & 0.873 & 0.723 & 0.837 & 0.737 \\ \hline 
D1, D2, D5 & MR / 1088 & 0.738 & 0.856 & 0.769 & 0.701 & 0.758 & 0.771 & 0.694 & 0.866 & 0.745 & 0.837 & 0.737 \\ \hline 
D1, D3, D4 & MR / 848 & 0.750 & 0.768 & 0.730 & 0.660 & 0.763 & 0.778 & 0.707 & 0.840 & 0.725 & 0.809 & 0.716 \\ \hline 
D1, D3, D5 & MR / 1152 & 0.727 & 0.788 & 0.720 & 0.638 & 0.717 & 0.716 & 0.541 & 0.747 & 0.706 & 0.790 & 0.706 \\ \hline 
D1, D4, D5 & MR / 1424 & 0.729 & 0.764 & 0.729 & 0.658 & 0.706 & 0.691 & 0.544 & 0.651 & 0.712 & 0.764 & 0.717 \\ \hline 
D2, D3, D4 & MR / 976 & 0.754 & 0.816 & 0.761 & 0.686 & 0.769 & 0.785 & 0.711 & 0.885 & 0.753 & 0.812 & 0.742 \\ \hline 
D2, D3, D5 & MR / 1280 & 0.771 & 0.870 & 0.798 & 0.731 & 0.784 & 0.807 & 0.693 & 0.878 & 0.764 & 0.869 & 0.753 \\ \hline 
D3, D4, D5 & MR / 1616 & 0.793 & 0.876 & 0.815 & 0.752 & 0.789 & 0.798 & 0.676 & 0.870 & 0.766 & 0.881 & 0.745 \\ \hline 
D1, D2, D3, D4 & MR / 1040 & 0.751 & 0.810 & 0.743 & 0.665 & 0.763 & 0.772 & 0.691 & 0.865 & 0.728 & 0.807 & 0.734 \\ \hline 
D1, D2, D3, D5 & MR / 1344 & 0.748 & 0.812 & 0.756 & 0.679 & 0.767 & 0.778 & 0.694 & 0.871 & 0.753 & 0.816 & 0.737 \\ \hline 
D2, D3, D4, D5 & MR / 1808 & 0.733 & 0.816 & 0.745 & 0.665 & 0.735 & 0.730 & 0.536 & 0.774 & 0.727 & 0.791 & 0.714 \\ \hline 
D1, D3, D4, D5 & MR / 1680 & 0.746 & 0.782 & 0.736 & 0.658 & 0.756 & 0.759 & 0.702 & 0.852 & 0.728 & 0.815 & 0.721 \\ \hline 
D1, D2, D3, D4, D5 & MR / 1872 & 0.751 & 0.810 & 0.745 & 0.667 & 0.766 & 0.775 & 0.699 & 0.872 & 0.730 & 0.808 & 0.740 \\ \hline 
\end{tabular}
}
\label{Table.A.SucDic_GoogLeNet}
\end{table}
\begin{figure}
\justify
\includegraphics[width=0.98\linewidth]{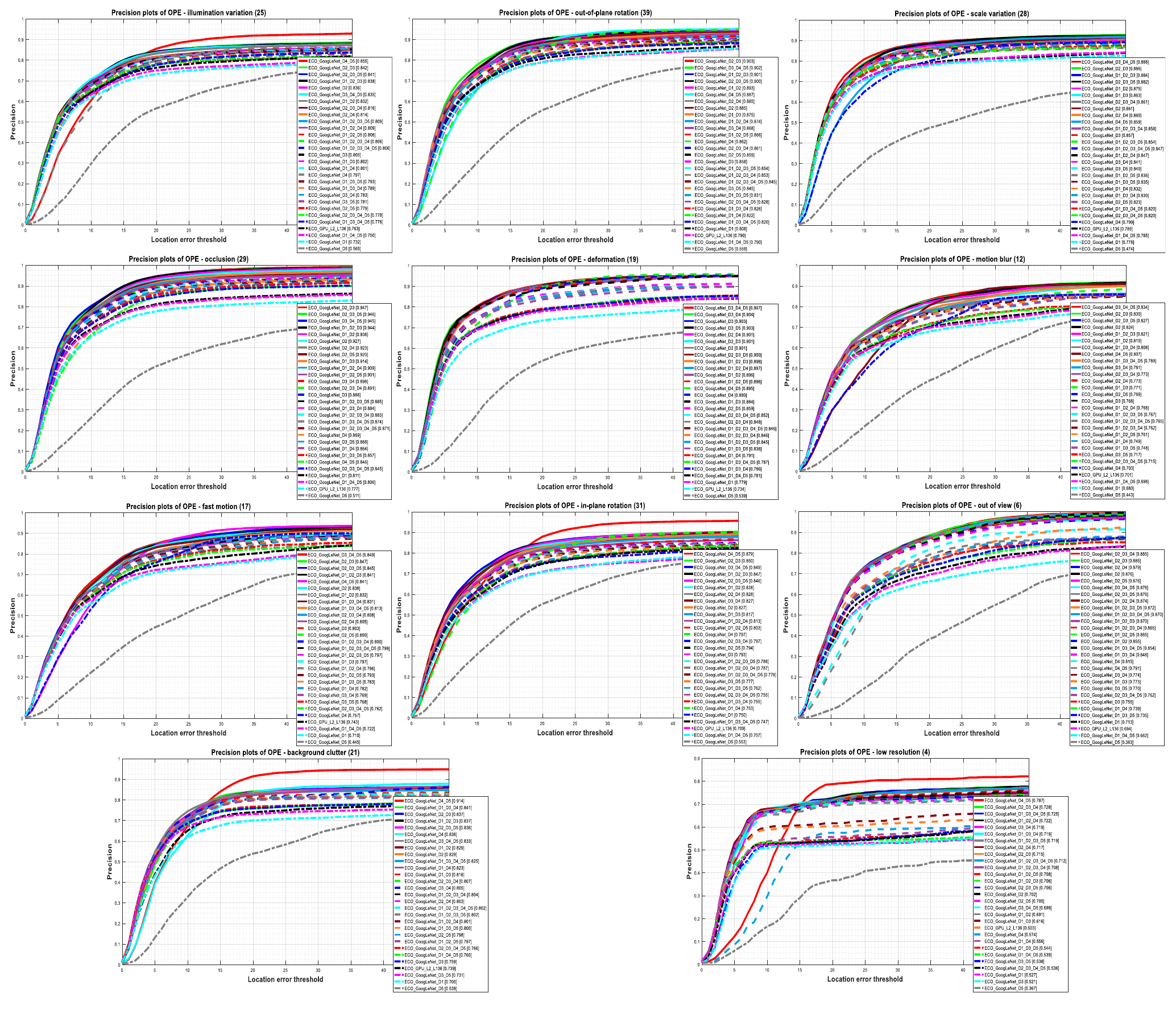}
\vspace{-2mm}
\caption{Attribute-based precision plots of GoogLeNet model on OTB-2013 dataset.}\label{fig.A.googlenet_pre}
\vspace{-4mm}
\end{figure}
\begin{table}
\caption{Precision analysis results for pre-trained GoogLeNet model on OTB-2013 dataset} 
\centering 
\resizebox{\textwidth}{!}{
\begin{tabular}{c |c |c |c |c |c |c |c |c |c |c |c |c} \hline \hline 
\multirow{3}{*}{Layers} & \multirow{3}{*}{Features: Resolution/Depth} & \multicolumn{11}{c}{Attributes } \\ \cline{3-13}
&  & \multicolumn{4}{c|}{Object} & \multicolumn{4}{c|}{Camera} & \multicolumn{3}{c}{Environment} \\ \cline{3-13}
&  & SV & DEF & OPR & IPR & FM & MB & LR & OV & BC & OCC & IV \\ \hline \hline
D1 & 112x112 / 64 & 0.778 & 0.779 & 0.808 & 0.750 & 0.710 & 0.680 & 0.527 & 0.713 & 0.700 & 0.811 & 0.732 \\ \hline 
D2 & 56x56 / 192 & 0.861 & 0.901 & 0.885 & 0.827 & 0.838 & 0.824 & 0.702 & 0.876 & 0.829 & 0.927 & 0.836 \\ \hline 
D3 & 28x28 / 256 & 0.857 & 0.903 & 0.858 & 0.793 & 0.803 & 0.768 & 0.521 & 0.755 & 0.759 & 0.888 & 0.805 \\ \hline 
D4 & 14x14 / 528 & 0.799 & 0.890 & 0.862 & 0.797 & 0.757 & 0.703 & 0.574 & 0.815 & 0.836 & 0.869 & 0.797 \\ \hline 
D5 & 7x7 / 832 & 0.474 & 0.539 & 0.559 & 0.553 & 0.445 & 0.443 & 0.367 & 0.383 & 0.539 & 0.511 & 0.565 \\ \hline 
D1, D2 & MR / 256 & 0.875 & 0.896 & 0.893 & 0.838 & 0.832 & 0.810 & 0.691 & 0.855 & 0.829 & 0.938 & 0.832 \\ \hline 
D1, D3 & MR / 320 & 0.863 & 0.884 & 0.875 & 0.817 & 0.797 & 0.771 & 0.616 & 0.773 & 0.816 & 0.914 & 0.802 \\ \hline 
D1, D4 & MR / 592 & 0.832 & 0.791 & 0.822 & 0.753 & 0.782 & 0.749 & 0.556 & 0.739 & 0.823 & 0.864 & 0.801 \\ \hline 
D1, D5 & MR / 896 & 0.789 & 0.734 & 0.790 & 0.709 & 0.743 & 0.701 & 0.603 & 0.694 & 0.739 & 0.777 & 0.763 \\ \hline 
D2, D3 & MR / 448 & 0.886 & 0.901 & 0.903 & 0.850 & 0.847 & 0.830 & 0.715 & 0.885 & 0.837 & 0.947 & 0.842 \\ \hline 
D2, D4 & MR / 720 & 0.860 & 0.901 & 0.885 & 0.828 & 0.805 & 0.773 & 0.717 & 0.879 & 0.803 & 0.923 & 0.814 \\ \hline 
D2, D5 & MR / 1024 & 0.823 & 0.859 & 0.859 & 0.794 & 0.800 & 0.769 & 0.700 & 0.876 & 0.798 & 0.920 & 0.779 \\ \hline 
D3, D4 & MR / 784 & 0.841 & 0.904 & 0.868 & 0.827 & 0.769 & 0.781 & 0.719 & 0.774 & 0.805 & 0.898 & 0.783 \\ \hline 
D3, D5 & MR / 1088 & 0.840 & 0.903 & 0.845 & 0.777 & 0.768 & 0.717 & 0.538 & 0.770 & 0.731 & 0.868 & 0.781 \\ \hline 
D4, D5 & MR / 1360 & 0.859 & 0.895 & 0.887 & 0.879 & 0.841 & 0.807 & 0.787 & 0.791 & 0.914 & 0.846 & 0.855 \\ \hline 
D1, D2, D3 & MR / 512 & 0.884 & 0.898 & 0.901 & 0.847 & 0.841 & 0.821 & 0.706 & 0.870 & 0.837 & 0.944 & 0.838 \\ \hline 
D1, D2, D4 & MR / 784 & 0.847 & 0.897 & 0.874 & 0.813 & 0.796 & 0.768 & 0.722 & 0.874 & 0.801 & 0.909 & 0.809 \\ \hline 
D1, D2, D5 & MR / 1088 & 0.836 & 0.896 & 0.866 & 0.803 & 0.793 & 0.761 & 0.708 & 0.865 & 0.797 & 0.901 & 0.806 \\ \hline 
D1, D3, D4 & MR / 848 & 0.830 & 0.786 & 0.826 & 0.755 & 0.831 & 0.808 & 0.719 & 0.848 & 0.841 & 0.884 & 0.789 \\ \hline 
D1, D3, D5 & MR / 1152 & 0.835 & 0.836 & 0.831 & 0.762 & 0.783 & 0.748 & 0.544 & 0.735 & 0.800 & 0.857 & 0.793 \\ \hline 
D1, D4, D5 & MR / 1424 & 0.785 & 0.781 & 0.790 & 0.707 & 0.722 & 0.698 & 0.539 & 0.662 & 0.760 & 0.806 & 0.750 \\ \hline 
D2, D3, D4 & MR / 976 & 0.861 & 0.848 & 0.861 & 0.797 & 0.808 & 0.773 & 0.728 & 0.885 & 0.807 & 0.891 & 0.816 \\ \hline 
D2, D3, D5 & MR / 1280 & 0.882 & 0.900 & 0.900 & 0.846 & 0.845 & 0.827 & 0.706 & 0.876 & 0.836 & 0.946 & 0.841 \\ \hline 
D3, D4, D5 & MR / 1616 & 0.888 & 0.907 & 0.902 & 0.849 & 0.849 & 0.834 & 0.699 & 0.876 & 0.833 & 0.945 & 0.835 \\ \hline 
D1, D2, D3, D4 & MR / 1040 & 0.858 & 0.846 & 0.853 & 0.787 & 0.800 & 0.762 & 0.708 & 0.865 & 0.804 & 0.883 & 0.806 \\ \hline 
D1, D2, D3, D5 & MR / 1344 & 0.854 & 0.845 & 0.854 & 0.788 & 0.797 & 0.767 & 0.719 & 0.872 & 0.802 & 0.885 & 0.809 \\ \hline 
D2, D3, D4, D5 & MR / 1808 & 0.820 & 0.852 & 0.828 & 0.755 & 0.762 & 0.715 & 0.536 & 0.762 & 0.766 & 0.845 & 0.778 \\ \hline 
D1, D3, D4, D5 & MR / 1680 & 0.820 & 0.787 & 0.820 & 0.747 & 0.813 & 0.789 & 0.725 & 0.854 & 0.825 & 0.874 & 0.776 \\ \hline 
D1, D2, D3, D4, D5 & MR / 1872 & 0.847 & 0.846 & 0.845 & 0.778 & 0.799 & 0.765 & 0.712 & 0.870 & 0.802 & 0.871 & 0.806 \\ \hline 
\end{tabular}
}
\label{Table.A.PreDic_GoogLeNet}
\end{table}
\begin{figure}
\justify
\includegraphics[width=0.98\linewidth]{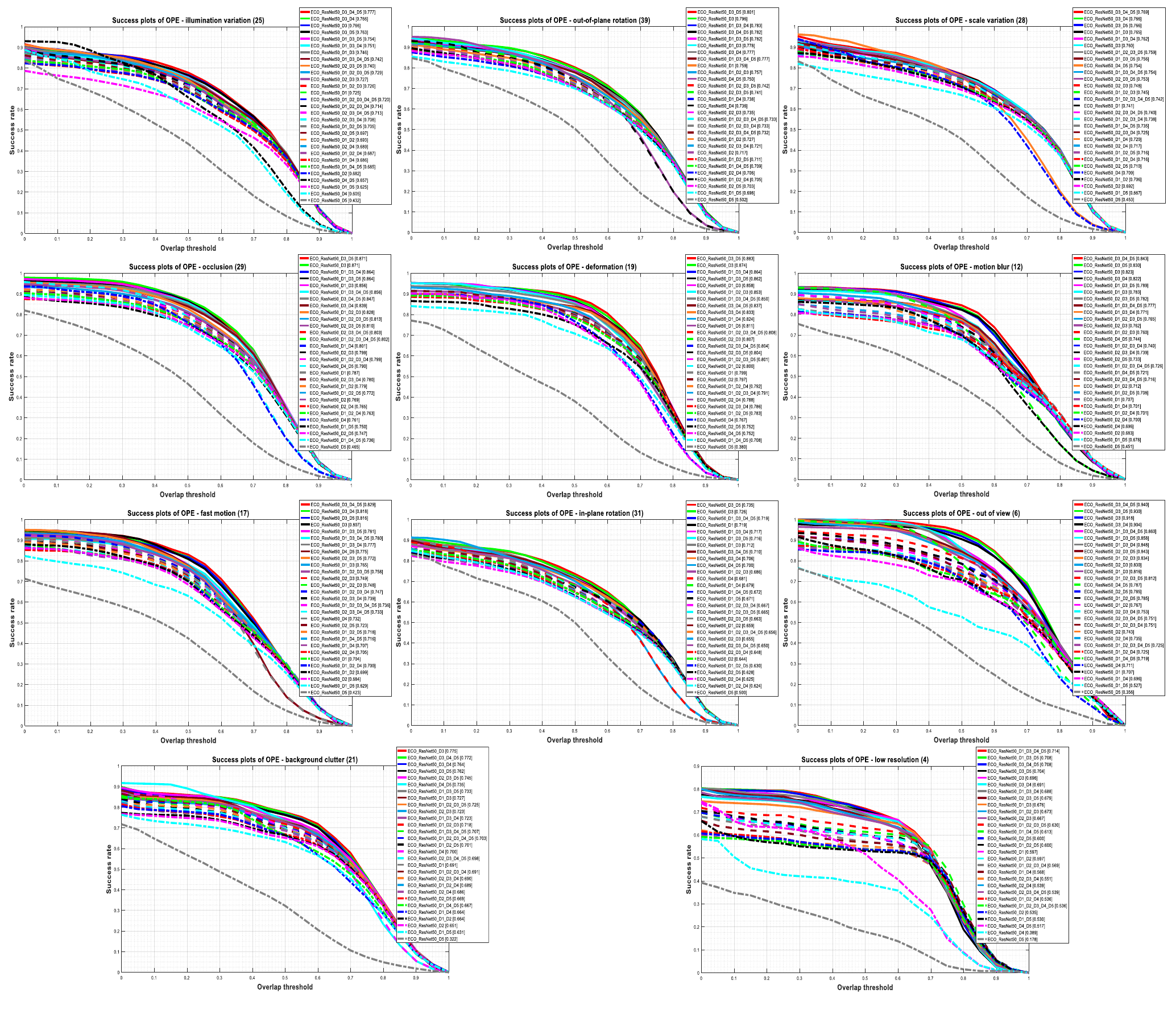}
\vspace{-2mm}
\caption{Attribute-based success plots of ResNet-50 model on OTB-2013 dataset.}\label{fig.A.resnet_suc}
\vspace{-4mm}
\end{figure}
\begin{table}
\caption{Success analysis results for pre-trained ResNet-50 model on OTB-2013 dataset.} 
\centering 
\resizebox{\textwidth}{!}{
\begin{tabular}{c |c |c |c |c |c |c |c |c |c |c |c |c} \hline \hline 
\multirow{3}{*}{Layers} & \multirow{3}{*}{Features: Resolution/Depth} & \multicolumn{11}{c}{Attributes } \\ \cline{3-13}
&  & \multicolumn{4}{c|}{Object} & \multicolumn{4}{c|}{Camera} & \multicolumn{3}{c}{Environment} \\ \cline{3-13}
&  & SV & DEF & OPR & IPR & FM & MB & LR & OV & BC & OCC & IV \\ \hline \hline
D1 & 112x112 / 64 & 0.741 & 0.799 & 0.759 & 0.719 & 0.704 & 0.707 & 0.597 & 0.707 & 0.691 & 0.787 & 0.725 \\ \hline 
D2 & 56x56 / 256 & 0.692 & 0.797 & 0.717 & 0.644 & 0.684 & 0.683 & 0.535 & 0.743 & 0.651 & 0.769 & 0.682 \\ \hline 
D3 & 28x28 / 512 & 0.760 & 0.874 & 0.796 & 0.726 & 0.807 & 0.823 & 0.696 & 0.918 & 0.775 & 0.871 & 0.766 \\ \hline 
D4 & 14x14 / 1024 & 0.709 & 0.767 & 0.736 & 0.681 & 0.732 & 0.696 & 0.389 & 0.711 & 0.700 & 0.761 & 0.605 \\ \hline 
D5 & 7x7 / 2048 & 0.453 & 0.380 & 0.502 & 0.500 & 0.423 & 0.451 & 0.178 & 0.356 & 0.322 & 0.465 & 0.432 \\ \hline 
D1, D2 & MR / 320 & 0.706 & 0.800 & 0.727 & 0.659 & 0.699 & 0.712 & 0.597 & 0.767 & 0.664 & 0.779 & 0.693 \\ \hline 
D1, D3 & MR / 576 & 0.765 & 0.858 & 0.778 & 0.712 & 0.765 & 0.783 & 0.676 & 0.816 & 0.727 & 0.856 & 0.746 \\ \hline 
D1, D4 & MR / 1088 & 0.720 & 0.824 & 0.738 & 0.679 & 0.707 & 0.701 & 0.568 & 0.696 & 0.664 & 0.801 & 0.686 \\ \hline 
D1, D5 & MR / 2112 & 0.667 & 0.811 & 0.698 & 0.671 & 0.629 & 0.678 & 0.530 & 0.527 & 0.631 & 0.750 & 0.625 \\ \hline 
D2, D3 & MR / 768 & 0.748 & 0.807 & 0.735 & 0.655 & 0.749 & 0.762 & 0.667 & 0.830 & 0.723 & 0.799 & 0.727 \\ \hline 
D2, D4 & MR / 1280 & 0.717 & 0.788 & 0.706 & 0.625 & 0.705 & 0.700 & 0.539 & 0.735 & 0.686 & 0.765 & 0.689 \\ \hline 
D2, D5 & MR / 2304 & 0.710 & 0.752 & 0.703 & 0.628 & 0.723 & 0.733 & 0.600 & 0.785 & 0.669 & 0.747 & 0.697 \\ \hline 
D3, D4 & MR / 1536 & 0.766 & 0.833 & 0.777 & 0.706 & 0.818 & 0.822 & 0.691 & 0.904 & 0.764 & 0.839 & 0.766 \\ \hline 
D3, D5 & MR / 2560 & 0.766 & 0.883 & 0.801 & 0.735 & 0.816 & 0.830 & 0.704 & 0.930 & 0.762 & 0.871 & 0.763 \\ \hline 
D4, D5 & MR / 3072 & 0.754 & 0.752 & 0.750 & 0.700 & 0.775 & 0.744 & 0.517 & 0.787 & 0.735 & 0.790 & 0.657 \\ \hline 
D1, D2, D3 & MR / 832 & 0.745 & 0.853 & 0.757 & 0.686 & 0.748 & 0.760 & 0.673 & 0.834 & 0.718 & 0.828 & 0.726 \\ \hline 
D1, D2, D4 & MR / 1344 & 0.716 & 0.792 & 0.705 & 0.624 & 0.700 & 0.701 & 0.536 & 0.725 & 0.689 & 0.763 & 0.687 \\ \hline 
D1, D2, D5 & MR / 2368 & 0.716 & 0.783 & 0.711 & 0.630 & 0.718 & 0.708 & 0.600 & 0.785 & 0.701 & 0.772 & 0.705 \\ \hline 
D1, D3, D4 & MR / 1600 & 0.762 & 0.864 & 0.783 & 0.717 & 0.777 & 0.771 & 0.688 & 0.848 & 0.723 & 0.864 & 0.751 \\ \hline 
D1, D3, D5 & MR / 2624 & 0.756 & 0.862 & 0.782 & 0.716 & 0.781 & 0.789 & 0.708 & 0.858 & 0.733 & 0.864 & 0.754 \\ \hline 
D1, D4, D5 & MR / 3136 & 0.735 & 0.708 & 0.709 & 0.672 & 0.716 & 0.721 & 0.613 & 0.719 & 0.667 & 0.736 & 0.685 \\ \hline 
D2, D3, D4 & MR / 1792 & 0.725 & 0.786 & 0.721 & 0.648 & 0.739 & 0.739 & 0.551 & 0.753 & 0.690 & 0.780 & 0.708 \\ \hline 
D2, D3, D5 & MR / 2816 & 0.753 & 0.804 & 0.741 & 0.663 & 0.758 & 0.782 & 0.679 & 0.843 & 0.745 & 0.810 & 0.740 \\ \hline 
D3, D4, D5 & MR / 3584 & 0.769 & 0.837 & 0.782 & 0.710 & 0.829 & 0.843 & 0.708 & 0.940 & 0.772 & 0.847 & 0.777 \\ \hline 
D1, D2, D3, D4 & MR / 1856 & 0.738 & 0.791 & 0.733 & 0.667 & 0.747 & 0.740 & 0.569 & 0.751 & 0.691 & 0.799 & 0.714 \\ \hline 
D1, D2, D3, D5 & MR / 2880 & 0.759 & 0.801 & 0.742 & 0.665 & 0.758 & 0.765 & 0.630 & 0.812 & 0.725 & 0.813 & 0.729 \\ \hline 
D2, D3, D4, D5 & MR / 3840 & 0.740 & 0.804 & 0.732 & 0.650 & 0.733 & 0.716 & 0.539 & 0.751 & 0.698 & 0.803 & 0.713 \\ \hline 
D1, D3, D4, D5 & MR / 3648 & 0.754 & 0.850 & 0.777 & 0.719 & 0.780 & 0.777 & 0.714 & 0.860 & 0.707 & 0.856 & 0.742 \\ \hline 
D1, D2, D3, D4, D5 & MR / 3904 & 0.742 & 0.808 & 0.733 & 0.656 & 0.736 & 0.726 & 0.536 & 0.725 & 0.703 & 0.802 & 0.720 \\ \hline 
\end{tabular}
}
\label{Table.A.SucDic_ResNet}
\end{table}
\begin{figure}
\justify
\includegraphics[width=0.98\linewidth]{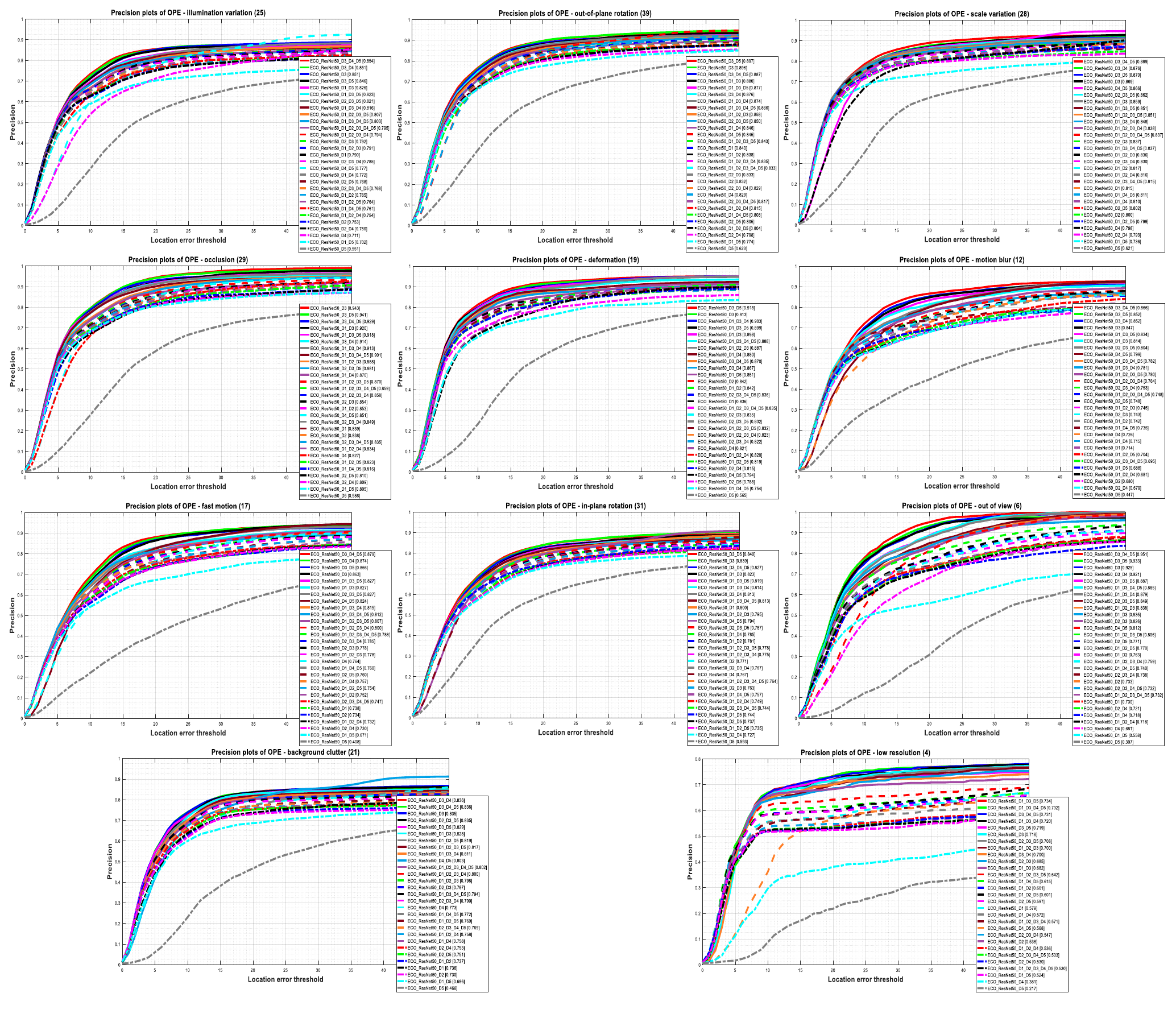}
\vspace{-2mm}
\caption{Attribute-based precision plots of ResNet-50 model on OTB-2013 dataset.}\label{fig.A.resnet_pre}
\vspace{-4mm}
\end{figure}
\begin{table}
\caption{Precision analysis results for pre-trained ResNet-50 model on OTB-2013 dataset.} 
\centering 
\resizebox{\textwidth}{!}{
\begin{tabular}{c |c |c |c |c |c |c |c |c |c |c |c |c} \hline \hline 
\multirow{3}{*}{Layers} & \multirow{3}{*}{Features: Resolution/Depth} & \multicolumn{11}{c}{Attributes } \\ \cline{3-13}
&  & \multicolumn{4}{c|}{Object} & \multicolumn{4}{c|}{Camera} & \multicolumn{3}{c}{Environment} \\ \cline{3-13}
&  & SV & DEF & OPR & IPR & FM & MB & LR & OV & BC & OCC & IV \\ \hline \hline
D1 & 112x112 / 64 & 0.815 & 0.836 & 0.840 & 0.800 & 0.738 & 0.714 & 0.579 & 0.730 & 0.736 & 0.839 & 0.790 \\ \hline 
D2 & 56x56 / 256 & 0.800 & 0.842 & 0.832 & 0.771 & 0.734 & 0.680 & 0.538 & 0.733 & 0.730 & 0.838 & 0.753 \\ \hline 
D3 & 28x28 / 512 & 0.869 & 0.913 & 0.896 & 0.839 & 0.863 & 0.847 & 0.716 & 0.925 & 0.835 & 0.943 & 0.851 \\ \hline 
D4 & 14x14 / 1024 & 0.798 & 0.821 & 0.829 & 0.767 & 0.764 & 0.726 & 0.381 & 0.681 & 0.773 & 0.827 & 0.711 \\ \hline 
D5 & 7x7 / 2048 & 0.621 & 0.565 & 0.623 & 0.593 & 0.408 & 0.447 & 0.217 & 0.307 & 0.466 & 0.586 & 0.551 \\ \hline 
D1, D2 & MR / 320 & 0.817 & 0.842 & 0.838 & 0.781 & 0.752 & 0.742 & 0.601 & 0.763 & 0.737 & 0.853 & 0.765 \\ \hline 
D1, D3 & MR / 576 & 0.859 & 0.898 & 0.880 & 0.823 & 0.827 & 0.814 & 0.682 & 0.835 & 0.828 & 0.920 & 0.826 \\ \hline 
D1, D4 & MR / 1088 & 0.810 & 0.880 & 0.846 & 0.785 & 0.757 & 0.715 & 0.572 & 0.718 & 0.758 & 0.870 & 0.772 \\ \hline 
D1, D5 & MR / 2112 & 0.736 & 0.851 & 0.774 & 0.744 & 0.671 & 0.688 & 0.524 & 0.558 & 0.686 & 0.805 & 0.702 \\ \hline 
D2, D3 & MR / 768 & 0.837 & 0.835 & 0.833 & 0.763 & 0.778 & 0.743 & 0.685 & 0.826 & 0.797 & 0.854 & 0.792 \\ \hline 
D2, D4 & MR / 1280 & 0.793 & 0.815 & 0.798 & 0.727 & 0.730 & 0.679 & 0.530 & 0.721 & 0.753 & 0.809 & 0.750 \\ \hline 
D2, D5 & MR / 2304 & 0.802 & 0.788 & 0.805 & 0.737 & 0.760 & 0.748 & 0.597 & 0.771 & 0.751 & 0.810 & 0.768 \\ \hline 
D3, D4 & MR / 1536 & 0.876 & 0.867 & 0.876 & 0.813 & 0.874 & 0.852 & 0.700 & 0.921 & 0.836 & 0.914 & 0.851 \\ \hline 
D3, D5 & MR / 2560 & 0.870 & 0.918 & 0.897 & 0.840 & 0.866 & 0.852 & 0.719 & 0.933 & 0.829 & 0.941 & 0.846 \\ \hline 
D4, D5 & MR / 3072 & 0.866 & 0.794 & 0.845 & 0.794 & 0.824 & 0.799 & 0.568 & 0.812 & 0.803 & 0.851 & 0.777 \\ \hline 
D1, D2, D3 & MR / 832 & 0.836 & 0.887 & 0.858 & 0.795 & 0.778 & 0.745 & 0.700 & 0.838 & 0.798 & 0.888 & 0.791 \\ \hline 
D1, D2, D4 & MR / 1344 & 0.816 & 0.820 & 0.815 & 0.749 & 0.732 & 0.681 & 0.536 & 0.718 & 0.758 & 0.834 & 0.754 \\ \hline 
D1, D2, D5 & MR / 2368 & 0.799 & 0.819 & 0.804 & 0.735 & 0.754 & 0.704 & 0.601 & 0.770 & 0.769 & 0.823 & 0.764 \\ \hline 
D1, D3, D4 & MR / 1600 & 0.848 & 0.903 & 0.874 & 0.814 & 0.815 & 0.781 & 0.720 & 0.879 & 0.811 & 0.913 & 0.816 \\ \hline 
D1, D3, D5 & MR / 2624 & 0.851 & 0.899 & 0.877 & 0.819 & 0.827 & 0.834 & 0.734 & 0.887 & 0.819 & 0.918 & 0.823 \\ \hline 
D1, D4, D5 & MR / 3136 & 0.811 & 0.754 & 0.808 & 0.757 & 0.760 & 0.735 & 0.615 & 0.740 & 0.772 & 0.816 & 0.761 \\ \hline 
D2, D3, D4 & MR / 1792 & 0.830 & 0.822 & 0.829 & 0.767 & 0.785 & 0.753 & 0.547 & 0.738 & 0.790 & 0.849 & 0.785 \\ \hline 
D2, D3, D5 & MR / 2816 & 0.862 & 0.832 & 0.850 & 0.787 & 0.827 & 0.804 & 0.708 & 0.849 & 0.835 & 0.881 & 0.821 \\ \hline 
D3, D4, D5 & MR / 3584 & 0.889 & 0.870 & 0.887 & 0.827 & 0.879 & 0.866 & 0.731 & 0.951 & 0.836 & 0.929 & 0.854 \\ \hline 
D1, D2, D3, D4 & MR / 1856 & 0.838 & 0.823 & 0.835 & 0.775 & 0.800 & 0.764 & 0.571 & 0.759 & 0.800 & 0.858 & 0.794 \\ \hline 
D1, D2, D3, D5 & MR / 2880 & 0.851 & 0.832 & 0.843 & 0.778 & 0.807 & 0.780 & 0.642 & 0.806 & 0.817 & 0.870 & 0.807 \\ \hline 
D2, D3, D4, D5 & MR / 3840 & 0.815 & 0.836 & 0.817 & 0.744 & 0.747 & 0.695 & 0.533 & 0.732 & 0.769 & 0.835 & 0.768 \\ \hline 
D1, D3, D4, D5 & MR / 3648 & 0.837 & 0.888 & 0.866 & 0.813 & 0.812 & 0.782 & 0.732 & 0.885 & 0.794 & 0.901 & 0.803 \\ \hline 
D1, D2, D3, D4, D5 & MR / 3904 & 0.837 & 0.835 & 0.833 & 0.764 & 0.788 & 0.748 & 0.530 & 0.732 & 0.802 & 0.859 & 0.795 \\ \hline 
\end{tabular}
}
\label{Table.A.PreDic_ResNet}
\end{table}
\end{center}
\end{appendices}

\end{document}